\documentclass[10pt,letterpaper]{article}
\usepackage[top=0.8in,bottom=0.8in,left=1in,right=1in]{geometry}
\usepackage[authoryear,square]{natbib}
\setcitestyle{citesep={,}}
\usepackage{amsmath,amsfonts,bm}

\def\eqref#1{equation~\ref{#1}}
\def\1{\bm{1}}

\DeclareMathAlphabet{\mathsfit}{\encodingdefault}{\sfdefault}{m}{sl}
\SetMathAlphabet{\mathsfit}{bold}{\encodingdefault}{\sfdefault}{bx}{n}

\newcommand{\E}{\mathbb{E}}

\newcommand{\R}{\mathbb{R}}

\DeclareMathOperator*{\argmin}{arg\,min}

\renewcommand{\eqref}[1]{\textup{(\ref{#1})}}

\usepackage{hyperref}
\hypersetup{
  hidelinks,
  pdftitle={Bellman Conformal Inference: Calibrating Prediction Intervals For Time Series},
  pdfauthor={Zitong Yang, Yu Ming Huang, Emmanuel Candes, Lihua Lei}
}
\usepackage{url}

\usepackage{amsthm}
\usepackage{amsmath, amsfonts, amssymb}
\usepackage{algorithm, algorithmic} 
\usepackage{multirow}               
\usepackage{bm}                     

\theoremstyle{plain}
\newtheorem{theorem}{Theorem}[section]
\newtheorem{proposition}{Proposition}[section]
\newtheorem{lemma}[theorem]{Lemma}

\theoremstyle{definition}
\newtheorem{definition}{Definition}[section]
\newtheorem{assumption}{Assumption}[section]

\renewcommand{\E}{\mathbb{E}}

\renewcommand{\R}{\mathbb R}
\newcommand{\TV}{\mathrm{TV}}
\newcommand{\Law}{\mathrm{Law}}
\newcommand{\Leb}{\mathrm{Leb}}
\newcommand{\Xsf}{\mathsf X}
\newcommand{\I}{\Lambda}
\newcommand{\Esf}{\mathcal E}
\newcommand{\bX}{\mathbf{X}}  
\newcommand{\F}{\mathcal{F}}

\newcommand{\btheta}{\boldsymbol{\theta}}
          
\newcommand{\ecc}{\mathrm{ecc}}      

\newcommand{\cY}{\mathcal{Y}}
\newcommand{\cF}{\mathcal{F}}
\usepackage{amssymb}
\renewcommand{\P}{\mathbb{P}}
\newcommand{\cN}{\mathcal{N}}
\newcommand{\err}{\mathrm{err}}

\providecommand{\argmin}{}
\renewcommand{\argmin}{\operatorname{argmin}}

\theoremstyle{remark}
\newtheorem{remark}{Remark}[section]
\usepackage{booktabs}
\usepackage{graphicx}
\graphicspath{{figures/}}
\usepackage{subcaption}

\title{Bellman Conformal Inference: Calibrating Prediction Intervals For Time Series}

\author{Zitong Yang \and Yu Ming Huang \and Emmanuel Cand\`es \and Lihua Lei}
\date{}

\makeatletter
\renewcommand{\maketitle}{%
  \begin{center}
    {\Large\bfseries \@title\par}
    \vspace{0.25in}
    {\large
      \begin{tabular}{@{}c@{}}
        Zitong Yang\textsuperscript{$\ast$}\quad
        Yu Ming Huang\textsuperscript{$\ddagger$}\quad
        Emmanuel Cand\`es\textsuperscript{$\diamond$}\quad
        Lihua Lei\textsuperscript{$\dagger\diamond$}\\
        \texttt{\char123 zitong, yumingh, candes, lihualei\char125 @stanford.edu}\\
        \textsuperscript{$\ast$}OpenAI\\
        \textsuperscript{$\ddagger$}Institute for Computational and Mathematical Engineering, Stanford University\\
        \textsuperscript{$\diamond$}Department of Statistics, Stanford University\\
        \textsuperscript{$\dagger$}Stanford Graduate School of Business
      \end{tabular}\par}
  \end{center}
}
\makeatother

\begin{document}

\maketitle

\begin{abstract}
  We introduce Bellman Conformal Inference (BCI), a framework that wraps around any time series forecasting models and provides calibrated prediction intervals.
  Unlike the existing methods, BCI is able to leverage multi-step ahead forecasts and explicitly optimize the average interval lengths by solving a one-dimensional stochastic control problem (SCP) at each time step.
  In particular, we use the dynamic programming algorithm to find the optimal policy for the SCP.
  We prove that BCI achieves long-term coverage under arbitrary distribution shifts and temporal dependence, even with poor multi-step ahead forecasts. When the raw forecast intervals are approximately calibrated, we prove that BCI is near optimal for its surrogate model predictive control problem.
  % In addition, we can lift the long-term average coverage guarantee to a short-term average marginal coverage guarantee when the time series is stationary.
We find empirically that BCI avoids uninformative intervals with infinite lengths under our main experimental configuration and generates substantially shorter prediction intervals when the raw forecast intervals are poorly calibrated.  \end{abstract}

\section{Introduction}

\begin{figure}[t]
    \centering
    \begin{subfigure}[b]{0.32\textwidth}
        \centering
        \includegraphics[width=\textwidth]{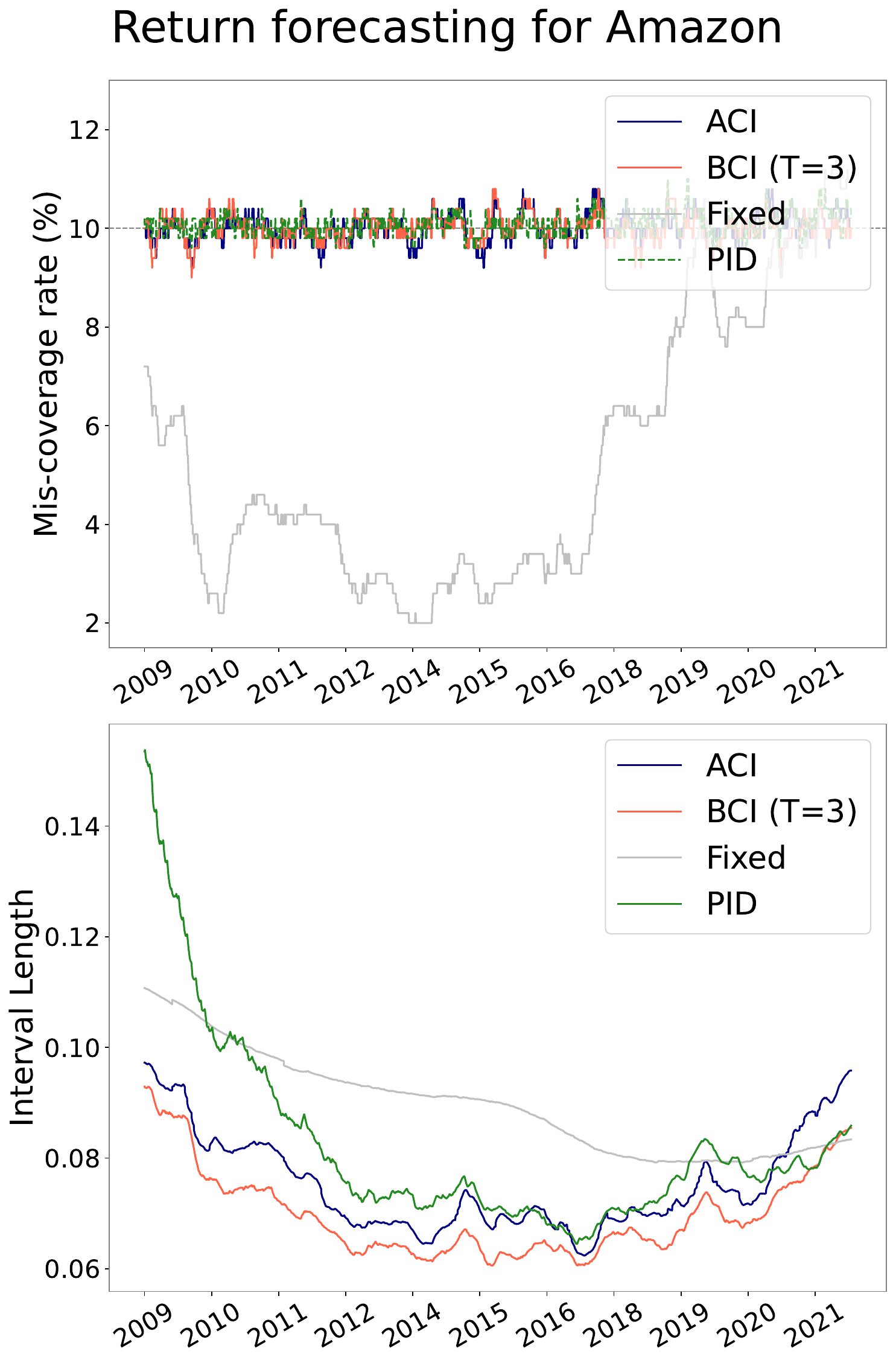}
        \caption{Return forecasting}
        \label{fig:rtfc-AMD-tight}
    \end{subfigure}
    \hfill
    \begin{subfigure}[b]{0.32\textwidth}
        \centering
        \includegraphics[width=\textwidth]{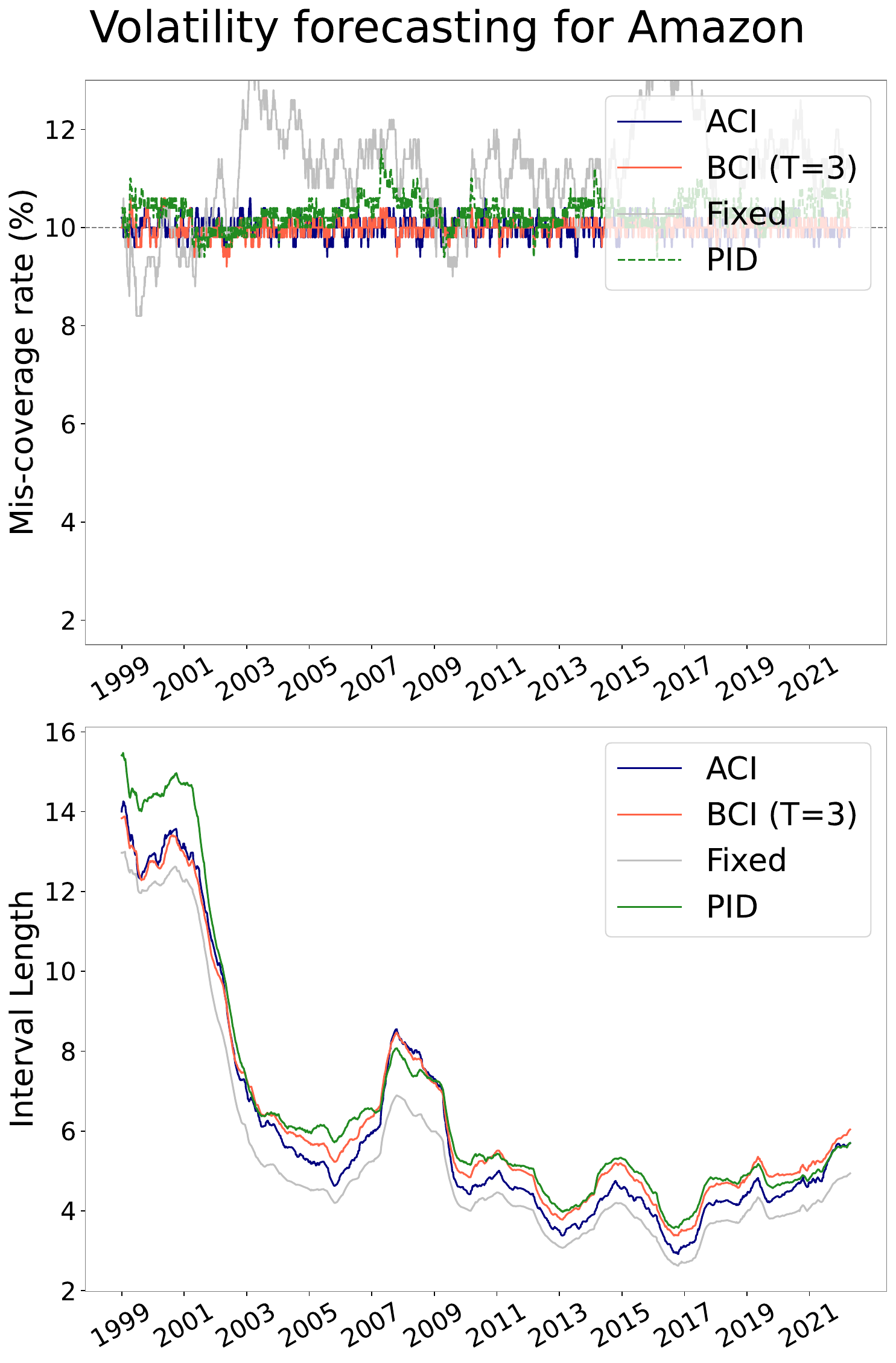}
        \caption{Volatility forecasting}
        \label{fig:vlfc-Amazon-tight}
    \end{subfigure}
    \hfill
    \begin{subfigure}[b]{0.32\textwidth}
        \centering
        \includegraphics[width=\textwidth]{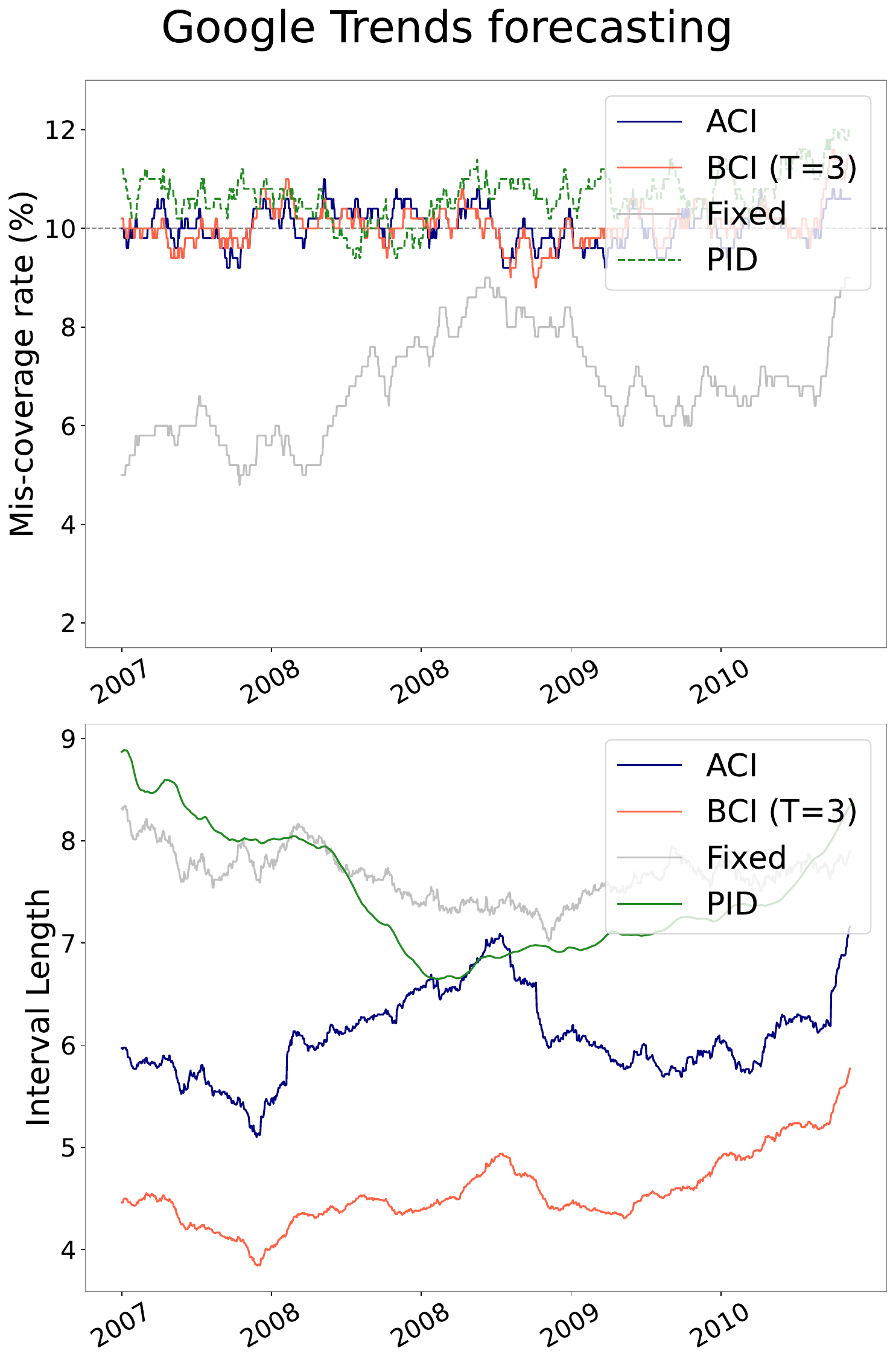}
        \caption{Google trend}
        \label{fig:google-trend-tight}
    \end{subfigure}
    \caption{Online time series forecasting for three different tasks: return forecasting on
    Amazon stock data, volatility forecasting on Amazon stock data, and Google search popularity
    data for the keyword ``deep learning''. \textbf{Top panel:} moving averages of miscoverage
    rates over 500 data points. \textbf{Bottom panel:} moving averages of prediction interval
    lengths. In all figures, the red and green curves correspond to our proposed BCI algorithm
    and the conformal PID method; the blue curves correspond to the ACI algorithm with stepsize
    $0.1$; the gray curves correspond to setting $\alpha_t=\overline{\alpha}$ for all $t$.}
    \label{fig:exp-main-tight}
\end{figure}

Uncertainty quantification for time series nowcasting and forecasting is crucial in many areas such as climate science, epidemiology, industrial engineering, and macroeconomics. 
Ideally, the forecaster would generate a prediction interval at each time period that is calibrated in the sense that the fraction of intervals covering the true outcomes is approximately equal to the target coverage level in the long run.
Classical approaches for generating prediction intervals are mostly model-based \cite{boxjen76, ar82, stowat10, Brown1964SmoothingFA, oscar05}.
However, time series models are often mis-specified due to non-stationarity or changing environments. 
As a result, the model-based prediction intervals tend to be poorly calibrated (see for instance the gray curves in Figure \ref{fig:exp-main-tight}).
Moreover, many forecasters have upgraded their workflows by incorporating black-box machine learning algorithms \citep[e.g.][]{taylor2018forecasting, makridakis2018statistical, darts}, for which valid uncertainty quantification proves to be challenging.

Due to complex temporal dependence and distribution shifts, distribution-free uncertainty quantification techniques such as conformal inference \citep[e.g.][]{saunders1999transduction, papadopoulos2002inductive, vovk05, lei2013conformal, lei2018distribution, angelopoulos2021gentle} are not guaranteed to achieve calibration. While many variants of conformal inference have been developed under weaker assumptions, most require restrictive dependence structure \cite{chernozhukov2018exact,tibshirani2019conformal, lei2021conformal, oliveira2022split, candes2023conformalized}, or limited distribution shifts \cite{barber2023conformal}, or accurate model estimates \cite{xu2021conformal, xu2023sequential}, or multiple independent copies of the time series \cite{stankeviciute2021conformal, dietterich2022conformal, sun2023copula}. 

An important departure from the aforementioned methods is Adaptive Conformal Inference (ACI) proposed by  \citet{gibbs2021adaptive}, as well as its variants \cite{gibbs2022conformal, zaffran2022adaptive, rolrc, angelopoulos2023conformal}, which can produce approximately calibrated prediction intervals without making any assumptions on the time series.
ACI generalizes standard conformal prediction for exchangeable data by choosing a time-varying nominal miscoverage rate to achieve calibration.
Alternatively, ACI can be formulated as an online gradient update that adjusts the nominal miscoverage rate downwards following each failure to cover the true outcome and upwards otherwise. 

While the appeal of assumption-free calibration is evident, ACI lacks a  mechanism to explicitly optimize the average interval lengths. 
For most practical time series forecasting models, multi-step ahead prediction intervals are readily available \cite{boxjen76,fan2003nonlinear, west2006bayesian, politis2023multi}. These intervals could be used to trade off between present and future interval lengths. For instance, if a two-day ahead interval accurately represents the one-day ahead interval for the following day, its length (at any given nominal miscoverage rate) could provide valuable guidance for judiciously selecting the nominal miscoverage rate to prevent unnecessarily wide intervals in two days. In this paper, we introduce Bellman Conformal Inference (BCI), which wraps around any multi-step ahead prediction intervals and formulates a stochastic control problem (SCP) to explicitly optimize the average interval lengths.
In particular, at each time point, BCI formulates an SCP by taking the nominal miscoverage rate as the action to optimize an objective function that trades off between the average length of multi-step ahead prediction intervals, as proxies for the actual future prediction intervals, and the estimated average future coverage. The SCP problem can be efficiently solved by dynamic programming (DP).

As with ACI, BCI is guaranteed to generate calibrated prediction intervals under Assumption~\ref{ass:pred-sets} that is satisfied by almost all commonly used prediction intervals -- in particular, it does not require the nominal multi-step prediction intervals to be well-calibrated.
Instead of updating the nominal miscoverage rates directly, BCI applies the online gradient update to adjust a ``security parameter'' in the SCP that controls the trade-off between average interval lengths and short-term coverage rate.
Applying an online gradient update to an alternative parameter is similar to the Rolling RC method introduced by \citet{rolrc} or the Conformal Decision Theory from \citet{lekeufack2023conformal}.
Both these works extend ACI to online risk control, akin to the extension of conformal inference in the offline setting (\citet{bates2021distribution, angelopoulos2021learn, angelopoulos2022conformal}). 

Another closely related method is the Conformal PID Control proposed by \cite{angelopoulos2023conformal}. They also extend ACI by leveraging ideas from control theory. This method modifies the ACI update rule by adding two more pieces -- error integration and scorecasting. The error integration stabilizes coverage and reduces the risk of the infinite-length intervals that commonly arise with ACI, while scorecasting enables the use of historical observations to take advantage of local stationarity. Both can be achieved by BCI, but through very different mechanisms. In particular, BCI leverages historical data to model dynamics and optimizes interval lengths explicitly.

We apply BCI to forecast stock price volatility, absolute return, and the popularity of Google search trend.
Our findings suggest that when the nominal multi-step ahead prediction intervals are poorly-calibrated, BCI can generate substantially shorter intervals compared to ACI and conformal PID, as seen in Figure \ref{fig:google-trend-tight}.
Conversely, when the nominal intervals are well-calibrated, BCI generates intervals of comparable lengths to ACI and conformal PID, and, under our main configuration, prevents the occurrence of infinitely long intervals as seen in Figure \ref{fig:vlfc-Amazon-tight}.
We discuss this correspondence in detail in Section \ref{sec:exp-results}.

\section{Setup}
\label{sec:setup}

Consider a time series  $Y_1, Y_2, \ldots, $ where $Y_t\in\cY$ is the outcome of interest the analyst wants to predict. We assume $Y_t$ is not observed until time $t+1$. Further, let $\cF_{t-1}$
 denote the $\sigma$-algebra generated by all information available at time $t$. In particular, $\cF_{t-1}$ includes all past outcomes $Y_{t-1}, Y_{t-2}, \ldots, Y_1$. It can also include other variables that have been observed before $t$.

%%%
\subsection{Multi-step ahead prediction intervals}
\label{sec:intro-multi}
The analyst applies the forecasting algorithm that can generate $T$-step ahead prediction intervals, where $T$ is a positive integer. At each time $t$ and future time $s$ for some $t\le s\le t+T-1$, let $C_{s|t}(1 - \beta)\subset \mathcal{Y}$ denote the $(1-\beta)$ prediction interval for $Y_{s}$ produced by the forecasting algorithm. Here, $\beta$ stands for the nominal miscoverage rate. We say $C_{s|t}(1-\beta)$ is a nominal prediction interval because $\P(Y_{s}\in C_{s|t}(1-\beta))= 1-\beta$ when the model behind the forecasting algorithm is correctly specified. However, when the model is misspecified, $\P(Y_{s}\in C_{s|t}(1-\beta))$ might deviate from $1 - \beta$.
Mose time series models generate multi-step prediction intervals. Below are two examples.

% \textbf{ARMA model} \cite{boxjen76, fanyao03}: an ARMA($p$, $q$) model assumes 
% \[
% Y_t = b_{1}Y_{t-1} + \ldots + b_{p}Y_{t-p} + \nu_t + a_{1}\nu_{t-1} + \ldots + a_{q}\nu_{t-q},
% \]
% where $\nu_1, \nu_2, \ldots \stackrel{i.i.d.}{\sim} \cN(0, \sigma^2)$. Then 
% $C_{s|t}(\beta) = [\hat{Y}_{t+s-1} + z_{\beta/2}\hat{\sigma}_{t+s-1}, \hat{Y}_{t+s-1} + z_{1-\beta/2}\hat{\sigma}_{t+s-1}]$, where $\hat{Y}_{t+s-1}$ is the estimate of the mean of $Y_{t+s-1}$ conditional on the past, $\hat{\sigma}_{t+s-1}$ is the estimated standard deviation of $Y_{t+s-1}-\hat{Y}_{t+s-1}$, and $z_{\beta/2}$ is the $\beta/2$-th quantile of the standard normal distribution.

\textbf{GARCH model} \cite{engle1982autoregressive, bollerslev1986generalized}: a GARCH($p$, $q$) model is a generalization of the ARMA model that allows for conditional heteroskedasticitiy. In particular, it assumes $Y_t \sim \cN(0, \sigma_{t}^2)$ where 
\[\sigma_{t}^2 = \omega + b_1\sigma_{t-1}^2 + \ldots + b_p\sigma_{t-p}^2 + a_1\epsilon_{t-1}^2 + \ldots + a_q\epsilon_{t-q}^2,\]
and $\epsilon_1, \epsilon_2, \ldots \stackrel{i.i.d.}{\sim} \cN(0, \sigma^2)$.
When the GARCH model is applied to forecast $Y_t^2$, $C_{s|t}(1-\beta)$ can be formed by the $(\beta/2)$-th and $(1-\beta/2)$-th quantile of $(\hat{\sigma}_{t+s-1} Z)^2$ where $Z \sim \cN(0, 1)$ and $\hat{\sigma}_{t+s-1}^2$ is the estimate of $\sigma_{t+s-1}^2$.

% \noindent Even without a complete time series model, we can obtain multi-step ahead prediction intervals by fitting a generic likelihood function. We will also use this approach in the experiments later.

\textbf{Sequence-to-sequence neural networks} \cite{lstm, deepar, transformer}: this approach first models the marginal distribution of each $Y_t$ (e.g., $Y_t\sim \cN(\mu_t, \sigma_t^2)$). Then it uses a sequence-to-sequence neural network, such as a transformer, to predict the likelihood parameters of a future observation $Y_t$ (e.g. $\hat{\mu}_t, \hat{\sigma}_t$) from the lagged observations $\{Y_{t-1}, Y_{t-2}, \dots\}$. The prediction intervals $C_{s|t}(1-\beta)$ can be formed using $\beta/2$ and $1-\beta/2$-th quantiles of $Y_t$ with likelihood parameters imputed by the model forecast.

Throughout the paper we make the following mild assumptions on the nominal prediction intervals, akin to \citet{gupta2022nested} and \citet{rolrc}. 

\begin{assumption}\label{ass:pred-sets}
 For any $t,s\ge 1$, the prediction interval $C_{s|t}(1-\beta)\subset \mathcal{Y}$ satisfies the following three conditions:\\
$~~~~\bullet$~ \textbf{Continuity:} Almost surely, $\beta\mapsto 1\{Y_s\in C_{s|t}(1-\beta)\}$ is right-continuous on $[0,1)$; We set $C_{s|t}(1-\beta)=C_{s|t}(1)$ for all $\beta<0$ and $C_{s|t}(1-\beta)=C_{s|t}(0)$ for all $\beta>1$.\\
$~~~~\bullet$~ \textbf{Monotonicity:} $C_{s|t}(1-\beta_1)\subset C_{s|t}(1-\beta_2)$ if $\beta_1>\beta_2$;\\
$~~~~\bullet$~ \textbf{Safeguard:} $C_{s|t}(1-0)=\cY, C_{s|t}(1-1)=\emptyset$, i.e., $\P(Y\in C_{s|t}(1))=1$,  $\P(Y\in C_{s|t}(0))=0$.
\end{assumption}

The monotonicity condition assumes that a smaller nominal miscoverage rate gives a wider interval, which is a natural condition for any reasonable prediction interval.
The safeguard condition assumes that it is always safe to set $\beta = 0$.
It is important for the coverage guarantee in Theorem~\ref{thm:global-guarantee}.
Otherwise, nature can always choose the time series adversarially to escape the prediction interval.
The monotonicity condition can be enforced by redefining $C_{s|t}(1-\beta) = \bigcup_{\beta\le \beta'}C_{s|t}(1-\beta')$ and $C_{s|t}(1) = \mathcal{Y}$.

\subsection{Calibrating prediction intervals}
Given the multi-step ahead prediction intervals $C_{s|t}(\cdot)$, BCI selects a nominal miscoverage index $\alpha_t$ and outputs $C_{t|t}(1-\alpha_t)\subset\cY$ as the prediction interval for $Y_t$.
We write $C_t(\cdot):=C_{t|t}(\cdot)$ for notational convenience.
Following \citet{gibbs2021adaptive}, our goal is to generate a sequence of nominal miscoverage indices $\{\alpha_t: t\in [K]\}$ such that $\alpha_t$ only depends on $\cF_{t-1}$ and 
\begin{equation}\label{eqn:global-guarantee}
    \lim_{K\to\infty} \frac{1}{K} \sum_{t=1}^K 1(Y_t\notin C_t(1-\alpha_t)) \le \overline{\alpha} \quad \text{almost surely}
\end{equation}
for some pre-specified target miscoverage level $\overline{\alpha}$ (e.g. $\overline{\alpha}=0.1$).
In particular, the inequality must hold for every joint distribution of $\{Y_t:t\ge1\}$ satisfying Assumption~\ref{ass:pred-sets}, including deterministic sequences satisfying these conditions.
The ACI algorithm chooses an initial nominal coverage $\alpha_0$ and updates $\alpha_t$ by 
\begin{equation}\label{eqn:aci}
\alpha_{t} = \alpha_{t-1} + \gamma(\overline{\alpha} - \err_{t-1}), 
\end{equation}
where $\err_{t-1} = 1(Y_{t-1}\notin C_{t-1}(1-\alpha_{t-1}))$.
Above, $\gamma$ is a fixed stepsize, which can be made data-adaptive using more advanced techniques in online learning \cite{gibbs2022conformal}. \citet{gibbs2021adaptive} prove that, the average miscoverage over the first $T$ time periods is at most $\overline{\alpha} + 2/T\gamma$.

Let $\beta_t$ be the supremum of the nominal miscoverage rates at which $Y_t$ is covered,
\begin{equation}\label{eq:beta_t}
\beta_t = \sup_{Y_t\in C_t(1-\beta)} \beta.
\end{equation}
We shall refer to $\beta_t$ as the uncalibrated probability inverse transform (PIT). 
Note that at time $t$, $\beta_{t-1}$ is observed as $Y_{t-1}$ is.
Under Assumption~\ref{ass:pred-sets}, right-continuity and the safeguard at $a=0$ give $\beta_t>0$. Monotonicity gives coverage for $0\le a<\beta_t$; right-continuity excludes coverage at $a=\beta_t<1$, and the safeguard excludes it at $a=\beta_t=1$. Thus $\{a\in[0,1]:Y_t\in C_t(1-a)\}=[0,\beta_t)$ with $0<\beta_t\le1$, and $\err_t=1\{Y_t\notin C_t(1-\alpha_t)\}=1\{\alpha_t\ge\beta_t\}$, almost surely.
Thus, for ACI, $\alpha_{t}$ depends exclusively on $(\alpha_{t-1}, \beta_{t-1})$.
While ACI indeed provides a valid calibration guarantee, it disregards two sources of information:
the first is the length of $C_t(1-\beta)$ and multi-step ahead intervals, which can inform a better choice of $\alpha_{t}$.
The second is the historical sequence of uncalibrated PITs $\{\beta_j\}_{j<t}$ that allows a better estimate of $\beta_{t+1}$ than just using $\beta_t$ alone.
% In particular, \citet{angelopoulos2023conformal}, which also take a control perspective from PID (Proportional–Integral–Derivative) controller, show that incorporating the historical errors into the update \eqref{eqn:aci} can stabilize the large variability in the prediction intervals generated by ACI. 
As we shall see, the BCI method we introduce in this paper chooses $\alpha_{t}$ as a function of all available information including the past actions $\alpha_{t-1}, \alpha_{t-2}, \ldots$, the past uncalibrated PITs $\beta_{t-1}, \beta_{t-2}, \ldots, $ and the multi-step ahead intervals $C_{t|t}(\cdot), \dots, C_{t+T-1|t}(\cdot)$.

\section{Bellman Conformal Inference}
\label{sec:bci}

% \subsection{More general formulation of BCI}

%
\subsection{BCI as Model Predictive Control}
\label{sec:bci-as-mpc}
At a high level, BCI is analogous to Model Predictive Control (MPC) \citep{mpc}. At each time point $t$, BCI models the ``dynamics'' of the process $(Y_1, \beta_1), (Y_2, \beta_2), \ldots $ from past observations, simulates the ``system'' over the next $T$ steps, and plans the ``control'' $\alpha_t$ by minimizing the ``cost'' driven by both the average interval length and miscoverage. Unlike traditional reactive control approaches, the proactive MPC approach is more suitable for forecasting problems under substantial distribution shifts.

To set up the cost minimization problem at time $t$, we denote by $L_{s|t}(\beta) = |C_{s|t}(1 - \beta)|$ the function that maps the miscoverage rate to the length of the nominal multi-step ahead prediction interval for $Y_{s}$ and $F_{s|t}$ the marginal distribution of $\beta_{s}$ estimated using the past observations. We do not make any assumption on how $L_{s|t}(\cdot)$ and $F_{s|t}$ are generated. For all experiments in this paper, we simply set $F_{s|t}=\widehat G_t^{(B)}$ where $\widehat G_t^{(B)}$ is the empirical CDF of $\{\beta_{t-1}, \ldots, \beta_{t-B}\}$ for some large $B$.
In our simulation, we set $B=100$.

At time $t$, we solve the following optimization problem:
\begin{align}
&\min_{\alpha_{t|t},\dots,\alpha_{t+T-1|t}} 
\mathbb{E}
\Bigg[ \underbrace{\sum_{s=t}^{t+T-1} L_{s|t}(\alpha_{s|t})}_{\text{Efficiency: interval length}}+ \lambda_t \underbrace{\max\left\{\frac{1}{T}\sum_{s=t}^{t+T-1}\err_{s|t} - \bar{\alpha}, 0\right\}}_{\text{Validity: miscoverage rate}}\Bigg]+\tau K(\alpha_{t|t},\bar\alpha), \label{eqn:cost}
\end{align}

\noindent where $(\beta_{t|t},\dots,\beta_{t+T-1|t})\sim F_{t|t}\otimes\cdots\otimes F_{t+T-1|t}$, $\alpha_{s|t}$ denotes the planned action for time $s$, $\err_{s|t} = 1(\alpha_{s|t} \ge \beta_{s|t})$ denotes the error indicator for the realized $\beta_{s|t}$, and $\lambda_t$ denotes the relative weight on the miscoverage that BCI uses to achieve the coverage guarantee, and $\tau K(\alpha_{t|t},\bar\alpha)$ is a regularization term. Let $\tau\ge0$ and suppose $K(\cdot,\bar\alpha)$ is nonnegative and finite on $[0,1]$. Following the convention of stochastic MPC, the minimization in \eqref{eqn:cost} is over the policies, namely functions of the state, described in Section~\ref{sec:bci-dynamic-programing}, rather than over a fixed vector. We will set $\tau=0$ for the applications and consider $\tau>0$ and a KL regularizer in Appendix~\ref{subsec:dominance}. When the minimum is not attained, we select a solution whose objective value is within $\epsilon$ of the infimum in \eqref{eqn:cost}, for a sufficiently small $\epsilon>0$.

Throughout the paper, we assume that the selected solution at time $t$ is $\cF_{t-1}$-measurable. The first term measures the average interval length and the second term measures the rescaled average miscoverage rate within the receding horizon. Here we assume $\beta_{t|t}, \ldots, \beta_{t+T-1|t}$ are independent so that the cost function only depends on the marginal distributions. This is not required for the coverage guarantee (see Theorem \ref{thm:global-guarantee}) but makes the cost-minimization problem more tractable.

\begin{remark}\label{rem:ecdf_beta}
For BCI, the above argument leads to a simple i.i.d. model for the dynamics of $\beta_{t+s|t}$ as well as a simple estimator -- the empirical distribution of past $\beta_t$s. Thus, BCI does not rely on another machine learning algorithm to learn the dynamics. By contrast, even when the underlying prediction intervals are perfectly calibrated, the dynamics of the conformity scores remain unspecified. We present a theoretical result (Theorem \ref{thm:oracle}) that justifies this choice.
\end{remark} 

When $\tau=0$, minimizers can be selected so that
\begin{equation}\label{eqn:trivial_property}
\lambda_t \le 0\Longrightarrow \alpha_{s|t}^{*} = 1, \,\, s = t, \ldots, t+T-1,
\end{equation}
since for a negative miscoverage penalty $\lambda_t$, the interval length and the negative miscoverage rate are both decreasing in $\alpha_{s|t}$, and hence the optimum is achieved at $\alpha_{s|t} = 1$.
For $\tau>0$, \eqref{eqn:trivial_property} is imposed as a safeguard instead of minimizing \eqref{eqn:cost}.
The analyst can apply any algorithm to minimize \eqref{eqn:cost} as long as it satisfies \eqref{eqn:trivial_property}. In particular, we discuss a DP-based algorithm in Section \ref{sec:bci-dynamic-programing} that provides the exact solution and is computationally efficient for moderate values of $T$. Let $\alpha_{s|t}^{*}$ be the solution, which does not have to be exact. Clearly, $\alpha_{s|t}^{*}$ only depends on the model forecasts $(L_{s|t}(\cdot), F_{s|t})$ and the relative weight $\lambda_t$. The standard MPC would set $\alpha_t = \alpha_{t|t}^{*}$ and discard all other planned actions. To achieve the coverage guarantee in Theorem~\ref{thm:global-guarantee}, BCI would modify $\alpha_t$ by considering both $\alpha_{t|t}^{*}$ and $\lambda_t$.

%% 
% \subsection{BCI update rule for $\lambda_t$ and $\alpha_t$}
% arXiv
\subsection{BCI update rule for \texorpdfstring{$\lambda_t$}{Lg} and \texorpdfstring{$\alpha_t$}{Lg}}
\label{sec:bci-valid}

The key lever of BCI is the relative weight $\lambda_t$ in \eqref{eqn:cost}. Intuitively, we shall increase $\lambda_t$ to penalize the miscoverage term more when errors occur too often and decrease $\lambda_t$ otherwise. Motivated by ACI \cite{gibbs2021adaptive} and the Rolling RC method \cite{rolrc}, we update $\lambda_t$ as follows:
\begin{equation}\label{eqn:lambda-update}
\lambda_{t+1} = \lambda_t - \gamma[\overline{\alpha} - \err_t],
\end{equation}
where $\gamma$ is a user-defined step size that controls how fast the procedure adapts to past errors and $\err_t = 1(\alpha_t \ge \beta_t)$ as before.

As mentioned in Section \ref{sec:setup}, the safeguard property in Assumption \ref{ass:pred-sets} is crucial to guarantee long-term coverage even in the adversarial case. Here, we force $\alpha_t$ to be zero, so that $C_{t|t}(1-\alpha_t) = \mathcal{Y}$ under Assumption \ref{ass:pred-sets}, when $\lambda_t$ passes some pre-specified threshold $\lambda_{\max} > 0$:
\begin{equation}\label{eqn:alpha-t-truncate}
\alpha_t =
\begin{cases}
0, &~\text{if}~ \lambda_t \geq \lambda_{\max}, \\
\alpha^\star_{t|t}, &~\text{otherwise},
\end{cases}
\end{equation}
where $\alpha_{t|t}^{*}$ is the minimizer of \eqref{eqn:cost}.

To match the scale, we can set $\gamma = c\lambda_{\max}$ for some $c\in (0, 1)$. Unlike ACI that updates $\alpha_t$ directly through a step of online gradient descent, BCI updates $\alpha_t$ indirectly through an online gradient descent-type update on $\lambda_t$. This allows BCI to take input other than past $(\alpha_t, \beta_t)$'s.
% We illustrate the difference between ACI and BCI in Figure \ref{fig:illustration}. 

% \begin{figure}[t]
%     \centering
%     \subfigure[ACI\label{fig:illustration-aci}]{\includegraphics[width=0.35\textwidth]{figures/cartoon/ACI.pdf}}
%     \subfigure[BCI\label{fig:illustration-model-aci}]{\includegraphics[width=0.35\textwidth]{figures/cartoon/modelACI.pdf}}
%     \caption{Schematic illustrations of standard ACI and BCI.}
%     \label{fig:illustration}
% \end{figure}

%%
\subsection{Solving \texorpdfstring{$\alpha_{t|t}^{*}$}{Lg} via dynamic programming}
\label{sec:bci-dynamic-programing}
When the value of $T$ is moderate (e.g. $T$=3), we can minimize \eqref{eqn:cost} exactly and efficiently by DP. In fact, we do not recommend choosing a large $T$ because the multi-step ahead prediction intervals tend to be less accurate or informative for events way into the future. For all our experiments, we choose $T = 3$. 

The reader may refer to Appendix \ref{sec:dp} for some background on the usage of DP in solving stochastic control problems. To set up the DP algorithm for our problem, we define the state variable $\rho_{s|t}$ as 
\[\rho_{t|t} = 0, \, \rho_{s|t} = \sum_{k=t}^{s-1}1(\alpha_{k|t} \ge \beta_{k|t}), \, s = t+1, \ldots, t+T.\]
The dynamics for the state variable can be written as 
\begin{equation}
\rho_{s+1|t}= \rho_{s|t} + 1(\alpha_{s|t} \ge \beta_{s|t}).
\label{eq:bci-state-dynamics}
\end{equation}
The minimization in \eqref{eqn:cost} is over collections of functions $\pi_{s|t}:\{0,\ldots,s-t\}\to[0,1]$, $s=t,\ldots,t+T-1$, with $\alpha_{s|t}=\pi_{s|t}(\rho_{s|t})$. Each $\pi_{s|t}$ is chosen at time $t$ and specifies $\alpha_{s|t}$ for every possible value of $\rho_{s|t}$. Specifically, let $\Pi_t$ denote the class of policies $\pi=(\pi_{t|t},\ldots,\pi_{t+T-1|t})$.
For $\pi\in\Pi_t$ and $\beta=(\beta_{t|t},\ldots,\beta_{t+T-1|t})\in[0,1]^T$, write
$\rho_{s|t}^{\pi}(\beta)$ and $\alpha_{s|t}^{\pi}(\beta)$ for the corresponding
states and nominal miscoverage levels.
For a distribution $P$ on $[0,1]^T$, define
\begin{equation}
J_t^P(\pi):=\E_{\beta\sim P}\left[
\sum_{s=t}^{t+T-1}L_{s|t}\bigl(\alpha_{s|t}^{\pi}(\beta)\bigr)
+\lambda_t\left(\frac{\rho_{t+T|t}^{\pi}(\beta)}{T}-\bar\alpha\right)_+
\right]+\tau K\bigl(\pi_{t|t}(0),\bar\alpha\bigr).
\label{eq:policy-objective}
\end{equation}
Thus \eqref{eqn:cost} minimizes $J_t^P(\pi)$ over $\pi\in\Pi_t$, with $P=F_{t|t}\otimes\cdots\otimes F_{t+T-1|t}$.

The terminal cost is
\begin{equation}
J_{t+T|t}(\rho)=\lambda_t\max\{\rho/T-\bar\alpha,0\},
\qquad \rho\in\{0,\ldots,T\}.
\label{eq:bci-terminal-cost}
\end{equation}
For $s=t+T-1,\ldots,t$ and $\rho\in\{0,\ldots,s-t\}$,
$J_{s|t}(\rho)$ is the continuation cost at time $s$ onwards,
given $\rho_{s|t}=\rho$. Then the standard Bellman equation gives
\begin{equation}
J_{s|t}(\rho)=\inf_{\alpha\in[0,1]}\Big\{
L_{s|t}(\alpha)+\tau1\{s=t\}K(\alpha,\bar\alpha)
+\E_{\beta_{s|t}\sim F_{s|t}}
J_{s+1|t}\bigl(\rho+1\{\alpha\ge\beta_{s|t}\}\bigr)\Big\}.
\label{eqn:bci-bellman-update}
\end{equation}
Expanding the expectation in \eqref{eqn:bci-bellman-update} gives
\[
J_{s|t}(\rho)=\inf_{\alpha\in[0,1]}\Big\{
L_{s|t}(\alpha)+\tau1\{s=t\}K(\alpha,\bar\alpha)
+\bigl(1-F_{s|t}(\alpha)\bigr)J_{s+1|t}(\rho)
+F_{s|t}(\alpha)J_{s+1|t}(\rho+1)\Big\}.
\]
Define $D_{s|t}(\rho)=J_{s+1|t}(\rho+1)-J_{s+1|t}(\rho)$. Then
\[
J_{s|t}(\rho)=J_{s+1|t}(\rho)+\inf_{\alpha\in[0,1]}
\left\{L_{s|t}(\alpha)+D_{s|t}(\rho)F_{s|t}(\alpha)
+\tau1\{s=t\}K(\alpha,\bar\alpha)\right\}.
\]
When the infimum is attained, select
\begin{equation}\label{eqn:tilde_alpha}
\tilde\alpha_{s|t}(\rho)\in\argmin_{\alpha\in[0,1]}
\left\{L_{s|t}(\alpha)+D_{s|t}(\rho)F_{s|t}(\alpha)
+\tau1\{s=t\}K(\alpha,\bar\alpha)\right\}.
\end{equation}
The following proposition shows $D_{s|t}(\rho)$ in \eqref{eqn:tilde_alpha} is always non-negative.

\begin{proposition}\label{prop:Dt-increasing}
Suppose $\lambda_t\ge0$, $\tau\ge0$, and $K(\cdot,\bar\alpha)$ is nonnegative
and finite on $[0,1]$. For $s=t,\ldots,t+T$, $J_{s|t}(\rho)$ is finite,
nonnegative, and non-decreasing on $\{0,\ldots,s-t\}$. Consequently,
$D_{s|t}(\rho)\ge0$ for $s=t,\ldots,t+T-1$ and $\rho=0,\ldots,s-t$.
\end{proposition}
The proof is in Appendix~\ref{subsec:proof-dt-increasing}.
By Proposition~\ref{prop:Dt-increasing}, $D_{s|t}(\rho)F_{s|t}(\alpha)$ is
non-decreasing in $\alpha$, whereas $L_{s|t}(\alpha)$ is non-increasing.
These two terms in \eqref{eqn:tilde_alpha} express the efficiency-coverage
tradeoff.
Recall that we just need the first component $\alpha_{t|t}^{*}=\tilde{\alpha}_{t|t}(0)$.
Since $\rho_{s|t}$ is integer-valued, the computation of $\tilde{\alpha}_{t|t}(0)$ involves solving $T + (T-1) + \ldots + 1 = O(T^2)$ one-dimensional optimization problems in the form of \eqref{eqn:tilde_alpha}. When $T$ is moderate, it is computationally efficient.

For the empirical model in Section~\ref{sec:bci-as-mpc}, write $\widehat J_{s|t}$ and $\widehat D_{s|t}$ for
$J_{s|t}$ and $D_{s|t}$ with
$F_{s|t}=\widehat G_t^{(B)}$. For $s=t+T-1,\ldots,t$ and $r=0,\ldots,s-t$, write
\begin{equation}
\widehat Q_{s|t}(r,a):=L_{s|t}(a)+\tau1\{s=t\}K(a,\bar\alpha)
+\bigl(1-\widehat G_t^{(B)}(a)\bigr)\widehat J_{s+1|t}(r)
+\widehat G_t^{(B)}(a)\widehat J_{s+1|t}(r+1).
\label{eq:policy-bellman-recursion}
\end{equation}
When each infimum in \eqref{eq:policy-exact-selection} is attained,
select a policy $\widehat\pi_t\in\Pi_t$ satisfying
\begin{equation}
\widehat Q_{s|t}\bigl(r,\widehat\pi_{s|t}(r)\bigr)
=\widehat J_{s|t}(r)=\inf_{a\in[0,1]}\widehat Q_{s|t}(r,a),
\quad s=t,\ldots,t+T-1,\quad r=0,\ldots,s-t.
\label{eq:policy-exact-selection}
\end{equation}

We summarize BCI with the DP algorithm described in this subsection in Algorithm \ref{alg:bci}.

\begin{algorithm*}[ht]
\caption{Bellman conformal inference at time $t$}\label{alg:bci}
\begin{algorithmic}[1]
\vspace{0.03in}
\STATE \textbf{Hyperparameters:} target miscoverage level $\overline{\alpha}$; length of receding horizon $T$; maximum weight $\lambda_{\max}$; initial value $\lambda_1\in[0,\lambda_{\max}]$; step size $\gamma=c\lambda_{\max}$ for $c\in(0,1)$.
\STATE \textbf{Input:}
Estimated marginal CDF of future uncalibrated PITs $F_{t|t}, \ldots, F_{t+T-1|t}$;\\
\quad\quad\quad Length functions for multi-step ahead prediction intervals $L_{t|t}(\cdot), \dots, L_{t+T-1|t}(\cdot)$;\\
\quad\quad\quad Weight $\lambda_{t-1}$ and miscoverage indicator $\err_{t-1}$ from the previous iteration.
\STATE \textbf{Step 1:} Update the security parameter as $\lambda_{t} = \lambda_{t-1} - \gamma[\overline{\alpha} - \err_{t-1}].
$
\STATE \textbf{Step 2:} Using $L_{s|t}$ and $\widehat G_t^{(B)}$, instantiate the stochastic control problem defined in \eqref{eqn:cost} from Section \ref{sec:bci-as-mpc}.
\STATE \textbf{Step 3:} Apply the DP algorithm described in Section \ref{sec:bci-dynamic-programing} to get $\tilde{\alpha}_{t+T-1|t}(\cdot), \ldots, \tilde{\alpha}_{t|t}(\cdot)$ 
\STATE \textbf{Output:} $\alpha_t = \tilde{\alpha}_{t|t}(0)$ if $\lambda_t \le \lambda_{\max}$ and $\alpha_t = 0$ otherwise.
\end{algorithmic}
\end{algorithm*}

\subsection{Validity Theory of BCI}

We start by proving that BCI achieves approximate average coverage control for any large consecutive blocks. The proof for Theorem \ref{thm:global-guarantee} can be found in Appendix \ref{sec:thm-proof}.
\begin{theorem}\label{thm:global-guarantee}
Consider Algorithm~\ref{alg:bci}. Under Assumption \ref{ass:pred-sets}, for any $m \ge 0$, almost surely,
\begin{equation}\label{eqn:local-coverage}
\left|\frac{1}{K} \sum_{t=m+1}^{m+K} \err_t -\overline{\alpha}\right| \leq \frac{c+1}{c K}.
\end{equation}
In particular, \eqref{eqn:global-guarantee} holds by letting $m=0$ and $K\rightarrow \infty$.
\end{theorem}

Next, we consider the case where $(Y_t)_{t\ge 1}$ is a time-homogeneous Markov process of order $p$, i.e., $Y_{t+1}\perp (Y_{t-p}, Y_{t-p-1}, \ldots)\mid (Y_t, \ldots, Y_{t-p+1})$. In this case, we prove that the average coverage guarantee can be strengthened to a guarantee on marginal coverage $\E[\err_t]$ when $t\to\infty$, insofar as $\bar{\alpha} = 1/q$ for some integer $q\ge2$ (e.g., $q = 10$ when $\bar{\alpha} = 0.1$). We present an informal version below and defer the formal statement and proofs to Appendix \ref{sec:proof_stationarity}.

\begin{theorem} [Informal]
\label{thm:stationarity}
Assume $(Y_t)_{t\ge 1}$ is a time-homogeneous Markov process of order $p$, $\bar{\alpha} = 1/q$ for some integer $q\ge2$, and $B$ is a constant. Under the technical assumptions in Appendix \ref{subsec:stationarity_assumptions}, for any $\lambda_1 \in (0, \lambda_{\max})$, as $t\to \infty$
\[
\lim_{t\to\infty}
\frac1q\sum_{j=0}^{q-1}\E[\err_{t+j}]
=
\bar\alpha.
\]
Furthermore, the joint process of the $p$ most recent observations, the $B$ most recent uncalibrated PITs, and the weight admits an invariant distribution $\overline\mu$. If
\begin{equation}
\bigl((Y_0,Y_{-1},\ldots,Y_{-(p-1)}),(\beta_{1-B},\ldots,\beta_0),\lambda_1\bigr)\sim\overline\mu,
\label{eq:stationary-joint-initialization}
\end{equation}
then $\E[\err_t]=\bar\alpha$ for every $t\ge1$.
\end{theorem}
\subsection{Efficiency theory of BCI}

Next, we show that the policy obtained from the empirical PIT distribution is close to the best policy under the true conditional law, when the nominal prediction intervals are approximately calibrated and $B$ is large. The proof is in Appendix~\ref{sec:proof_oracle}.

\begin{theorem}\label{thm:oracle}
Fix $T\ge1$, $B\ge4$, $t\ge B+1$, and $\bar\alpha\in(0,1)$. Define
\[
P_t^\star:=\Law\bigl((\beta_t,\ldots,\beta_{t+T-1})\mid\F_{t-1}\bigr),
\qquad
P_t^{(B)}:=\bigl(\widehat\mu_t^{(B)}\bigr)^{\otimes T},
\]
where $\widehat\mu_t^{(B)}$ is the empirical distribution of $(\beta_{t-1},\ldots,\beta_{t-B})$.
Assume $L_{s|t}\ge0$ and $L_{s|t}(1)=0$ for $s=t,\ldots,t+T-1$, $|\lambda_t|<\infty$, $\tau\ge0$, and $K(\cdot,\bar\alpha)$ is nonnegative and finite.
Define
\[
G_s^\circ(u):=\P(\beta_s\le u\mid\F_{s-1}),\qquad u\in[0,1].
\]
Suppose that, for all $s\ge1$,
\begin{equation}
\sup_{u\in[0,1]}|G_s^\circ(u)-u|\le\kappa_s,
\label{eq:kappa-assumption}
\end{equation}
where $(\kappa_s)_{s\ge1}$ is deterministic, nonnegative, and nonincreasing, and, for some constant $M>0$,
\begin{equation}
|G_s^\circ(u)-G_s^\circ(v)|\le M|u-v|,
\qquad u,v\in[0,1].
\label{eq:key-lipschitz}
\end{equation}
Let $\widehat\pi_t$ satisfy \eqref{eq:policy-exact-selection}.
Then, for every $\delta\in(0,1)$, with probability at least $1-\delta$,
\begin{equation}
0\le J_t^{P_t^\star}(\widehat\pi_t)
-\inf_{\pi\in\Pi_t}J_t^{P_t^\star}(\pi)
\le4|\lambda_t|\left[
\frac{\sqrt{\log(B/\delta)}+M}{\sqrt B}+\kappa_{t-B}
\right].
\label{eq:main-theorem-bound}
\end{equation}
\end{theorem}

In Appendix~\ref{subsec:dominance}, we consider a setting in which the expected interval lengths of ACI are provably larger than those of BCI. In this setting, we assume that both ACI and BCI have stationary trajectories. Such trajectories can be obtained by initializing the respective joint processes from invariant distributions, whose existence is established under the assumptions of \citet[Appendix~A.5]{gibbs2021adaptive} for ACI and Theorem~\ref{thm:stationarity} for BCI.

% the average marginal miscoverage $(1/q)\sum_{j=1}^q\mathbb{E}[\err_{t+j}]$ over $q$ converges to $\bar{\alpha} = 1/q$ for any initial $\lambda_1$ as $t\to \infty$. Moreover, we prove that $(Y_t, \Gamma_t, \lambda_t)$ admits an invariant distribution $\mu$, where $\Gamma_t = (\beta_{t-1}, \ldots, \beta_{t-B})$, and the marginal miscoverage $\E[\err_t] = \bar{\alpha}$ for every $t$ if $\lambda_1$ is sampled from $\mu$. This bridges the gap between average coverage and marginal coverage under stationarity. 

\section{Empirical results}
\label{sec:exp}

In this section, we present empirical experiments on real time series forecasting problems to demonstrate the effectiveness of BCI.
We consider auto-regressive time series forecasting task for three datasets: daily Google trend popularity for keyword \texttt{deep learning}, daily stock return for companies \texttt{AMD, Amazon, Nvidia}, and stock volatility for the same companies.
We use the ACI procedure \eqref{eqn:aci} as a baseline for comparison.
Making fair comparisons among ACI, BCI, and conformal PID is not straightforward due to the different structures of tuning parameters. To address this issue, we propose a matching method for fair comparison in Section \ref{sec:exp-comparsion-setup}.

% All experiments are done on one AMD EPYC 7502P 32-Core CPU.
% removed for neurips submission
% The code for reproducing the results in the paper can be found at \url{https://github .com/ZitongYang/bellman-conformal-inference.git}.

%%%
\subsection{Dataset and model fitting}
\label{sec:exp-data-model}

\textbf{Return forecasting.}
\label{sec:exp-rtfc}
Our first example studies the relative return of stock prices from various companies.
We download the daily stock price of companies \texttt{Amazon, AMD, Nvidia} from the \textit{Wall Street Journal Market Data}.\footnote{\url{https://www.wsj.com/market-data}}
The length of history varies from company to company. That said, for all companies we have approximately 15 years of data with roughly 250 trading days each year.
For each company, we compute the one-day log return $Y_t$ of the stock as
\[
Y_t = \log (P_t/P_{t-1}).
\]
On each day $t$, we use the lagged sequence $\{Y_{t-1}, \dots, Y_{t-100}\}$ as our predictors.
We use a decoder-only transformer \cite{transformer} of embedding size $16$ with $8$ heads and $2$ layers.
This choice leads to a head size of $16/8=2$, which makes this a low-dimensional self-attention operation that is suitable for univariate time series.
For the output, we use a linear layer to map the embedding to a $2\times T$ dimensional vector, representing the predicted mean and standard deviation of the for the next $T$ days.
Mathematically, we can think of our transformer as a mapping from $\R^{100}$ to $\R^{2\times T}$, denoted by 
$[\mu_{s|t}(\bX_t;\btheta), \sigma_{s|t}(\bX_t;\btheta)]$ for $s=t,\dots,t+T-1$.
Here, $\bX_t=\{Y_{t-1}, \dots, Y_{t-100}\}$ are the lagged returns and $\btheta$ denotes the weights of the neural network.
At time $t$, we estimate $\btheta$ by solving the following optimization problem:
\[\min_{\btheta} \sum_{\tau<T} \sum_{s=\tau-T}^{\tau-1}- \log \cN\left(Y_{t+s}; \mu_{s|t}(\bX_t;\btheta), \sigma_{s|t}(\bX_t;\btheta)\right).\]
This formulation assumes the distribution of $Y_{t+s}$ follows a normal distribution $\cN\left(\mu_{s|t}(\bX_t;\btheta), \sigma_{s|t}(\bX_t;\btheta)\right)$.
It is a straightforward adaptation from the standard techniques for training variational auto-encoders \cite{vae}.
Then we can construct $C_{s|t}(1-\beta)$ as $\bigl(\mu_{s|t}(\bX_t;\btheta)-z_{1-\beta/2}\sigma_{s|t}(\bX_t;\btheta),\,\mu_{s|t}(\bX_t;\btheta)+z_{1-\beta/2}\sigma_{s|t}(\bX_t;\btheta)\bigr)$ 
% where $z_{1-\beta/2}$ is the $(1-\beta/2)$-th quantile of the standard normal distribution.

%%
\textbf{Volatility forecasting.}
Our next example is the volatility forecasting problem explored in \citet{gibbs2021adaptive}.
We work with the same daily stock data as in the return forecasting problem.
Instead of the log return, we compute the squared volatility as
$Y_t = (P_{t}/P_{t-1}-1)^2$.
We then use the lagged volatility $\{Y_{t-1}, \dots, Y_{t-100}\}$ as a the predictors.
Instead of using modern neural networks, we use the classical time series model \texttt{GARCH(1, 1)}\cite{engle1982autoregressive, bollerslev1986generalized} that is often used to model stock price volatility.
As we will see in Section \ref{sec:exp-results}, this classic model delivers nearly (marginally) calibrated prediction intervals.
The \texttt{GARCH(1, 1)} model assumes that the squared volatility $Y_{s}$ conditioned on the current observation follows a non-central $\chi$-squared distribution and we can use its $\beta/2$- and $1-\beta/2$-th quantile to form the prediction interval $C_{s|t}(1-\beta)$.
We provide a detailed description of the \texttt{GARCH(1, 1)} model in Appendix \ref{sec:appendix-garch}.

\textbf{Google trend popularity.}
Finally, we apply BCI to non-financial data. We consider the daily Google search popularity for the keyward \texttt{deep learning}
% It is non-trivial to obtain the raw dataset as Google imposes rate limitations for querying the trend dataset.
% We obtain the daily popularity data 
from 2006 to 2011.
For this dataset, we use the same neural-network-based fitting procedure as in the return forecasting task, except that we use a 5-layer LSTM \cite{lstm} recurrent network instead of the transformer to demonstrate the compatibility of BCI with a broader class of forecasters.

\begin{figure}[h]
    \centering
    \begin{subfigure}[b]{0.32\textwidth}
        \centering
        \includegraphics[width=\textwidth]{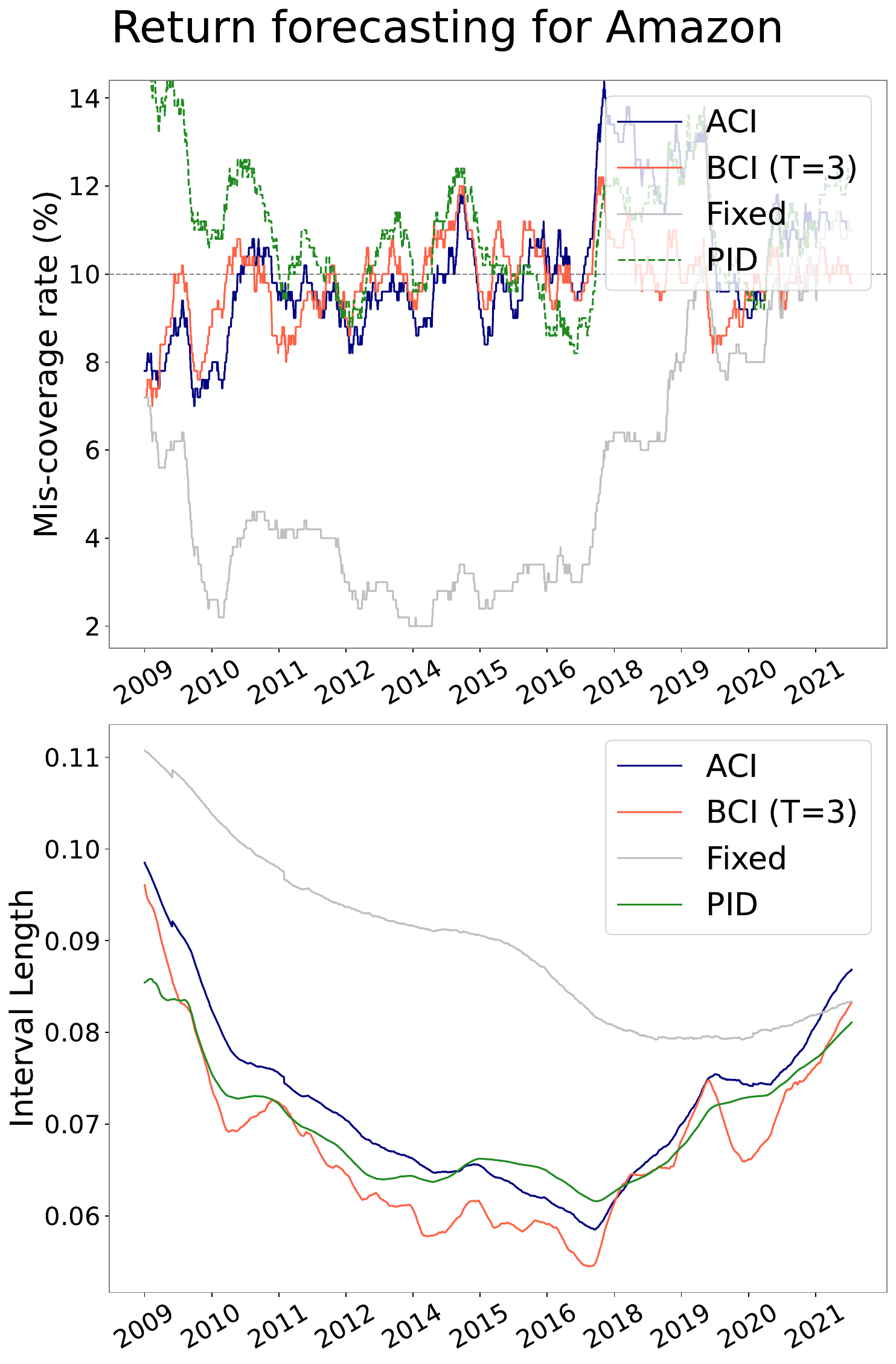}
        \caption{Return forecasting}
        \label{fig:rtfc-AMD-loose}
    \end{subfigure}
    \hfill
    \begin{subfigure}[b]{0.32\textwidth}
        \centering
        \includegraphics[width=\textwidth]{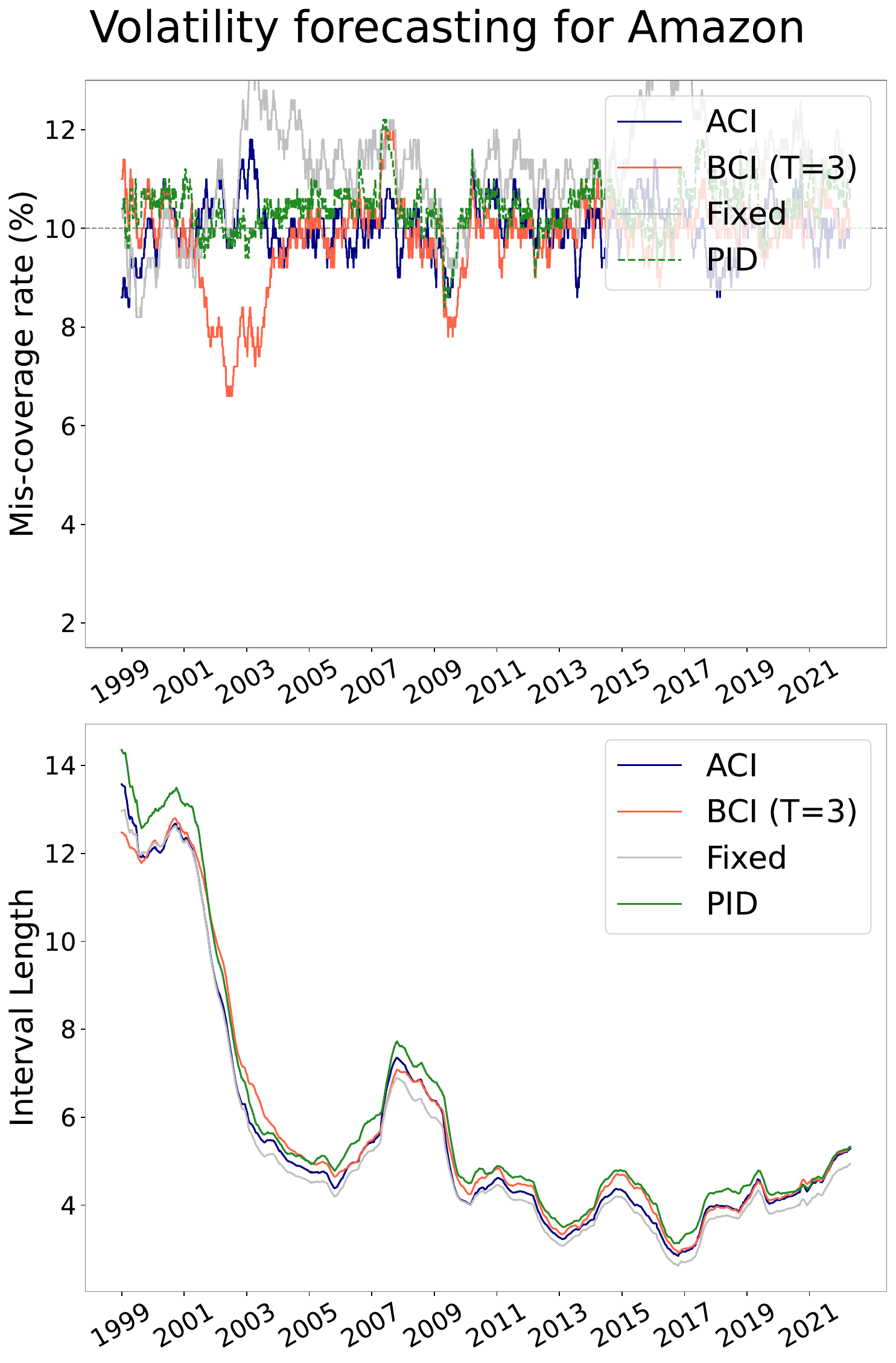}
        \caption{Volatility forecasting}
        \label{fig:vlfc-Amazon-loose}
    \end{subfigure}
    \hfill
    \begin{subfigure}[b]{0.32\textwidth}
        \centering
        \includegraphics[width=\textwidth]{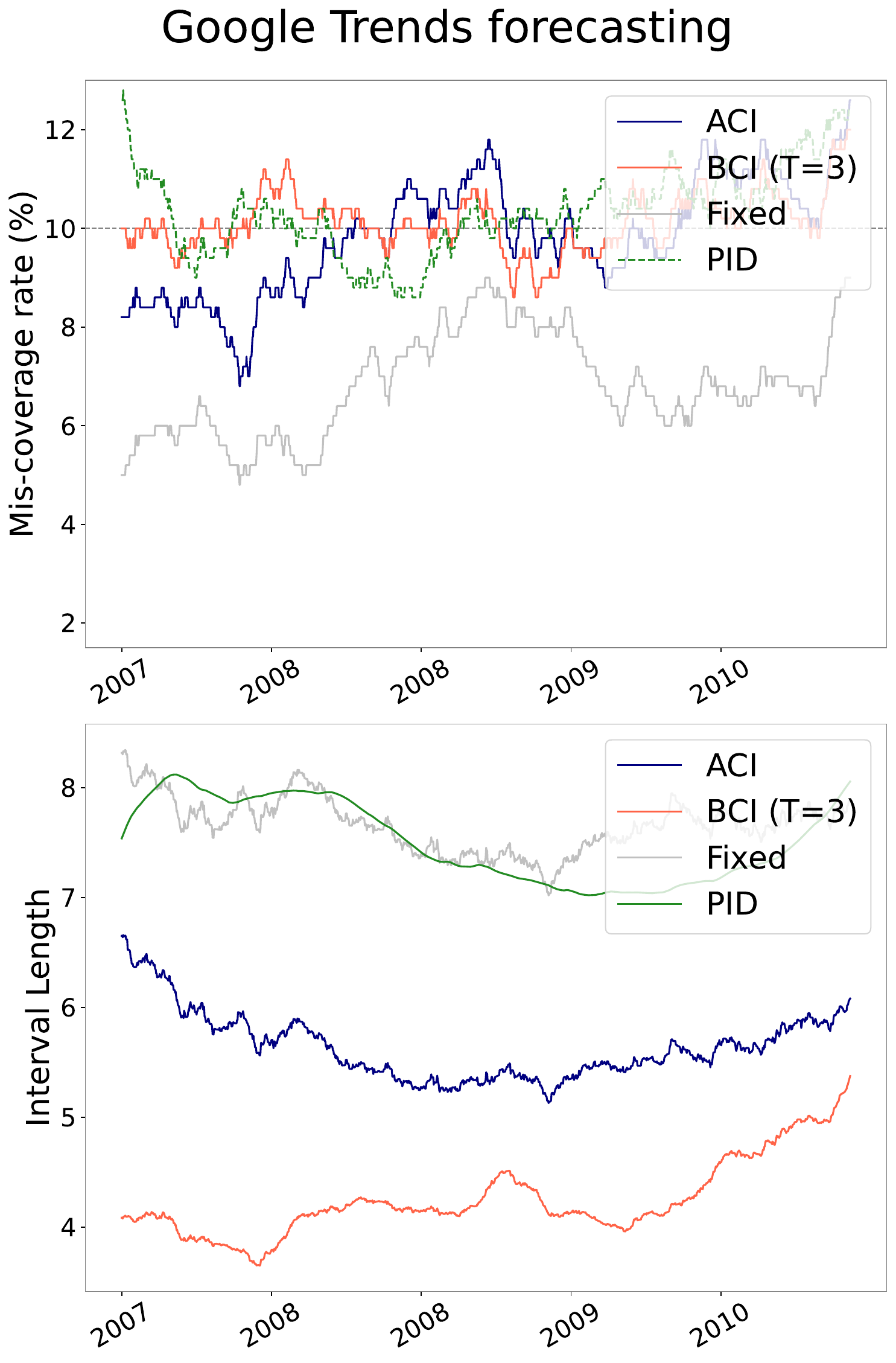}
        \caption{Google trend}
        \label{fig:google-trend-loose}
    \end{subfigure}
    \caption{Same as Figure \ref{fig:exp-main-tight}, except that the stepsize $\gamma$ for ACI is $0.008$ for looser control.}    
    \label{fig:exp-main-loose}
\end{figure}

\subsection{Performance evaluation}
\label{sec:exp-comparsion-setup}

We turn to discussing how we evaluate the performance of BCI. In particular, we point out caveats when comparing the performance of online prediction intervals and propose a fair evaluation.

\textbf{Evaluation metrics.} We evaluate the performance of online prediction intervals using two metrics: average miscoverage rate and average interval length. Following \citet{gibbs2021adaptive}, we measure the local average of both measures over a moving window of size $500$: 
\[
\begin{bmatrix}
\mathrm{LocalMiscov}_t\\
\mathrm{LocalLength}_t
\end{bmatrix}
= \frac{1}{500}\sum_{s=t-250}^{t+250} \begin{bmatrix}
\err_s\\
|C_s(1-\alpha_s)|
\end{bmatrix}.
\]

We report the standard deviation of the local miscoverage rate ($\text{LocalMiscov}_t$) in Tables \ref{tbl:exp-tight} and \ref{tbl:exp-loose}. 

\textbf{An approach for fair comparison.} For ACI, BCI, and conformal PID, the stepsize $\gamma$ can trade off the tightness of the coverage and the rate of change of the interval length. Typically, a smaller stepsize means that the local miscoverage rate will exhibit larger excursions away from the target, hence looser control, and that the length of the prediction intervals will be smoother due to the smaller increment in $\alpha_t$'s. We illustrate this tradeoff in Figure \ref{fig:exp-main-loose}. Ideally, we want to compare the average interval lengths when both methods achieve similar levels of coverage control.  
To make the comparison fair, we first choose a set of ACI parameters ($\gamma=0.1$ for tight control and $\gamma=0.008$ for loose control). For each choice of $\gamma$ for ACI, we tune BCI and PID; see Appendix \ref{sec:tuning}. 
% Note that this tuning method is used to make fair comparisons among the three methods, since shorter intervals can be obtained at the cost of higher miscoverage. 
While the step size parameters are chosen post-hoc, length comparisons can be misleading without matching the tightness of miscoverate control due to the miscoverate-length tradeoff.

\subsection{Empirical results}
\label{sec:exp-results}
For each forecasting problem, we apply ACI with $\gamma = 0.1$ (tight coverage control) and $\gamma = 0.008$ (loose coverage control). We run BCI and conformal PID with stepsizes chosen through the grid search method discussed in the last subsection. We also include the naive benchmark with $\alpha_t = \bar{\alpha}$. 

The time series of $\mathrm{LocalMiscov}_t$ and $\mathrm{LocalLength}_t$ are plotted in Figure \ref{fig:exp-main-tight} and Figure \ref{fig:exp-main-loose} for a subset of experiments. 
All three methods achieve comparable levels of coverage control across every combination of dataset and $\gamma$ (Tables \ref{tbl:exp-tight}, \ref{tbl:exp-loose}). BCI achieves the shortest average interval length in most cases; ACI is shortest on volatility forecasting specifically, where the underlying GARCH model already produces nearly well-calibrated nominal intervals. Conformal PID does not achieve the shortest interval length in any of the tabulated combinations we test (Tables \ref{tbl:exp-tight}, \ref{tbl:exp-loose}); results on the additional datasets in Appendix \ref{sec:more-exp} are qualitatively consistent.

\textbf{Uninformative infinite-length intervals.} One issue pointed out by \citet{angelopoulos2023conformal} is that ACI may generate infinite-length intervals. From Table \ref{tbl:exp-tight}, we observe that ACI generates a moderate fraction of infinite-length prediction intervals under tight coverage control. In contrast, BCI avoids infinite intervals entirely at our main horizon $T=3$ and generally tends to produce shorter intervals (see the horizon-sensitivity discussion below for how this changes at larger $T$). Under loose control, neither ACI nor BCI generate any uninformative intervals.

\textbf{Quality of nominal prediction intervals.} From Figure \ref{fig:exp-main-tight} and Figure \ref{fig:exp-main-loose}, we can see that BCI generally outperforms ACI. For return forecasting and Google trend forecasting, the gain is more prominent than for volatility forecasting. 
Intuitively, if the nominal prediction intervals are well-calibrated, BCI would not gain much from using the multi-step ahead prediction intervals. We confirm this intuition through the expected calibration curves (ECC) for each of the datasets; see Appendix \ref{subsec:ECC}. 

\textbf{Sensitivity to Prediction Horizon.} Finally, we investigate the sensitivity of BCI to the choice of the prediction horizon $T$. While our main results utilize $T=3$, we also evaluated the framework across $T \in \{1, 3, 5, 7, 9\}$ and found that larger $T$ yields only marginal further reduction in interval length but a substantially higher fraction of unbounded intervals (up to 5--6\% for Google Trends and Volatility-AMD under tight control by $T=7$--$9$, versus under 1\% at $T=3$), making $T=3$ a reasonable balance between horizon length and interval boundedness.

\section{Conclusion}
 We propose Bellman Conformal Inference (BCI) as an extension of ACI that calibrates nominal prediction intervals produced by any forecasting algorithms with coverage guarantees. BCI leverages multi-step ahead prediction intervals and applies Model Predictive Control (MPC) techniques to explicitly optimize the interval lengths. When the nominal prediction intervals are poorly calibrated, BCI improves substantially upon ACI in terms of the average interval lengths; otherwise, BCI has comparable performance to ACI.

% \section{Acknowledgement}
% Zitong Yang would like to acknowledge the support of Albion Walter Hewlett Stanford Graduate Fellowship.
% Lihua Lei is grateful for the support of  National Science Foundation grant DMS-2338464.
% E.J.C. was supported by the Office of Naval Research grant N00014-20-1-2157, the National Science Foundation grant DMS-2032014, the Simons Foundation under award 814641, and the ARO grant 2003514594.

\bibliographystyle{plainnat}
\bibliography{reference}

\begin{thebibliography}{46}
\providecommand{\natexlab}[1]{#1}
\providecommand{\url}[1]{\texttt{#1}}
\expandafter\ifx\csname urlstyle\endcsname\relax
  \providecommand{\doi}[1]{doi: #1}\else
  \providecommand{\doi}{doi: \begingroup \urlstyle{rm}\Url}\fi

\bibitem[Angelopoulos and Bates(2021)]{angelopoulos2021gentle}
Anastasios~N Angelopoulos and Stephen Bates.
\newblock A gentle introduction to conformal prediction and distribution-free
  uncertainty quantification.
\newblock \emph{arXiv preprint arXiv:2107.07511}, 2021.

\bibitem[Angelopoulos et~al.(2021)Angelopoulos, Bates, Cand{\`e}s, Jordan, and
  Lei]{angelopoulos2021learn}
Anastasios~N Angelopoulos, Stephen Bates, Emmanuel~J Cand{\`e}s, Michael~I
  Jordan, and Lihua Lei.
\newblock Learn then test: Calibrating predictive algorithms to achieve risk
  control.
\newblock \emph{arXiv preprint arXiv:2110.01052}, 2021.

\bibitem[Angelopoulos et~al.(2022)Angelopoulos, Bates, Fisch, Lei, and
  Schuster]{angelopoulos2022conformal}
Anastasios~N Angelopoulos, Stephen Bates, Adam Fisch, Lihua Lei, and Tal
  Schuster.
\newblock Conformal risk control.
\newblock \emph{arXiv preprint arXiv:2208.02814}, 2022.

\bibitem[Angelopoulos et~al.(2023)Angelopoulos, Candes, and
  Tibshirani]{angelopoulos2023conformal}
Anastasios~N Angelopoulos, Emmanuel~J Candes, and Ryan~J Tibshirani.
\newblock Conformal {PID} control for time series prediction.
\newblock \emph{arXiv preprint arXiv:2307.16895}, 2023.

\bibitem[Barber et~al.(2023)Barber, Candes, Ramdas, and
  Tibshirani]{barber2023conformal}
Rina~Foygel Barber, Emmanuel~J Candes, Aaditya Ramdas, and Ryan~J Tibshirani.
\newblock Conformal prediction beyond exchangeability.
\newblock \emph{The Annals of Statistics}, 51\penalty0 (2):\penalty0 816--845,
  2023.

\bibitem[Bates et~al.(2021)Bates, Angelopoulos, Lei, Malik, and
  Jordan]{bates2021distribution}
Stephen Bates, Anastasios Angelopoulos, Lihua Lei, Jitendra Malik, and Michael
  Jordan.
\newblock Distribution-free, risk-controlling prediction sets.
\newblock \emph{Journal of the ACM (JACM)}, 68\penalty0 (6):\penalty0 1--34,
  2021.

\bibitem[Bertsekas(1976)]{dp}
Dimitri~P. Bertsekas.
\newblock \emph{Dynamic Programming and Stochastic Control}.
\newblock Academic Press, Inc., USA, 1976.
\newblock ISBN 0120932504.

\bibitem[Bollerslev(1986)]{bollerslev1986generalized}
Tim Bollerslev.
\newblock Generalized autoregressive conditional heteroskedasticity.
\newblock \emph{Journal of econometrics}, 31\penalty0 (3):\penalty0 307--327,
  1986.

\bibitem[Borrelli et~al.(2017)Borrelli, Bemporad, and Morari]{mpc}
Francesco Borrelli, Alberto Bemporad, and Manfred Morari.
\newblock \emph{Predictive Control for Linear and Hybrid Systems}.
\newblock Cambridge University Press, 2017.
\newblock \doi{10.1017/9781139061759}.

\bibitem[Box and Jenkins(1976)]{boxjen76}
George.E.P. Box and Gwilym~M. Jenkins.
\newblock \emph{Time Series Analysis: Forecasting and Control}.
\newblock Holden-Day, 1976.

\bibitem[Brown(1964)]{Brown1964SmoothingFA}
Robert~Goodell Brown.
\newblock Smoothing, forecasting and prediction of discrete time series.
\newblock In \emph{.}, 1964.
\newblock URL \url{https://api.semanticscholar.org/CorpusID:120260777}.

\bibitem[Cand{\`e}s et~al.(2023)Cand{\`e}s, Lei, and
  Ren]{candes2023conformalized}
Emmanuel Cand{\`e}s, Lihua Lei, and Zhimei Ren.
\newblock Conformalized survival analysis.
\newblock \emph{Journal of the Royal Statistical Society Series B: Statistical
  Methodology}, 85\penalty0 (1):\penalty0 24--45, 2023.

\bibitem[Chernozhukov et~al.(2018)Chernozhukov, W{\"u}thrich, and
  Yinchu]{chernozhukov2018exact}
Victor Chernozhukov, Kaspar W{\"u}thrich, and Zhu Yinchu.
\newblock Exact and robust conformal inference methods for predictive machine
  learning with dependent data.
\newblock In \emph{Conference On learning theory}, pages 732--749. PMLR, 2018.

\bibitem[Dietterich and Hostetler(2022)]{dietterich2022conformal}
Thomas~G Dietterich and Jesse Hostetler.
\newblock Conformal prediction intervals for markov decision process
  trajectories.
\newblock \emph{arXiv preprint arXiv:2206.04860}, 2022.

\bibitem[Engle(1982{\natexlab{a}})]{ar82}
Robert Engle.
\newblock Autoregressive conditional heteroscedasticity with estimates of the
  variance of united kingdom inflation.
\newblock \emph{Econometrica}, 50\penalty0 (4):\penalty0 987--1007,
  1982{\natexlab{a}}.
\newblock URL
  \url{https://EconPapers.repec.org/RePEc:ecm:emetrp:v:50:y:1982:i:4:p:987-1007}.

\bibitem[Engle(1982{\natexlab{b}})]{engle1982autoregressive}
Robert~F Engle.
\newblock Autoregressive conditional heteroscedasticity with estimates of the
  variance of united kingdom inflation.
\newblock \emph{Econometrica: Journal of the econometric society}, pages
  987--1007, 1982{\natexlab{b}}.

\bibitem[Fan and Yao(2003)]{fan2003nonlinear}
Jianqing Fan and Qiwei Yao.
\newblock \emph{Nonlinear time series: nonparametric and parametric methods},
  volume~20.
\newblock Springer, 2003.

\bibitem[Feldman et~al.(2023)Feldman, Ringel, Bates, and Romano]{rolrc}
Shai Feldman, Liran Ringel, Stephen Bates, and Yaniv Romano.
\newblock Achieving risk control in online learning settings, 2023.

\bibitem[Gibbs and Cand\`{e}s(2021)]{gibbs2021adaptive}
Isaac Gibbs and Emmanuel Cand\`{e}s.
\newblock Adaptive conformal inference under distribution shift.
\newblock In A.~Beygelzimer, Y.~Dauphin, P.~Liang, and J.~Wortman Vaughan,
  editors, \emph{Advances in Neural Information Processing Systems}, 2021.
\newblock URL \url{https://openreview.net/forum?id=6vaActvpcp3}.

\bibitem[Gibbs and Candès(2022)]{gibbs2022conformal}
Isaac Gibbs and Emmanuel Candès.
\newblock Conformal inference for online prediction with arbitrary distribution
  shifts, 2022.

\bibitem[Gupta et~al.(2022)Gupta, Kuchibhotla, and Ramdas]{gupta2022nested}
Chirag Gupta, Arun~K Kuchibhotla, and Aaditya Ramdas.
\newblock Nested conformal prediction and quantile out-of-bag ensemble methods.
\newblock \emph{Pattern Recognition}, 127:\penalty0 108496, 2022.

\bibitem[Herzen et~al.(2022)Herzen, LAossig, Piazzetta, Neuer, Tafti, Raille,
  Pottelbergh, Pasieka, Skrodzki, Huguenin, Dumonal, KoAcisz, Bader, Gusset,
  Benheddi, Williamson, Kosinski, Petrik, and Grosch]{darts}
Julien Herzen, Francesco LAossig, Samuele~Giuliano Piazzetta, Thomas Neuer,
  LAoo Tafti, Guillaume Raille, Tomas~Van Pottelbergh, Marek Pasieka, Andrzej
  Skrodzki, Nicolas Huguenin, Maxime Dumonal, Jan KoAcisz, Dennis Bader,
  FrACdAcrick Gusset, Mounir Benheddi, Camila Williamson, Michal Kosinski,
  Matej Petrik, and GaAl Grosch.
\newblock Darts: User-friendly modern machine learning for time series.
\newblock \emph{Journal of Machine Learning Research}, 23\penalty0
  (124):\penalty0 1--6, 2022.
\newblock URL \url{http://jmlr.org/papers/v23/21-1177.html}.

\bibitem[Hochreiter and Schmidhuber(1997)]{lstm}
Sepp Hochreiter and J\"{u}rgen Schmidhuber.
\newblock Long short-term memory.
\newblock \emph{Neural Comput.}, 9\penalty0 (8):\penalty0 1735–1780, nov
  1997.
\newblock ISSN 0899-7667.
\newblock \doi{10.1162/neco.1997.9.8.1735}.
\newblock URL \url{https://doi.org/10.1162/neco.1997.9.8.1735}.

\bibitem[Jorda(2005)]{oscar05}
Oscar Jorda.
\newblock Estimation and inference of impulse responses by local projections.
\newblock \emph{American Economic Review}, 95\penalty0 (1):\penalty0 161--182,
  March 2005.
\newblock \doi{10.1257/0002828053828518}.
\newblock URL
  \url{https://www.aeaweb.org/articles?id=10.1257/0002828053828518}.

\bibitem[Kingma and Welling(2014)]{vae}
Diederik~P. Kingma and Max Welling.
\newblock {Auto-Encoding Variational Bayes}.
\newblock In \emph{2nd International Conference on Learning Representations,
  {ICLR} 2014, Banff, AB, Canada, April 14-16, 2014, Conference Track
  Proceedings}, 2014.

\bibitem[Lei et~al.(2015)Lei, Rinaldo, and Wasserman]{lei2013conformal}
Jing Lei, Alessandro Rinaldo, and Larry Wasserman.
\newblock A conformal prediction approach to explore functional data.
\newblock \emph{Annals of Mathematics and Artificial Intelligence},
  74:\penalty0 29--43, 2015.
\newblock \doi{10.1007/s10472-013-9366-6}.

\bibitem[Lei et~al.(2018)Lei, G’Sell, Rinaldo, Tibshirani, and
  Wasserman]{lei2018distribution}
Jing Lei, Max G’Sell, Alessandro Rinaldo, Ryan~J Tibshirani, and Larry
  Wasserman.
\newblock Distribution-free predictive inference for regression.
\newblock \emph{Journal of the American Statistical Association}, 113\penalty0
  (523):\penalty0 1094--1111, 2018.

\bibitem[Lei and Cand{\`e}s(2021)]{lei2021conformal}
Lihua Lei and Emmanuel~J Cand{\`e}s.
\newblock Conformal inference of counterfactuals and individual treatment
  effects.
\newblock \emph{Journal of the Royal Statistical Society Series B: Statistical
  Methodology}, 83\penalty0 (5):\penalty0 911--938, 2021.

\bibitem[Lekeufack et~al.(2023)Lekeufack, Angelopoulos, Bajcsy, Jordan, and
  Malik]{lekeufack2023conformal}
Jordan Lekeufack, Anastasios~N. Angelopoulos, Andrea Bajcsy, Michael~I. Jordan,
  and Jitendra Malik.
\newblock Conformal decision theory: Safe autonomous decisions from imperfect
  predictions, 2023.

\bibitem[Makridakis et~al.(2018)Makridakis, Spiliotis, and
  Assimakopoulos]{makridakis2018statistical}
Spyros Makridakis, Evangelos Spiliotis, and Vassilios Assimakopoulos.
\newblock Statistical and machine learning forecasting methods: Concerns and
  ways forward.
\newblock \emph{PloS one}, 13\penalty0 (3):\penalty0 e0194889, 2018.

\bibitem[Oliveira et~al.(2022)Oliveira, Orenstein, Ramos, and
  Romano]{oliveira2022split}
Roberto~I Oliveira, Paulo Orenstein, Thiago Ramos, and Jo{\~a}o~Vitor Romano.
\newblock Split conformal prediction for dependent data.
\newblock \emph{arXiv preprint arXiv:2203.15885}, 2022.

\bibitem[Papadopoulos et~al.(2002)Papadopoulos, Proedrou, Vovk, and
  Gammerman]{papadopoulos2002inductive}
Harris Papadopoulos, Kostas Proedrou, Volodya Vovk, and Alex Gammerman.
\newblock Inductive confidence machines for regression.
\newblock In \emph{Machine Learning: ECML 2002: 13th European Conference on
  Machine Learning Helsinki, Finland, August 19--23, 2002 Proceedings 13},
  pages 345--356. Springer, 2002.

\bibitem[Politis and Wu(2023)]{politis2023multi}
Dimitris~N Politis and Kejin Wu.
\newblock Multi-step-ahead prediction intervals for nonparametric
  autoregressions via bootstrap: Consistency, debiasing, and pertinence.
\newblock \emph{Stats}, 6\penalty0 (3):\penalty0 839--867, 2023.

\bibitem[Salinas et~al.(2020)Salinas, Flunkert, Gasthaus, and
  Januschowski]{deepar}
David Salinas, Valentin Flunkert, Jan Gasthaus, and Tim Januschowski.
\newblock Deepar: Probabilistic forecasting with autoregressive recurrent
  networks.
\newblock \emph{International Journal of Forecasting}, 36\penalty0
  (3):\penalty0 1181--1191, 2020.
\newblock ISSN 0169-2070.
\newblock \doi{https://doi.org/10.1016/j.ijforecast.2019.07.001}.
\newblock URL
  \url{https://www.sciencedirect.com/science/article/pii/S0169207019301888}.

\bibitem[Saunders et~al.(1999)Saunders, Gammerman, and
  Vovk]{saunders1999transduction}
Craig Saunders, Alexander Gammerman, and Volodya Vovk.
\newblock Transduction with confidence and credibility.
\newblock In \emph{Proceedings of the Sixteenth International Joint Conference
  on Artificial Intelligence}, IJCAI '99, page 722–726, San Francisco, CA,
  USA, 1999. Morgan Kaufmann Publishers Inc.
\newblock ISBN 1558606130.

\bibitem[Stankeviciute et~al.(2021)Stankeviciute, M~Alaa, and van~der
  Schaar]{stankeviciute2021conformal}
Kamile Stankeviciute, Ahmed M~Alaa, and Mihaela van~der Schaar.
\newblock Conformal time-series forecasting.
\newblock \emph{Advances in neural information processing systems},
  34:\penalty0 6216--6228, 2021.

\bibitem[Stock and Watson(2010)]{stowat10}
James Stock and Mark Watson.
\newblock \emph{Dynamic Factor Models}, chapter~., page~.
\newblock Oxford University Press, Oxford, 2010.
\newblock URL
  \url{http://www.economics.harvard.edu/faculty/stock/files/dfm_oup_4.pdf}.

\bibitem[Sun and Yu(2023)]{sun2023copula}
Sophia Sun and Rose Yu.
\newblock Copula conformal prediction for multi-step time series forecasting,
  2023.

\bibitem[Taylor and Letham(2018)]{taylor2018forecasting}
Sean~J Taylor and Benjamin Letham.
\newblock Forecasting at scale.
\newblock \emph{The American Statistician}, 72\penalty0 (1):\penalty0 37--45,
  2018.

\bibitem[Tibshirani et~al.(2019)Tibshirani, Foygel~Barber, Candes, and
  Ramdas]{tibshirani2019conformal}
Ryan~J Tibshirani, Rina Foygel~Barber, Emmanuel Candes, and Aaditya Ramdas.
\newblock Conformal prediction under covariate shift.
\newblock \emph{Advances in neural information processing systems}, 32, 2019.

\bibitem[Vaswani et~al.(2017)Vaswani, Shazeer, Parmar, Uszkoreit, Jones, Gomez,
  Kaiser, and Polosukhin]{transformer}
Ashish Vaswani, Noam Shazeer, Niki Parmar, Jakob Uszkoreit, Llion Jones,
  Aidan~N Gomez, \L~ukasz Kaiser, and Illia Polosukhin.
\newblock Attention is all you need.
\newblock In I.~Guyon, U.~Von Luxburg, S.~Bengio, H.~Wallach, R.~Fergus,
  S.~Vishwanathan, and R.~Garnett, editors, \emph{Advances in Neural
  Information Processing Systems}, volume~30. Curran Associates, Inc., 2017.
\newblock URL
  \url{https://proceedings.neurips.cc/paper_files/paper/2017/file/3f5ee243547dee91fbd053c1c4a845aa-Paper.pdf}.

\bibitem[Vovk et~al.(2005)Vovk, Gammerman, and Shafer]{vovk05}
Vladimir Vovk, Alex Gammerman, and Glenn Shafer.
\newblock \emph{Algorithmic Learning in a Random World}.
\newblock Springer-Verlag, Berlin, Heidelberg, 2005.
\newblock ISBN 0387001522.

\bibitem[West and Harrison(2006)]{west2006bayesian}
Mike West and Jeff Harrison.
\newblock \emph{Bayesian forecasting and dynamic models}.
\newblock Springer Science \& Business Media, 2006.

\bibitem[Xu and Xie(2021)]{xu2021conformal}
Chen Xu and Yao Xie.
\newblock Conformal prediction interval for dynamic time-series.
\newblock In \emph{International Conference on Machine Learning}, pages
  11559--11569. PMLR, 2021.

\bibitem[Xu and Xie(2023)]{xu2023sequential}
Chen Xu and Yao Xie.
\newblock Sequential predictive conformal inference for time series.
\newblock In \emph{International Conference on Machine Learning}, pages
  38707--38727. PMLR, 2023.

\bibitem[Zaffran et~al.(2022)Zaffran, F{\'e}ron, Goude, Josse, and
  Dieuleveut]{zaffran2022adaptive}
Margaux Zaffran, Olivier F{\'e}ron, Yannig Goude, Julie Josse, and Aymeric
  Dieuleveut.
\newblock Adaptive conformal predictions for time series.
\newblock In \emph{International Conference on Machine Learning}, pages
  25834--25866. PMLR, 2022.

\end{thebibliography}

%%%%%%%%%%%%%%%%%%%%%%%%%%%%%%%%%%%%%%%%%%%%%%%%%%%%%%%%%%%%%%%%%%%%%%%%%%%%%%%
%%%%%%%%%%%%%%%%%%%%%%%%%%%%%%%%%%%%%%%%%%%%%%%%%%%%%%%%%%%%%%%%%%%%%%%%%%%%%%%
% APPENDIX
%%%%%%%%%%%%%%%%%%%%%%%%%%%%%%%%%%%%%%%%%%%%%%%%%%%%%%%%%%%%%%%%%%%%%%%%%%%%%%%
%%%%%%%%%%%%%%%%%%%%%%%%%%%%%%%%%%%%%%%%%%%%%%%%%%%%%%%%%%%%%%%%%%%%%%%%%%%%%%%
\newpage
\appendix
\onecolumn

\section{Additional experimental details}
\label{sec:more-exp}
%%%% RTFC %%%%
\subsection{When does BCI improve most upon ACI? }\label{subsec:ECC}

\begin{figure}[ht]
    \centering
    \begin{subfigure}[b]{0.32\textwidth}
        \centering
        \includegraphics[width=\textwidth, trim=0 0 0 15pt, clip]{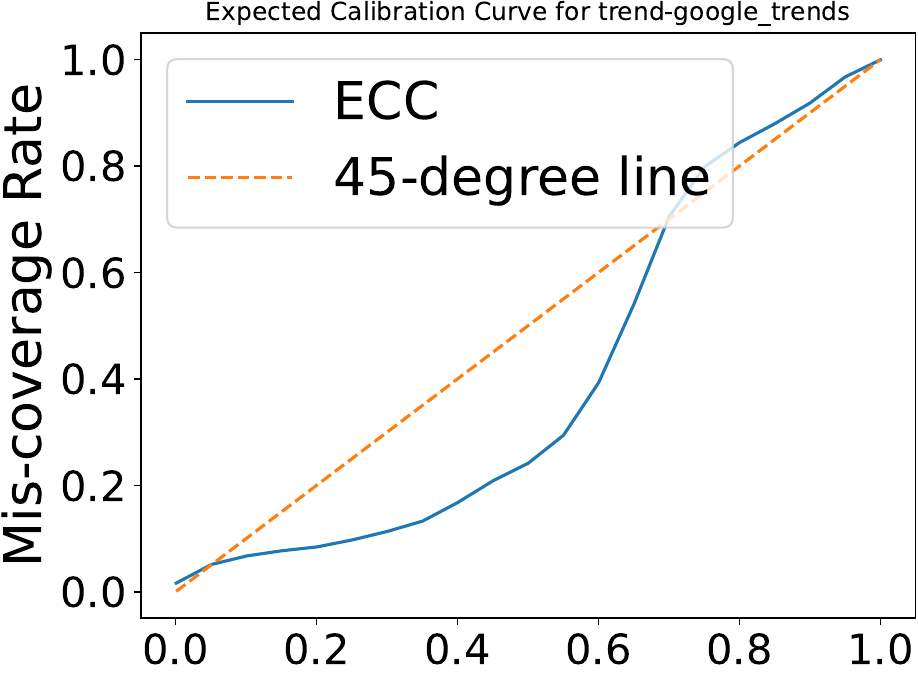}
        \caption{Google trend forecasting}
    \end{subfigure}
    \hfill
    \begin{subfigure}[b]{0.32\textwidth}
        \centering
        \includegraphics[width=\textwidth, trim=0 0 0 15pt, clip]{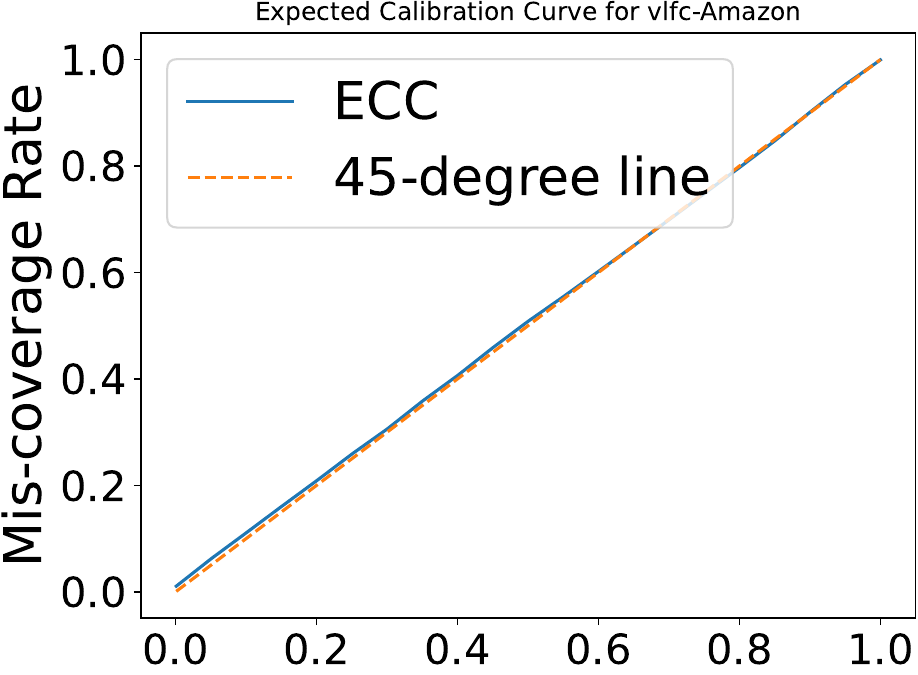}
        \caption{Volatility forecasting}
    \end{subfigure}
    \hfill
    \begin{subfigure}[b]{0.32\textwidth}
        \centering
        \includegraphics[width=\textwidth, trim=0 0 0 15pt, clip]{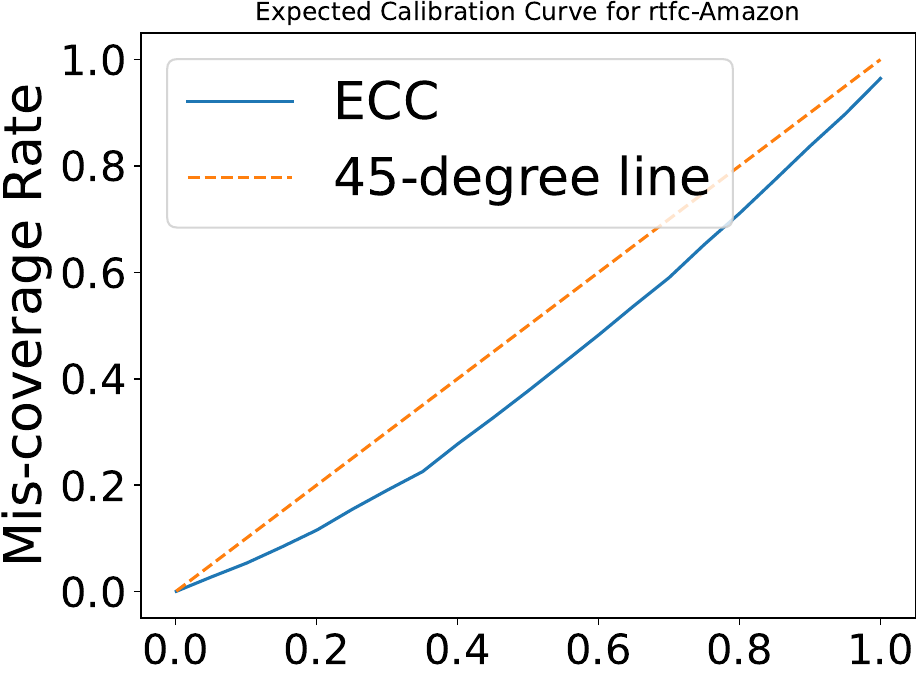}
        \caption{Return forecasting}
    \end{subfigure}
    \caption{Expected calibration curve of nominal prediction intervals for Google trend, volatility, and return forecasting.}
    \label{fig:ecc}
\end{figure} 

To examine the intuition that the amount of improvement of BCI over ACI depends on calibration of nominal prediction intervals, we define the expected calibration curve for nominal prediction intervals $C_t(\cdot)$ as follows:
\vspace{-0.1cm
}
\[
\ecc(\alpha) = \frac{1}{T}\sum_{t=1}^{T} 1\{{Y_t\notin C_t(1-\alpha)}\}.
\]
% In words, $\ecc(\alpha)$ is the average miscoverage rate when we fix a constant $\alpha_t$ as $\alpha_t = \overline{\alpha}$. 
If $\ecc(\alpha)\approx\alpha$ for all $\alpha\in (0, 1)$, it means that the nominal prediction intervals are already well-calibrated.

% \begin{wrapfigure}{r}{0.5\textwidth} % "r" for right side, and width of the wrap figure
% \centering
% \subfigure[Google trend forecasting\label{fig:trend-ece}]{\includegraphics[width=0.24\textwidth]{figures/trend-deeplearning-ece.pdf}}
% \subfigure[Volatility forecasting\label{fig:vlfc-Amazon-ece}]{\includegraphics[width=0.24\textwidth]{figures/vlfc-Amazon-ece.pdf}}
% \caption{Expected calibration curve of nominal prediction intervals for Google trend forecasting and volatility forecasting.}
% \label{fig:ece}
% \end{wrapfigure}

From Figure \ref{fig:ecc}, we can see that the Google trend forecasting and return forecasting with LSTM neural network is too conservative for small $\alpha$ and too aggressive for large $\alpha$.
In contrast, the \texttt{GARCH(1, 1)} generates nearly calibrated prediction invervals, 
% which means that, on average, $C_t(1-\alpha)$ covers the true $Y_t$ $(1-\alpha)$ fraction of the time, 
leaving little possibility for BCI to alter the prediction intervals.
This supports our intuition that BCI improves upon ACI more substantially when the nominal prediction intervals are poorly calibrated.

\subsection{Additional experiments}
\begin{figure}[ht]
    \centering
    \begin{subfigure}[b]{0.48\textwidth}
        \centering
        \includegraphics[width=0.48\linewidth]{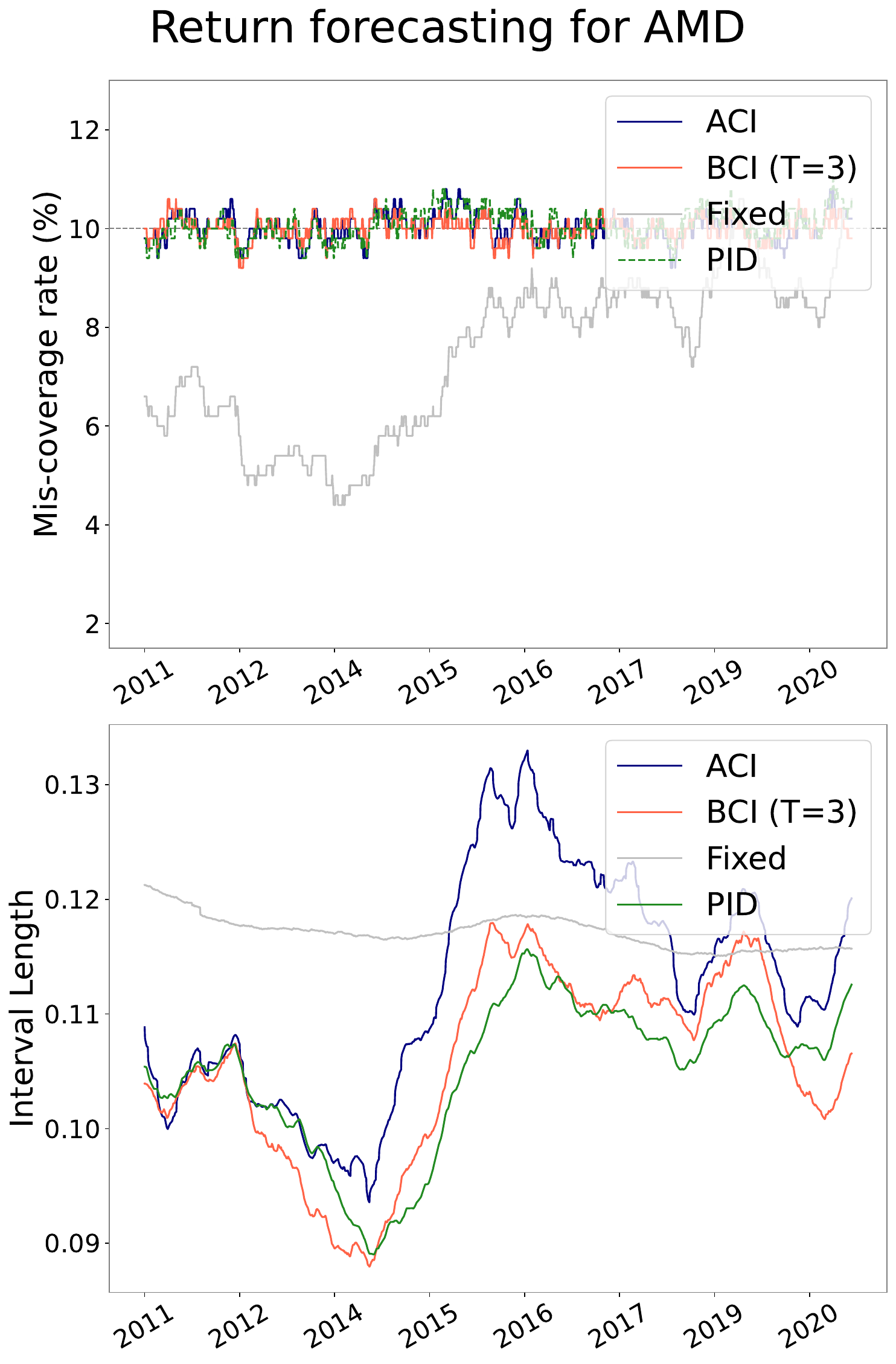}
        \includegraphics[width=0.48\linewidth]{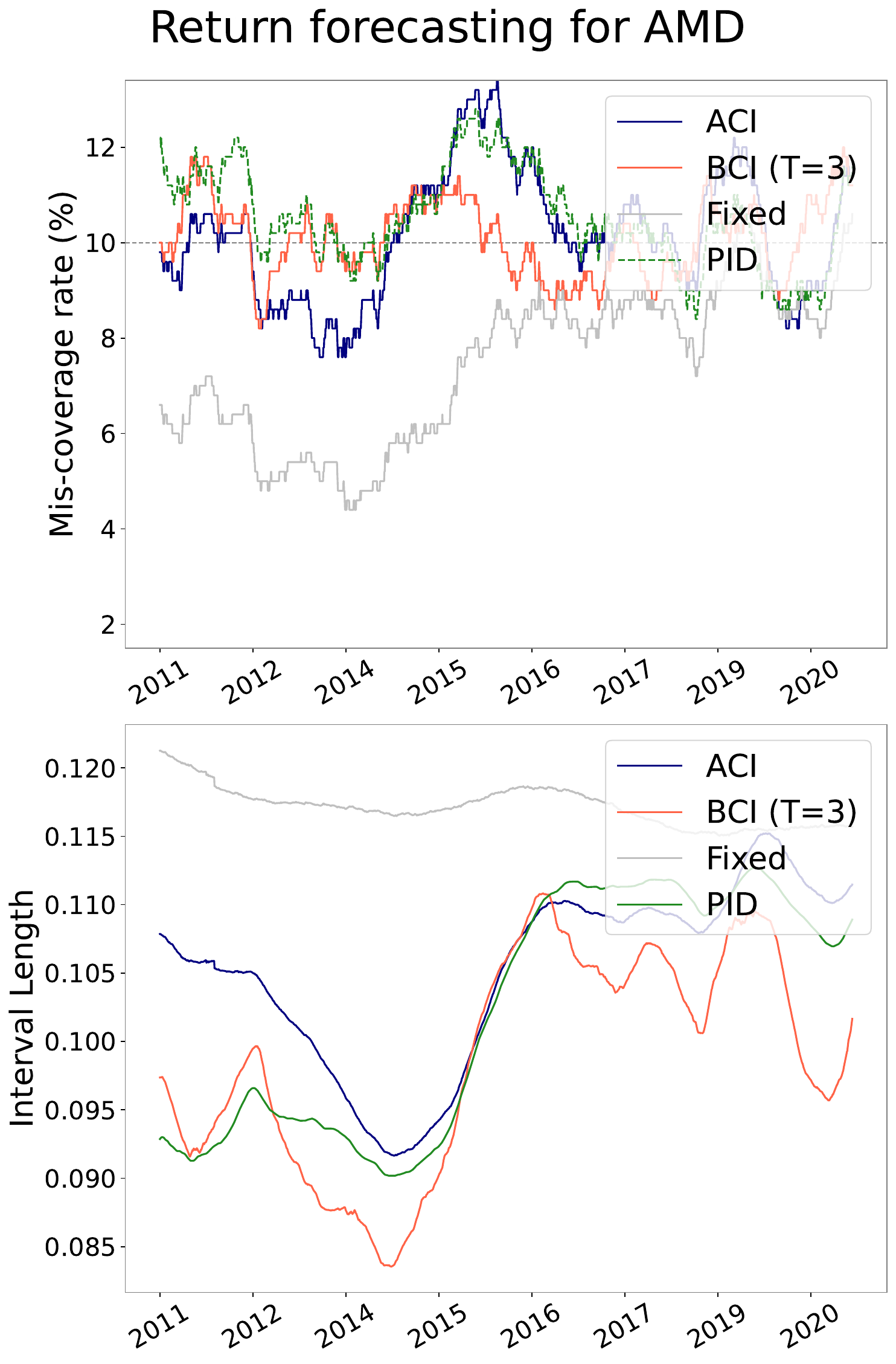}
        \caption{Amazon tight v.s. loose control}
        \label{fig:rtfc-AMD}
    \end{subfigure}
    \hfill
    \begin{subfigure}[b]{0.48\textwidth}
        \centering
        \includegraphics[width=0.48\linewidth]{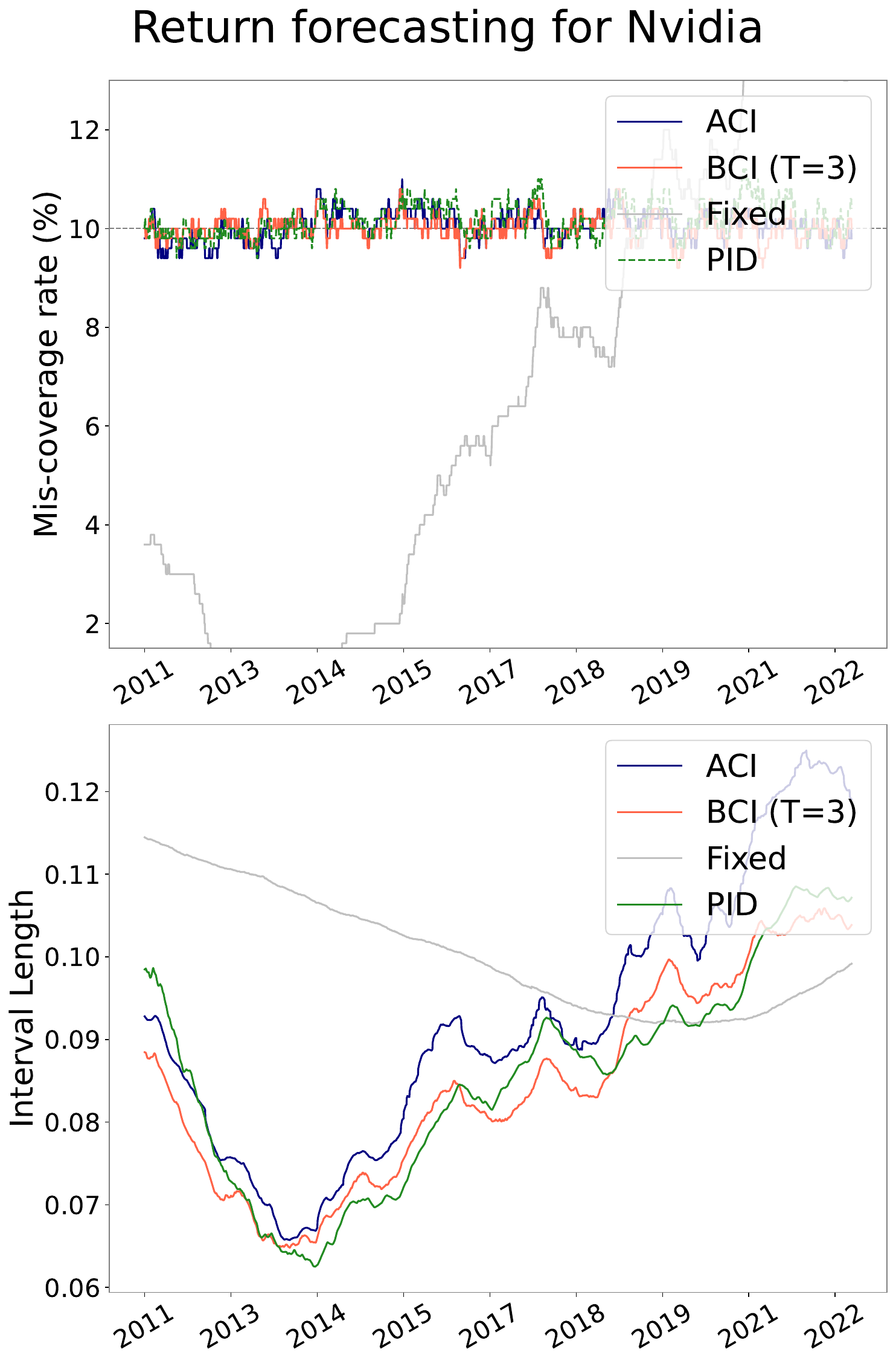}
        \includegraphics[width=0.48\linewidth]{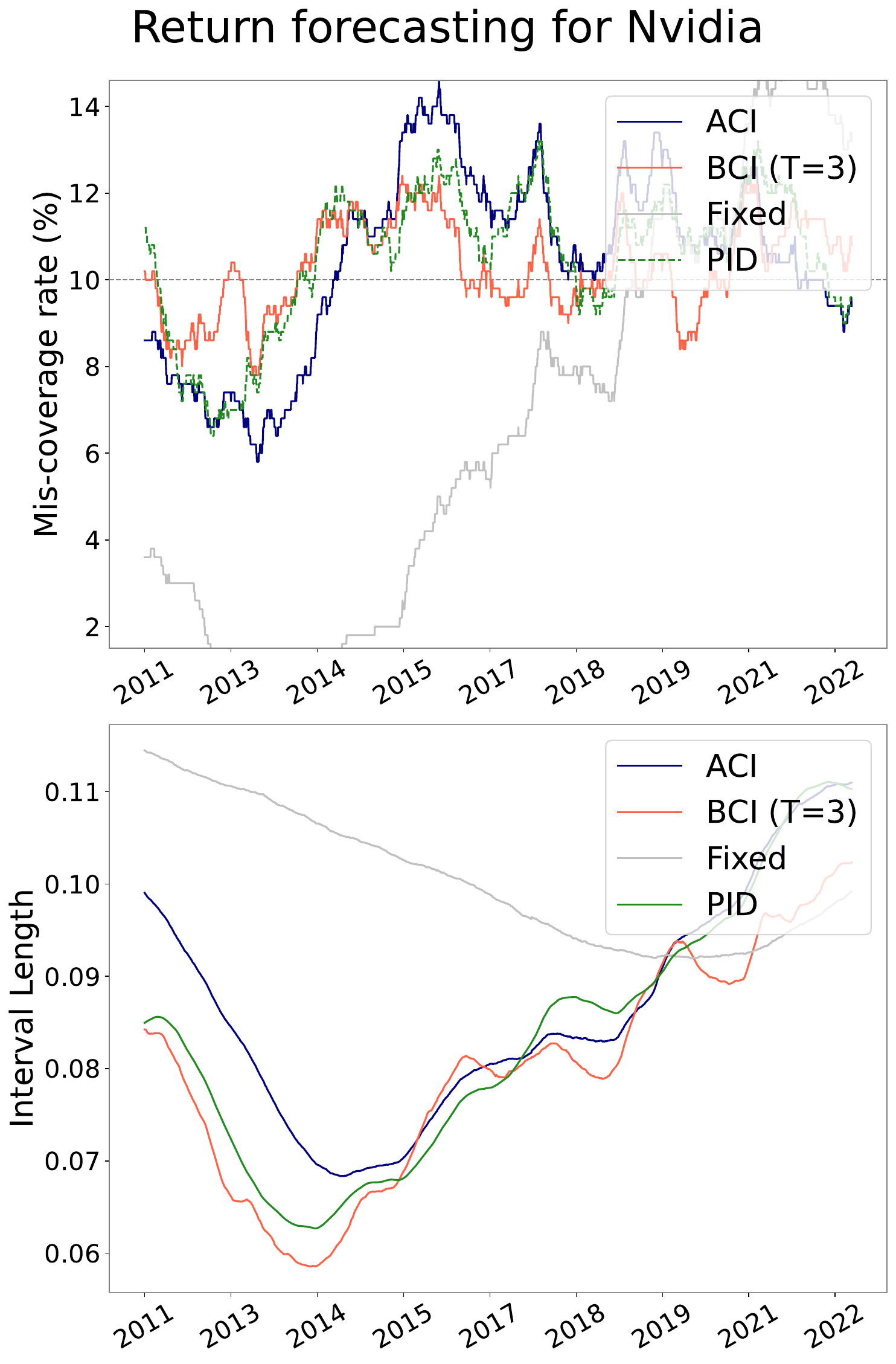}
        \caption{Nvidia tight v.s. loose control}
        \label{fig:rtfc-Nvidia}
    \end{subfigure}
    \caption{Additional return forecasting problems.}
    \label{fig:exp-more-rtfc}
\end{figure}

%%%% RTFC %%%%

%%%% VLFC %%%%
\begin{figure}[ht]
    \centering
    \begin{subfigure}[b]{0.48\textwidth}
        \centering
        \includegraphics[width=0.48\textwidth]{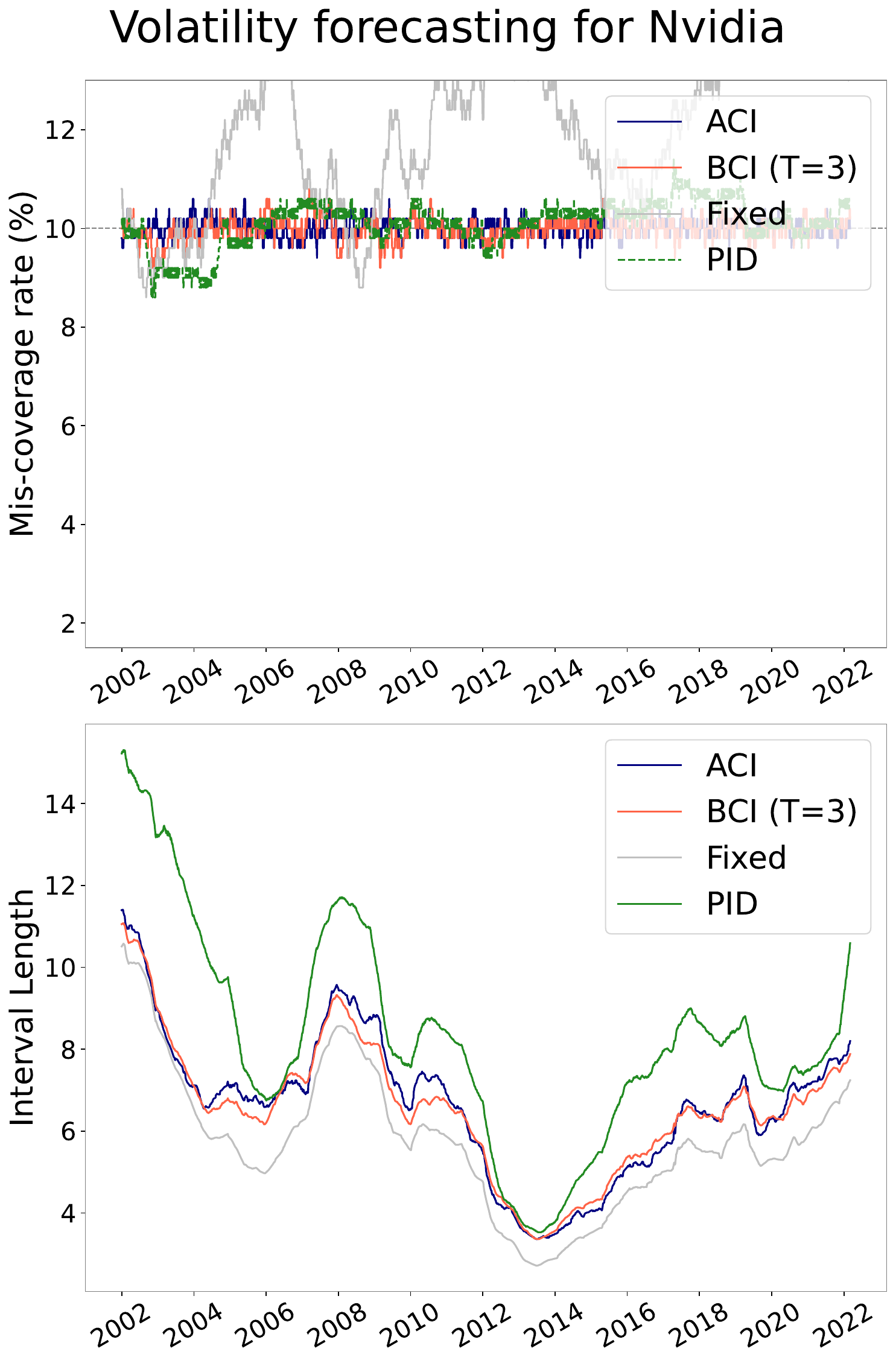}
        \includegraphics[width=0.48\textwidth]{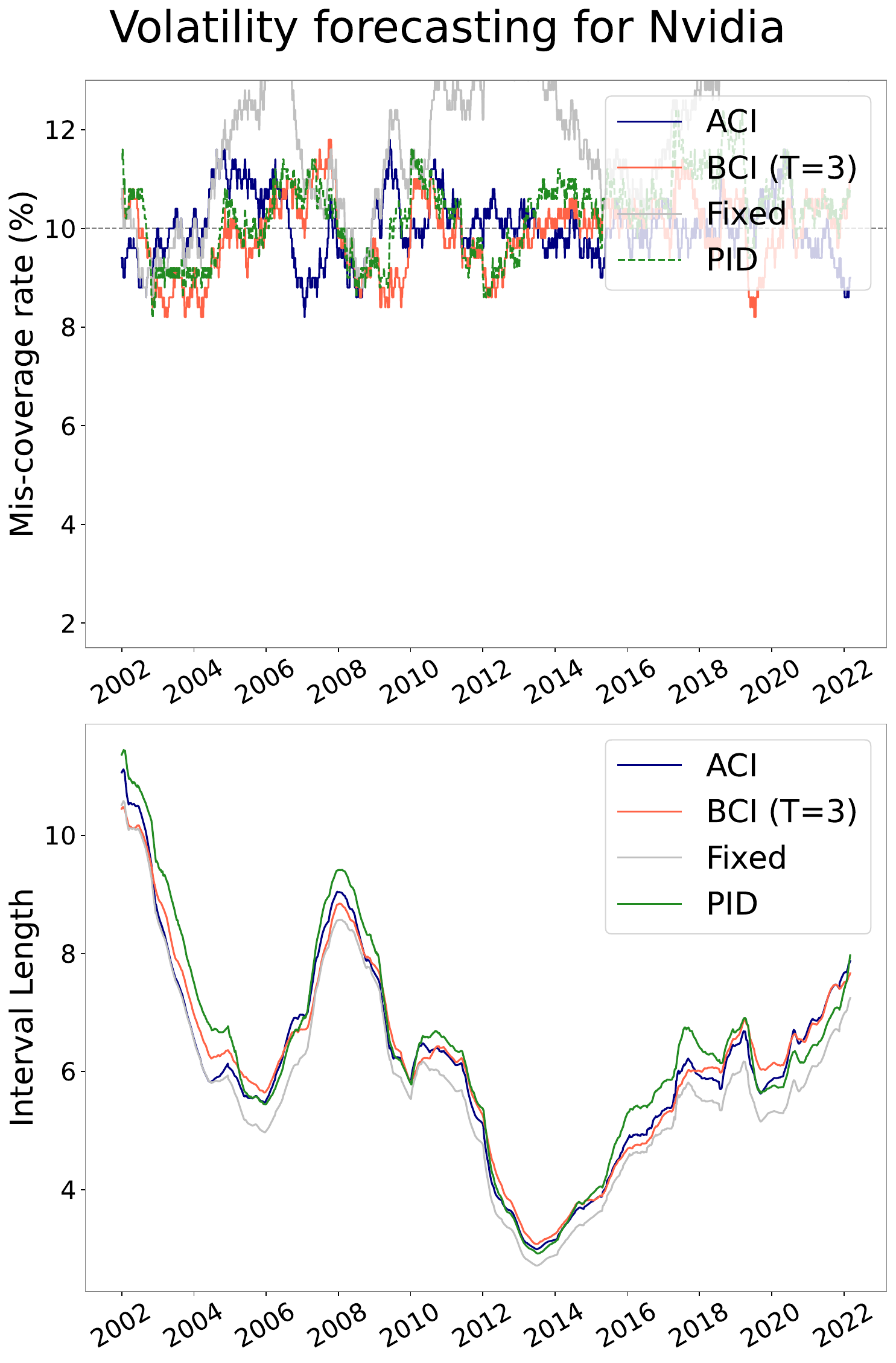}
        \caption{Nvidia tight v.s. loose control}
        \label{fig:vlfc-Nvidia}
    \end{subfigure}
    \hfill
    \begin{subfigure}[b]{0.48\textwidth}
        \centering
        \includegraphics[width=0.48\textwidth]{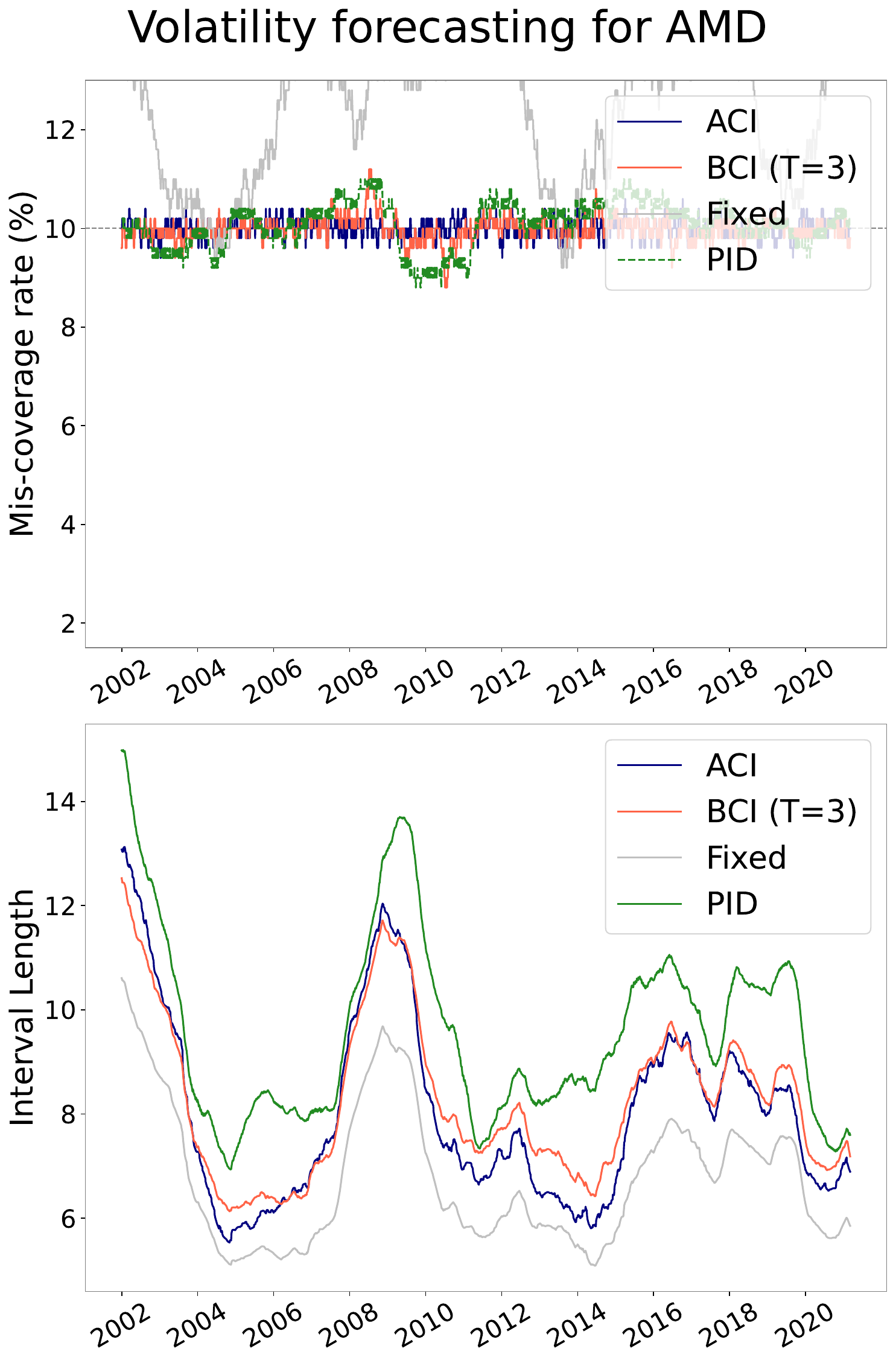}
        \includegraphics[width=0.48\textwidth]{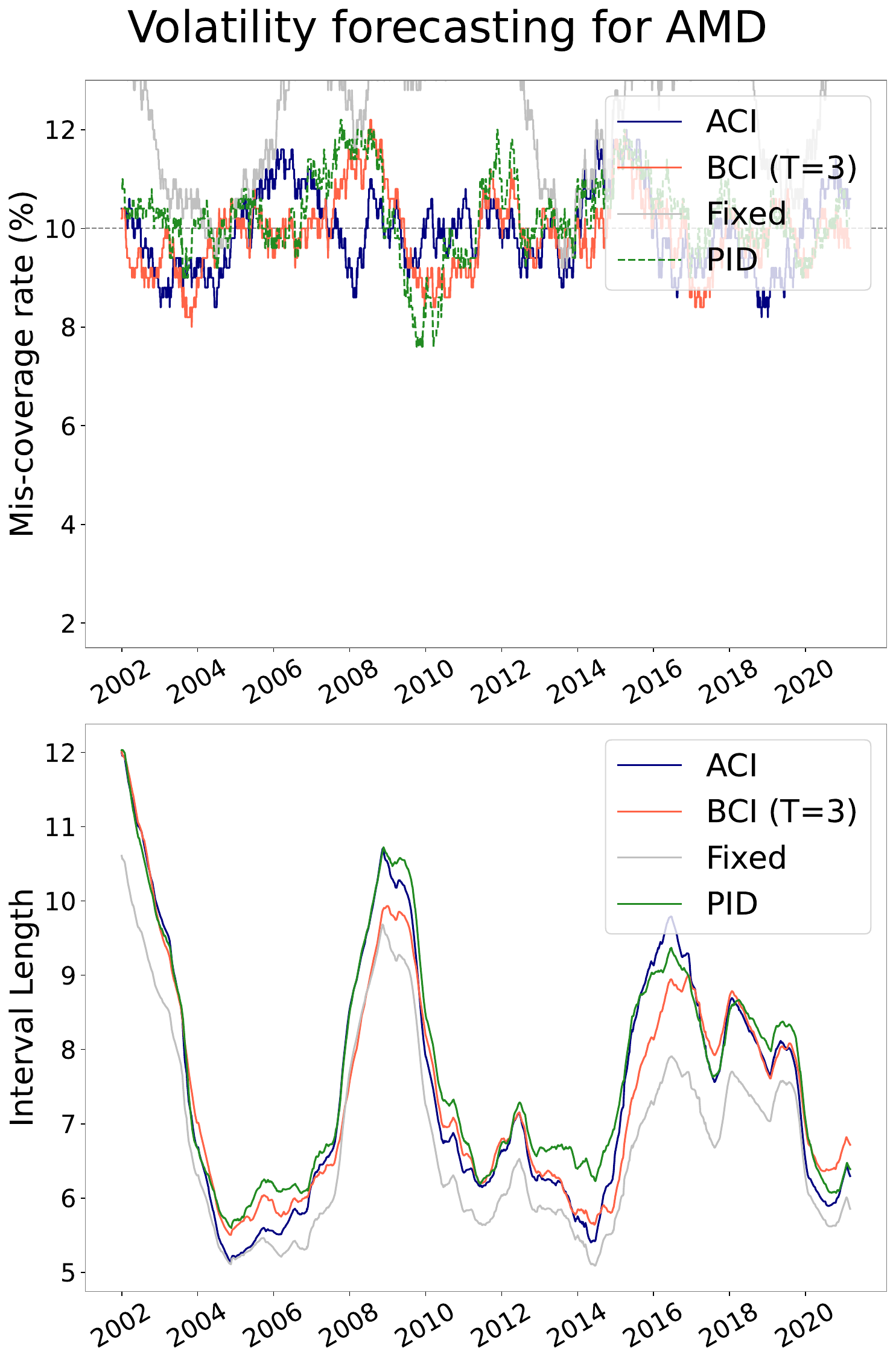}
        \caption{AMD tight v.s. loose control}
        \label{fig:vlfc-AMD}
    \end{subfigure}
    \caption{Additional volatility forecasting problems.}
    \label{fig:exp-more-vlfc}
\end{figure}

\begin{table*}[t]
\centering
\footnotesize
\setlength{\tabcolsep}{2pt}
\caption{Summary statistics for ACI with $\gamma = 0.1$ (tight control), BCI with $T=3$, and
PID. Miscoverage columns report mean $\pm$ std (\%); length excludes
$\infty$-sized intervals.}
\label{tbl:exp-tight}
\begin{tabular}{lccccccccc}
\toprule
& \multicolumn{3}{c}{Miscoverage (\%) $\pm$ Std.} & \multicolumn{3}{c}{Avg.\ length} & \multicolumn{3}{c}{Frac.\ $\infty$ (\%)} \\
\cmidrule(l){2-4} \cmidrule(l){5-7} \cmidrule(l){8-10}
Task/dataset & BCI & ACI & PID & BCI & ACI & PID & BCI & ACI & PID \\
\midrule
Return-AMD      & $9.95\pm0.256$  & $9.95\pm0.284$  & $10.06\pm0.441$ & 0.1051 & 0.1128 & 0.1059 & 0.00 & 3.37  & 0.00 \\
Return-Amazon   & $9.97\pm0.270$  & $9.99\pm0.298$  & $10.08\pm0.246$ & 0.0730 & 0.0787 & 0.0891 & 0.00 & 2.89  & 0.00 \\
Return-Nvidia   & $10.04\pm0.255$ & $10.04\pm0.287$ & $10.21\pm0.330$ & 0.0865 & 0.0928 & 0.0886 & 0.00 & 4.29  & 0.00 \\
\midrule
Volatility-AMD    & $9.97\pm0.235$ & $9.99\pm0.189$ & $10.09\pm0.498$ & 8.3956 & 8.2062 & 9.7662 & 0.00 & 10.75 & 0.00 \\
Volatility-Amazon & $9.97\pm0.168$ & $9.99\pm0.198$ & $10.31\pm0.288$ & 6.7463 & 6.4999 & 7.0544 & 0.00 & 8.85  & 0.00 \\
Volatility-Nvidia & $10.01\pm0.249$ & $10.01\pm0.190$ & $10.14\pm0.485$ & 6.7998 & 6.9466 & 8.7391 & 0.00 & 10.71 & 0.00 \\
\midrule
Google Trend & $10.36\pm0.545$ & $10.15\pm0.377$ & $10.92\pm0.695$ & 4.7895 & 6.3510 & 7.9425 & 0.00 & 2.81 & 0.00 \\
\bottomrule
\end{tabular}
\end{table*}

% ============================================================
% TABLE 2: LOOSE CONTROL (gamma = 0.008)
% ============================================================
\begin{table*}[t]
\centering
\footnotesize
\setlength{\tabcolsep}{2pt}
\caption{Same as Table~\ref{tbl:exp-tight}, except ACI uses $\gamma = 0.008$ (loose control).}
\label{tbl:exp-loose}
\begin{tabular}{lccccccccc}
\toprule
& \multicolumn{3}{c}{Miscoverage (\%) $\pm$ Std.} & \multicolumn{3}{c}{Avg.\ length} & \multicolumn{3}{c}{Frac.\ $\infty$ (\%)} \\
\cmidrule(l){2-4} \cmidrule(l){5-7} \cmidrule(l){8-10}
Task/dataset & BCI & ACI & PID & BCI & ACI & PID & BCI & ACI & PID \\
\midrule
Return-AMD      & $10.20\pm0.953$ & $10.13\pm1.284$ & $10.83\pm1.463$ & 0.0985 & 0.1057 & 0.1017 & 0.00 & 0.00 & 0.00 \\
Return-Amazon   & $9.53\pm1.067$  & $10.02\pm1.541$ & $11.47\pm2.171$ & 0.0712 & 0.0756 & 0.0720 & 0.00 & 0.00 & 0.00 \\
Return-Nvidia   & $10.28\pm1.390$ & $10.07\pm2.109$ & $10.62\pm2.212$ & 0.0822 & 0.0896 & 0.0854 & 0.00 & 0.00 & 0.00 \\
\midrule
Volatility-AMD    & $9.92\pm0.762$  & $10.03\pm0.803$ & $10.31\pm0.868$ & 7.6328 & 7.6393 & 7.8388 & 0.00 & 0.00 & 0.00 \\
Volatility-Amazon & $9.88\pm1.020$  & $9.93\pm0.559$  & $10.42\pm0.505$ & 6.0761 & 6.0013 & 6.3920 & 0.00 & 0.00 & 0.00 \\
Volatility-Nvidia & $10.01\pm0.829$ & $9.96\pm0.669$  & $10.37\pm0.885$ & 6.4198 & 6.4208 & 6.6602 & 0.00 & 0.00 & 0.00 \\
\midrule
Google Trend & $10.41\pm0.730$ & $10.10\pm1.593$ & $11.12\pm1.881$ & 4.4483 & 5.9183 & 7.6147 & 0.00 & 0.00 & 0.00 \\
\bottomrule
\end{tabular}
\end{table*}

\subsection{BCI and PID tuning for fair comparison}

\paragraph{Tuning procedure.} 
\label{sec:tuning}

A natural approach 
would be to tune each method to minimize average interval length, but this 
would unfairly favor methods that achieve coverage with narrower intervals 
by being less stable. Conversely, tuning purely on coverage 
stability would ignore interval efficiency entirely.

Instead, we use the 
standard deviation of the local miscoverage rate as our tuning criterion. 
We define $\hat{\sigma} = 
\mathrm{std}(\mathrm{LocalMiscov}_t)$ as our stability metric. For a fixed 
choice of ACI stepsize $\gamma$ (either $0.1$ for tight control or $0.008$ 
for loose control), we first run ACI and record its $\hat{\sigma}_{\text{ACI}}$. 
We then tune the hyperparameters of BCI and PID to match $\hat{\sigma}_{\text{ACI}}$ as closely as
possible, such that that miscoverage remains within
$\pm 0.5\%$ of the target miscoverage level $\bar{\alpha} = 0.1$. This ensures that ACI, BCI, and PID
have a comparable level of miscoverage control, making interval length comparison fair.

\paragraph{BCI Tuning.}
BCI requires two hyperparameters: $\lambda_{\max}$ and $c$, where $\gamma = c \cdot \lambda_{\max}$.

To find $\lambda_{\max}$, we exploit the theoretical property that the 
optimal nominal miscoverage rate $\beta^*(\lambda)$, defined as
\[
\beta^*(\lambda) = \arg\min_{\beta} \left[\bar{L}(\beta) + 
\lambda \cdot \max\!\left(\bar{m}(\beta) - \bar{\alpha},\, 0\right)\right],
\]
where $\bar{L}(\beta)$ and $\bar{m}(\beta)$ are the average interval length 
and miscoverage rate over the first 100 days, converges to a floor value 
as $\lambda \to \infty$ -- beyond which increasing $\lambda$ no longer 
changes the solution. We conduct a geometric grid search over 
$\lambda \in [10^{-4}, 10^6]$ and identify the smallest $\lambda^*$ at 
which $\beta^*(\lambda)$ has plateaued -- specifically, the first 
$\lambda^*$ such that $\beta^*(\lambda^*) \leq 1.01 \cdot 
\beta^*(\lambda_{\max}^{\text{grid}})$, where 
$\beta^*(\lambda_{\max}^{\text{grid}})$ is the value at the right boundary 
of the grid. We set $\lambda_{\max} = \lambda^*$ and 
$\lambda_{\text{init}} = 0.01 \cdot \lambda_{\max}$. This stage uses only the first 100 days and is discarded from all reported metrics.

To tune $c$, we conduct a geometric grid search over $c \in [10^{-4}, 1)$ with 40 points on the full dataset. Our tuning criterion is designed to be fair to BCI: we first identify all candidate values
\[
\mathcal{C} = \left\{c : \left|\hat{m}(c) - \bar{\alpha}\right| \leq 0.5\%\right\},
\]
where $\hat{m}(c)$ is the empirical miscoverage rate on days 100+ and $\bar{\alpha} = 0.1$ is the target. Among $\mathcal{C}$, we select the $c$ whose 50-day rolling miscoverage standard deviation is closest to that of ACI:
\[
c^* = \arg\min_{c \in \mathcal{C}} \left| \hat{\sigma}(c) - \hat{\sigma}_{\text{ACI}} \right|,
\]
where $\hat{\sigma}(\cdot)$ denotes the standard deviation of the 50-day rolling miscoverage rate on days 100+. This criterion allows BCI to be as stable as it naturally wants to be while remaining calibrated, rather than forcing it into an artificially volatile regime to match ACI's standard deviation. If no candidate matches the miscoverage target, the tolerance is relaxed to $\pm 1\%$.

BCI has an $O(T^2)$ complexity for each $t$, making it computationally efficient when $T$ is small (e.g. $T=3$). The full experimental pipeline, including hyperparameter tuning, completes within a few hours on a standard laptop.

\paragraph{PID Tuning.}
PID requires two hyperparameters: the learning rate $\eta$ and the integral 
gain $K_I$. The saturation parameter $C_{\text{sat}}$ is set following \citet{angelopoulos2023conformal}:
\[
C_{\text{sat}} = \frac{2}{\pi}\left(\left\lceil \log(T) 
\cdot 0.05 \right\rceil - \frac{1}{\log T}\right),
\]
where $T$ is the dataset length.

\subsection{Stochastic model for the volatility series}
\label{sec:appendix-garch}
In the definition below, we introduce the GARCH(1, 1) process which we will later use as a stochastic model for the return series $r_1, \dots, r_K$.
\begin{definition}[GARCH(1, 1)]\label{def:garch11}
Let $p_k(r_1, \dots, r_k;\mu, \omega, a, b)$ be a joint probability density function of $r_1, \dots, r_k$ with parameters $(\mu, \omega, a, b)$.
$p_k$ is defined through the following sampling process: For each $k\geq 1$
\begin{enumerate}
	\item Define ${\sigma_k}^2 = \omega + a {\epsilon_{k-1}}^2 + b {\sigma_{k-1}}^2$ with the convention that $\epsilon_0 = 0$ and $\sigma_0=0$.
	\item Sample $e_k\sim\cN(0, 1).$
	\item Set $\epsilon_k = \sigma_k e_k$ and $r_k = \epsilon_k + \mu$.
\end{enumerate}
\end{definition}

\paragraph{Forecasting with GARCH.}
Now we introduce how to use the GARCH model to generate prediction intervals.
The stochastic model specifies the conditional distribution
\[
r_{K+1} | r_1, \dots, r_K \sim \cN(\mu, \sigma_{K+1}^2).
\]
Therefore, once we know the values of $\sigma_{K+1}$ and $\mu$, we know the conditional distribution $r_{K+1}|r_1, \dots, r_K$.
With a bit of algebra, we can show that
\[
\sigma_{K+1}^2 = \omega \frac{1-b^{K+1}}{1-b} + a \sum_{k=1}^{K} b^{K-k} (r_k - \mu)^2
\]
The unknown variables in the equation above are $(\mu, \omega, a, b)$, which can be estimated by applying MLE on historical data $r_1, r_2, \dots, r_K$:
\[
(\hat{\mu}, \hat{\omega}, \hat{a}, \hat{b}) =
\arg\max_{(\mu, \omega, a, b)} \log p_K(r_1, \dots, r_K;\mu, \omega, a, b).
\]
We use python package \texttt{arch}\footnote{\url{https://arch.readthedocs.io/en/latest/univariate/introduction.html}} to perform the fitting.
This gives an estimate of the conditional distribution $r_{K+1}|r_1, \dots, r_K \sim \cN(\hat{\mu}, \hat{\sigma}_{K+1}^2)$, where
\[
\hat{\sigma}_{K+1}^2 = \hat{\omega} \frac{1-\hat{b}^{K+1}}{1-\hat{b}} +
\hat{a} \sum_{k=1}^{K}\hat{b}^{K-k} (r_k-\hat{\mu})^2.
\]
Let $Q_{\mu, \sigma}:[0, 1]\rightarrow [0, \infty]$ be the quantile function of the squared normal distribution $\cN(\mu, \sigma)^2$, meaning that
\[
\P(\cN(\mu, \sigma^2)^2 \leq Q_{\mu, \sigma}(1-\beta)) = 1-\beta.
\]
Using $Q_{\mu, \sigma}$, a natural prediction interval for $r_{K+1}$ is
\[
\hat{C}_{K+1}(\beta) =
\bigl(Q_{\hat{\mu}, \hat{\sigma}_{K+1}^2}(\beta/2),
 Q_{\hat{\mu}, \hat{\sigma}_{K+1}^2}(1-\beta/2)\bigr).
\]
Intuitively, we expect $\hat{C}_{K+1}(\beta)$ to have mis-coverage rate $\beta$.

\section{The dynamic programming algorithm for \texorpdfstring{$\alpha_{t|t}^{*}$}{alpha t|t*}}\label{sec:bg}

\subsection{Background on stochastic control and dynamic programming}\label{sec:dp}
This is a self-contained section outlining some background on stochastic control \cite{dp} needed for this paper.
To be consistent with the control literature, we use the standard notation in this section.
Consider a finite horizon stochastic system defined by the update rule
\begin{equation}
	x_{t+1} = f_t(x_t, u_t, w_t)~ \text{for}~ t=1, 2, \dots, T,
\end{equation}
where $x_t$ is the state variable, $u_t$ is the control variable chosen by the analyst, and $w_t$ is the random disturbance to the system at time $t$ that affects the next state $x_{t+1}$. The control $u_t$ can depend on any information available at time $t$ but not on future knowledge. The disturbance $w_t$ are sampled from $w_t\sim P(\cdot|x_t, u_t)$ and can depend on the state $x_t$ and the control $u_t$. At each time $t$, the analyst bears the cost
\begin{equation}
	g_t(x_t, u_t, w_t).
\end{equation}
At the terminal time $T+1$, the analyst bears the cost $g_{T+1}(x_{T+1})$.
The planning strategy is described by policy $\pi = \{\mu_t\}_{t=1,\dots, T}$, where each $\mu_t$ determines the control at time $t$ through
\begin{equation}
	u_t = \mu_t(x_t).
\end{equation}
Given an initial state $x_1$, the expected cost under policy $\pi$ is
\begin{equation}
	J_\pi(x_1) = \E_{w_{1:T}} \left[g_{T+1}(x_{T+1}) + \sum_{t=1}^{T} g_t(x_t, \mu_t(x_t), w_t)\right].
\end{equation}
The problem of optimal control is to find the optimal policy
\begin{equation}\label{eqn:optimal-policy}
	\pi_{x_1}^* = \arg\min_\pi J_\pi(x_1).
\end{equation}

The optimal control problem admits exact solution through dynamic programming. To introduce the dynamic programming algorithm, we first define auxiliary functions
\begin{equation}
	J_t(x_t) := \text{The optimal expected cost given that we start at state $x_t$ at time $t$.}
\end{equation}
These functions are called ``cost-to-go'' function in dynamic programming literature.
Under this definition, we have that
\[J_{T+1}(x_{T+1}) = g_{T+1}(x_{T+1}).\]
Now suppose that we have solved function $J_{t+1}$ exactly. We can solve $J_t$ according the update rule
\begin{equation}\label{eqn:dp-update}
J_t(x) = \min_{u} \E_{w\sim P(\cdot|x, u)}\left[g_t(x, u, w) + J_{t+1}(f_t(x, u, w))\right].	
\end{equation}
Mathematically, the minimizer $u^*$ of the program in \eqref{eqn:dp-update} will depend on $x$, which defines the relation $u^*=\mu_t(x)$. A standard result that establishes the optimality of dynamic programming algorithm dictates the policy $\{\mu_t\}$ find through \eqref{eqn:dp-update} is the same as the optimal policy in \eqref{eqn:optimal-policy}.

\subsection{Proof of Proposition \ref{prop:Dt-increasing}}
\label{subsec:proof-dt-increasing}
\begin{proof}
All costs in \eqref{eq:bci-terminal-cost} and \eqref{eqn:bci-bellman-update}
are nonnegative. Assumption~\ref{ass:pred-sets} gives $L_{s|t}(1)=0$.
Choosing $\alpha=1$ at every remaining step therefore gives
\[
0\le J_{s|t}(\rho)\le\lambda_t+\tau1\{s=t\}K(1,\bar\alpha)<\infty.
\]
The terminal function in \eqref{eq:bci-terminal-cost} is non-decreasing.
Suppose $J_{s+1|t}$ is non-decreasing. For $0\le\rho<s-t$ and
$\alpha\in[0,1]$, since $0\le F_{s|t}(\alpha)\le1$,
\begin{equation}
\bigl(1-F_{s|t}(\alpha)\bigr)J_{s+1|t}(\rho)
+F_{s|t}(\alpha)J_{s+1|t}(\rho+1)
\le\bigl(1-F_{s|t}(\alpha)\bigr)J_{s+1|t}(\rho+1)
+F_{s|t}(\alpha)J_{s+1|t}(\rho+2).
\label{eq:bci-monotone-continuation}
\end{equation}
Adding $L_{s|t}(\alpha)+\tau1\{s=t\}K(\alpha,\bar\alpha)$ to both sides
of \eqref{eq:bci-monotone-continuation} and taking infima over $\alpha$
in \eqref{eqn:bci-bellman-update} yields $J_{s|t}(\rho)\le J_{s|t}(\rho+1)$.
Backward induction proves monotonicity for $s=t,\ldots,t+T$ and hence
$D_{s|t}(\rho)=J_{s+1|t}(\rho+1)-J_{s+1|t}(\rho)\ge0$.
\end{proof}

\section{Proof of Theorem \ref{thm:global-guarantee}}
\label{sec:thm-proof}
\begin{proof}
By \eqref{eqn:lambda-update},
\begin{equation}\label{eqn:coverage_bound}
\begin{aligned}
\lambda_{m+K+1}-\lambda_{m+1}
&=\gamma\sum_{t=m+1}^{m+K}(\err_t-\bar\alpha),\\
\left|\frac1K\sum_{t=m+1}^{m+K}\err_t-\bar\alpha\right|
&=\frac{|\lambda_{m+K+1}-\lambda_{m+1}|}{K\gamma}.
\end{aligned}
\end{equation}

Now we prove that $\lambda_k\in [ - \gamma\bar{\alpha}, \lambda_{\max} + \gamma(1-\bar{\alpha})]$ for all $k$ by induction.
For $k=1$, this follows from $\lambda_1\in[0,\lambda_{\max}]$. Suppose the claim holds for some $k\ge1$.
\begin{enumerate}
\item If $\lambda_k < 0$, by \eqref{eqn:trivial_property}, $\alpha_k = 1$ and hence $\err_k = 1$. As a result, 
\[
-\gamma\bar{\alpha}\le \lambda_k < \lambda_{k+1} = \lambda_{k} + \gamma(1-\bar{\alpha})< \gamma \le \lambda_{\max}.
\]

\item If $\lambda_k>\lambda_{\max}$, then $\alpha_k = 0$ and hence $\err_k = 0$.
As a result,
\[
\lambda_{\max} + \gamma(1 - \bar{\alpha})\ge \lambda_{k} > \lambda_{k+1} = \lambda_{k} - \gamma\bar{\alpha} > \lambda_{\max} - \gamma\bar{\alpha}\ge \lambda_{\max}-\gamma > 0.
\]
\item If $\lambda_k\in[0, \lambda_{\max}]$, then we either subtract $\gamma\bar{\alpha}$ from $\lambda_{k}$ or add $\gamma(1 - \bar{\alpha})$ onto $\lambda_k$.
This guarantees that $\lambda_{k+1}\in [- \gamma\bar{\alpha}, \lambda_{\max} + \gamma(1-\bar{\alpha})]$.
\end{enumerate}
Thus the induction hypothesis holds for every $k\ge1$, and
\begin{equation}\label{eq:coverage-weight-difference}
|\lambda_{m+K+1}-\lambda_{m+1}|
\le\lambda_{\max}+\gamma(1-\bar\alpha)+\gamma\bar\alpha
=\lambda_{\max}+\gamma.
\end{equation}
Combining \eqref{eqn:coverage_bound} and \eqref{eq:coverage-weight-difference} with $\gamma=c\lambda_{\max}$ gives \eqref{eqn:local-coverage}.

\end{proof}

\section{Formal statement and proof of Theorem \ref{thm:stationarity}}\label{sec:proof_stationarity}

\subsection{Setting}
Fix integers $p\ge 1$, $B\ge 1$, and $\bar{\alpha} = 1/q$ for some  integer $q\ge 2$. Fix $\lambda_{\max}>0$ and $c\in(0,1)$, set $\gamma:=c\lambda_{\max}$, and set
\[
\alpha:=\frac1q,
\quad
\I:=\left[-\frac\gamma q,\,\lambda_{\max}+\gamma\Bigl(1-\frac1q\Bigr)\right] = \lambda_{\max}\cdot \left[-\frac c q,\,1+c\Bigl(1-\frac1q\Bigr)\right]\] 
Write
\[
Z_t:=(Y_t,\dots,Y_{t-p+1})\in \mathcal{Z}=\R^p.
\]
We assume that $(Y_t)$ is Markov of order $p$ with one-step transition density $p_\star(\cdot\mid z)$ relative to Lebesgue measure,  meaning that for every Borel set $A\subset\R$,
\[
\P\bigl(Y_{t+1}\in A\mid Y_t,Y_{t-1},\dots\bigr)
=
\int_A p_\star(x\mid Z_t)\,dx.
\]
At time $t$, the rolling window prediction algorithm uses $Z_t$ and produces a one-step predictive CDF: 
\[
y\mapsto F(z,y),
\quad z\in \mathcal{Z}.
\]
Then the nominal prediction interval is given by 
\[C(\beta) = \bigl(F^{-1}(z, \beta/2), F^{-1}(z, 1-\beta/2)\bigr).\]
Then 
\[
\beta_t:=s(Z_t,Y_{t+1}),
\]
where 
\[
s(z,y):=2\min\{F(z,y),1-F(z,y)\}\in[0,1],
\quad (z,y)\in \mathcal{Z}\times\R,
\]
Define the sequence of $\beta$'s as 
\[
\Gamma_t:=(\beta_{t-B},\dots,\beta_{t-1})\in[0,1]^B.
\]
Let
\[
a:\mathcal{Z}\times[0,1]^B\times\I\to[0,1]
\]
be the map from the input $(Z_t, \Gamma_t, \lambda_t)$ to the resulting solution $\alpha_{t+1}$. Using the fact that both the length and miscoverage terms are monotone in $(\alpha_{s|t})_{s=t}^{t+T-1}$, we have 
\eqref{eqn:alpha-t-truncate}: 
\begin{equation}
a(z,\gamma,\lambda)=1\quad\text{if }\lambda\le 0,
\quad
a(z,\gamma,\lambda)=0\quad\text{if }\lambda\ge \lambda_{\max}.
\label{eq:clipping}
\end{equation}
Define
\begin{equation}
\err_t:=1\{a(Z_t,\Gamma_t,\lambda_t)\ge\beta_t\},
\quad
\lambda_{t+1}=\lambda_t+\gamma\left(\err_t-\frac1q\right),
\quad t\ge 1.
\label{eq:lambda-update}
\end{equation}
Consider the concatenated state
\[
X_t:=(Z_t,\Gamma_t,\lambda_t)\in \Xsf:=\mathcal{Z}\times[0,1]^B\times\I.
\]
Since $Z_t$ is Markov and $(\Gamma_t, \lambda_t)$ is a function of $Z_t$, $X_t$ is also Markov. Let $P$ denote the transition kernel of $(X_t)_{t\ge1}$.

For a weight $\lambda\in\I$, define 
\begin{equation}
\label{eq:admissible-error-set}
\Esf(\lambda):=
\begin{cases}
\{1\}, & \lambda\le 0,\\[0.2em]
\{0,1\}, & 0<\lambda<\lambda_{\max},\\[0.2em]
\{0\}, & \lambda\ge\lambda_{\max}.
\end{cases}
\end{equation}
We say that a sequence $w_1,\dots,w_N\in\{0,1\}$ is admissible from
$\lambda$ if, after setting $\lambda_0=\lambda$ and
\[
\lambda_j=\lambda_{j-1}+\gamma\left(w_j-\frac1q\right),
\qquad j=1,\dots,N,
\]
one has 
\[w_j\in\mathcal E(\lambda_{j-1}) \text{ for every }j.\]
Equivalently, the sequence never asks for a bit that is forbidden by the clipped boundary rule.

Fix an initial value $\lambda_1\in\I$. For $r=1,\dots,q$, define
\begin{align}
\Lambda_r&:=\I\cap\bigl(\lambda_1-(r-1)\gamma/q+\gamma\mathbb Z\bigr),
\label{eq:phase-level-set}\\
\Xsf_r&:=\mathcal{Z}\times[0,1]^B\times\Lambda_r,
\label{eq:phase-state-space}
\end{align}
where $\mathbb Z$ is the set of integers. We will show in Lemma \ref{lem:phase} that $\lambda_t \in \Lambda_r$ if $t - r$ is divisible by $q$. 

For each $r\in\{1,\dots,q\}$, define the kernel $Q_r$ on $\Xsf_r$ by
\[
Q_r(x,A):=P^q(x,A),
\quad x\in\Xsf_r,
\quad A\subseteq\Xsf_r\text{ measurable}.
\]

Lastly, we define 
\begin{equation}\label{eq:T*_L_n}
\begin{aligned}
T_*&:=\left\lceil q\left(\frac1c+1\right)\right\rceil,
\quad L = \left\lceil \frac{1}{q}(p + B + T_*)\right\rceil,\\
n&=Lq-(p+B),\quad N:=Lq=n+p+B.
\end{aligned}
\end{equation}

\subsection{Assumptions}\label{subsec:stationarity_assumptions}

\begin{assumption}
\label{ass:true-compact-density}
For every nonempty compact set $A\subset\R$,
\[
\inf_{x\in A,\ z\in \mathcal{Z}}p_\star(x\mid z)>0.
\]
\end{assumption}

\begin{assumption}
\label{ass:locscale-F}
There exist a continuous strictly increasing CDF $F_0$ on $\R$ with $F_0(0)=1/2$, functions $\widehat m:\mathcal{Z}\to\R$ and $\widehat\sigma:\mathcal{Z}\to(0,\infty)$, and constants
\[
m_-\le m_+,
\quad
0<\sigma_-\le\sigma_+,
\]
such that
\begin{equation}
F(z,y)=F_0\left(\frac{y-\widehat m(z)}{\widehat\sigma(z)}\right),
\quad
m_-\le\widehat m(z)\le m_+,
\quad
\sigma_-\le\widehat\sigma(z)\le\sigma_+,
\label{eq:loc-scale-F}
\end{equation}
for every $z\in \mathcal{Z}$ and $y\in\R$.
\end{assumption}

\begin{assumption}
\label{ass:finite-actions}
There exists $0 < \ell_a \le h_a < 1$ such that
\[
a(z,\gamma,\lambda)\in [\ell_a, h_a]
\quad\text{for every }z\in \mathcal{Z},\ \gamma\in[0,1]^B,\ 0<\lambda<\lambda_{\max}.
\]
\end{assumption}
\begin{remark}
Assumption \ref{ass:finite-actions} can be satisfied by forcing $\alpha_t$ to be $\ell_a$ when $\lambda_t < \lambda_{\max}$ but the resulting DP solution is less than $\ell_a$ and $h_a$ when $\lambda_t> 0$ but the resulting DP solution is less than $\ell_a$. In real implementations, we can simply round $\lambda$ to achieve it.
% For example, we can set $\ell_a=0.01\bar{\alpha}$ and $h_a=2\bar{\alpha}$.
\end{remark}

\begin{assumption}
\label{ass:central-overlap}
With the constants from Assumption~\ref{ass:locscale-F} and the value $h_a$ from Assumption~\ref{ass:finite-actions},
\begin{equation}
m_+ + \sigma_- F_0^{-1}(h_a/2)
<
m_- + \sigma_- F_0^{-1}(1-h_a/2).
\label{eq:central-overlap-a}
\end{equation}
\end{assumption}

\subsection{Main result}
With the assumptions, we state the formal version of Theorem \ref{thm:stationarity}. 

\begin{theorem}
\label{thm:cycle}
Assume Assumptions~\ref{ass:true-compact-density}, \ref{ass:locscale-F}, \ref{ass:finite-actions}, \ref{ass:central-overlap}, and a constant $B$. Let
\[
\eta:=\Law(Z_1,\Gamma_1)\otimes\delta_{\lambda_1},
\quad
\eta_r:=\eta P^{r-1},
\quad r=1,\dots,q.
\]
Then the following hold.
\begin{enumerate}
\item For each $r = 1,\ldots,q$, there exists a measure $\mu_r$ on $\Xsf_r$ such that
\[
\Law(X_{mq+r})\longrightarrow\mu_r
\,\,\text{in total variation}\quad\text{as }m\to\infty
\]
Moreover, these measures form a cyclic group:
\[
\mu_{r+1}=\mu_rP,
\quad r=1,\dots,q-1,
\quad
\mu_1=\mu_qP.
\]
Consequently,
\[
\overline\mu:=\frac1q\sum_{r=1}^q\mu_r
\]
is invariant for $P$.

\item Let $\E_\lambda$ denote expectation with respect to $X_t$ with $\lambda_1 = \lambda$. Then, for any $\lambda_1 \in (0, \lambda_{\max})$, 
\[
\lim_{t\to\infty}
\frac1q\sum_{j=0}^{q-1}\E_{\lambda_1}[\err_{t+j}]
=
\frac1q.
\]
If $X_1 = (Z_1, \Gamma_1, \lambda_1)\sim\overline\mu$, then $(X_t)_{t\ge1}$ is stationary. Under this stationary distribution,
\[
\E[\err_t]=\frac1q
\quad\text{for every }t\ge1.
\]
\end{enumerate}
\end{theorem}

\subsection{Proof of Theorem \ref{thm:cycle}}

\subsubsection{Lemmas}
We first show that $X_t \in \Xsf_r$ whenever $t - r$ is divisible by $q$.

\begin{lemma}
\label{lem:phase}
Assume $\lambda_1\in\I$ is fixed. Then
\[
[\lambda_{t+1}]_\gamma=[\lambda_t-\gamma/q]_\gamma,
\quad t\ge 1,
\]
where $[\cdot]_\gamma$ denotes residual modulo $\gamma$. In particular,
\[
[\lambda_{t+q}]_\gamma=[\lambda_t]_\gamma,
\quad t\ge 1.
\]
For the sets $\Lambda_r$ and $\Xsf_r$ in \eqref{eq:phase-level-set} and \eqref{eq:phase-state-space}, each $\Lambda_r$ is finite and
\begin{equation}\label{eq:phase-transition}
P(x,\Xsf_{r+1})=1
\quad\text{for }x\in\Xsf_r,\quad r=1,\dots,q-1,
\quad
P(x,\Xsf_1)=1
\quad\text{for }x\in\Xsf_q.
\end{equation}
\end{lemma}

\begin{proof}
Reducing \eqref{eq:lambda-update} modulo $\gamma$ gives
\[
[\lambda_{t+1}]_\gamma=[\lambda_t+\gamma \err_t-\gamma/q]_\gamma=[\lambda_t-\gamma/q]_\gamma,
\]
because $\err_t\in\{0,1\}$. Iterating $q$ times gives
\[
[\lambda_{t+q}]_\gamma=[\lambda_t-\gamma]_\gamma=[\lambda_t]_\gamma.
\]
Since $\I$ is bounded, each intersection $\Lambda_r$ is finite. \eqref{eq:phase-transition} follows from \eqref{eq:lambda-update} and the definitions in \eqref{eq:phase-level-set} and \eqref{eq:phase-state-space}.
\end{proof}

% The minorization will randomize only the final $p+B$ observations. 
The next result introduces a useful formulation for the dynamic of $(Z_t, \Gamma_t)$.

\begin{lemma}
\label{lem:tail-state}
There is a measurable map
\[
\Theta:\R^{p+B}\to \mathcal{Z}\times[0,1]^B
\]
such that, for every integer $k\ge p+B$ and every starting state $x\in\Xsf$,
\begin{equation}
(Z_{k+1},\Gamma_{k+1})
=
\Theta(Y_{k-p-B+2},\dots,Y_{k+1})
\label{eq:tail-state-identity}
\end{equation}
for every value of the displayed observations.
\end{lemma}

\begin{proof}
For $y=(y_1,\dots,y_{p+B})\in\R^{p+B}$ and $j=1,\dots,B+1$, define
\[
z_j(y):=(y_{p+j-1},y_{p+j-2},\dots,y_j)\in \mathcal{Z}.
\]
For $j=1,\dots,B$, define
\[
\theta_j(y):=s\bigl(z_j(y),y_{p+j}\bigr).
\]
Set
\[
\Theta(y):=\bigl(z_{B+1}(y),(\theta_1(y),\dots,\theta_B(y))\bigr).
\]
This map is measurable.

Now fix $k\ge p+B$. Define
\[
Z_{k+1}=(Y_{k+1},\dots,Y_{k-p+2}),
\]
which is the last $p$ entries of $(Y_{k-p-B+2},\dots,Y_{k+1})$. Thus,
\[
\Gamma_{k+1}=(\beta_{k-B+1},\dots,\beta_{k}).
\]
For $j=1,\dots,B$, the PIT value $\beta_{k-B+j}$ is obtained by applying $s$ with a rolling window. These quantities are encoded by $z_j(y)$ and $y_{p+j}$. Therefore \eqref{eq:tail-state-identity} holds.
\end{proof}

The next result identifies two intervals of $Y_{t+1}$ on which $s(z, y)$ is guaranteed to be small and large, respectively. This serves as a building block for the Doeblin minorization condition.

\begin{lemma}
\label{lem:locscale-a2}
Assume Assumptions~\ref{ass:locscale-F} and \ref{ass:central-overlap}. Then there exist compact nondegenerate intervals $G_1,G_0\subset\R$ such that
\[
s(z,y)<\ell_a\quad\text{for every }(z,y)\in \mathcal{Z}\times G_1,
\]
and
\[
s(z,y)>h_a\quad\text{for every }(z,y)\in \mathcal{Z}\times G_0.
\]
% Moreover, $a(z,\gamma,\lambda)\in[\ell_a, h_a]$ for every $z\in \mathcal{Z}$, $\gamma\in[0,1]^B$, and $0<\lambda<\lambda_{\max}$.
\end{lemma}

\begin{proof}
Define the open interval
\[
\mathcal C_a
:=
\left(
 m_+ + \sigma_-F_0^{-1}(h_a/2),
 m_- + \sigma_-F_0^{-1}(1-h_a/2)
\right).
\]
Assumption~\ref{ass:central-overlap} says exactly that $\mathcal C_a$ is nonempty. Choose a compact interval of positive length
\[
G_0\subset\mathcal C_a.
\]
Also choose a compact interval of positive length
\[
G_1\subset\left(m_+ + \sigma_+F_0^{-1}(1-\ell_a/2),\infty\right).
\]

Let $y\in G_0$. By the definition of $\mathcal C_a$, for every $z\in \mathcal{Z}$,
\[
y-\widehat m(z)
>
\sigma_-F_0^{-1}(h_a/2)
\ge
\widehat\sigma(z)F_0^{-1}(h_a/2),
\]
because $F_0^{-1}(h_a/2)<0$ and $\widehat\sigma(z)\ge\sigma_-$. Similarly,
\[
y-\widehat m(z)
<
\sigma_-F_0^{-1}(1-h_a/2)
\le
\widehat\sigma(z)F_0^{-1}(1-h_a/2),
\]
because $F_0^{-1}(1-h_a/2)>0$ and $\widehat\sigma(z)\ge\sigma_-$. Hence
\[
h_a/2<F(z,y)<1-h_a/2,
\quad z\in \mathcal{Z},
\]
and therefore
\[
s(z,y)=2\min\{F(z,y),1-F(z,y)\}> h_a,
\quad z\in \mathcal{Z},\ y\in G_0.
\]

Now let $y\in G_1$. For every $z\in \mathcal{Z}$,
\[
y-\widehat m(z)
>
\sigma_+F_0^{-1}(1-\ell_a/2)
\ge
\widehat\sigma(z)F_0^{-1}(1-\ell_a/2),
\]
because $F_0^{-1}(1-\ell_a/2)>0$ and $\widehat\sigma(z)\le\sigma_+$. Thus $F(z,y)>1-\ell_a/2$, and hence
\[
s(z,y)<\ell_a,
\quad z\in \mathcal{Z},\ y\in G_1.
\]

% Then Assumption~\ref{ass:finite-actions} gives
% \[
% a(z,\gamma,\lambda)\in[\ell_a, h_a]
% \quad\text{for every }z\in \mathcal{Z},\ \gamma\in[0,1]^B,\ 0<\lambda<\lambda_{\max}.
% \]
% Therefore $\ell_a\le a(z,\gamma,\lambda)\le h_a$ on the interior strip. The displayed estimates give the asserted forcing-band bounds.
\end{proof}

For the remainder of the proof, fix compact intervals $G_1,G_0\subset\R$ supplied by Lemma~\ref{lem:locscale-a2}. The following lemma shows that $G_1$ and $G_0$ are able to pin down $\err_t$. 

\begin{lemma}
\label{lem:forcing}
Assume Assumptions~\ref{ass:locscale-F}, \ref{ass:finite-actions}, and \ref{ass:central-overlap}. Let $e\in\{0, 1\}$. Then
\[
Y_{t+1}\in G_e\Longrightarrow \err_t=e.
\]
\end{lemma}

\begin{proof}
If $e=1$, then $Y_{t+1}\in G_1$. If $\lambda_t\le0$, then $a(Z_t,\Gamma_t,\lambda_t)=1$ by clipping and
\[
\beta_t=s(Z_t,Y_{t+1})\le h_a < 1,
\]
so $\err_t=1$. If $e=1$ and $0<\lambda_t<\lambda_{\max}$, then Lemma \ref{lem:locscale-a2} implies
\[
\beta_t<\ell_a\le a(Z_t,\Gamma_t,\lambda_t),
\]
so again $\err_t=1$.

If $e=0$, then $Y_{t+1}\in G_0$. If $\lambda_t\ge\lambda_{\max}$, clipping gives $a(Z_t,\Gamma_t,\lambda_t)=0<\beta_t$, hence $\err_t=0$. If $0<\lambda_t<\lambda_{\max}$, then
\[
\beta_t=s(Z_t,Y_{t+1})>h_a\ge a(Z_t,\Gamma_t,\lambda_t),
\]
so $\err_t=0$.
\end{proof}

Lemma \ref{lem:locscale-a2} also implies a weaker version of the Doeblin minorization condition for $(Y_t)$.

\begin{lemma}
\label{lem:true-band-density}
Under Assumption \ref{ass:true-compact-density}, there exists a constant $c_G>0$ such that, for every $z\in \mathcal{Z}$ and every Borel set
$A\subset G_0\cup G_1$,
\[
\P\bigl(Y_{t+1}\in A\mid Z_t=z\bigr)\ge c_G\Leb(A).
\]
\end{lemma}

\begin{proof}
The set $G_0\cup G_1$ is compact and nonempty. By Assumption~\ref{ass:true-compact-density},
\[
c_G:=\inf_{x\in G_0\cup G_1,\ z\in \mathcal{Z}}p_\star(x\mid z)>0.
\]
For every Borel set $A\subset G_0\cup G_1$ and every $z\in \mathcal{Z}$,
\[
\P\bigl(Y_{t+1}\in A\mid Z_t=z\bigr)
=
\int_A p_\star(x\mid z)\,dx
\ge c_G\Leb(A),
\]
which proves the claim.
\end{proof}

Applying Lemma~\ref{lem:true-band-density} recursively, we can obtain the following result on $(Y_2, \ldots, Y_{k+1})$.

\begin{lemma}
\label{lem:block-lower}
Under Assumption \ref{ass:true-compact-density}, for every integer $k\ge1$, every starting state $x\in\Xsf$, and every Borel set $A\subset (G_0\cup G_1)^k$,
\[
\P_x\bigl((Y_2,\dots,Y_{k+1})\in A\bigr)
\ge c_G^k\,\Leb_k(A).
\]
\end{lemma}

\begin{proof}
Fix $x\in\Xsf$ and a rectangle $A=A_1\times\cdots\times A_k$ with each $A_j\subset G_0\cup G_1$ Borel. Write $\E_x$ for expectation under $\P_x$.
Under $\P_x$, $Z_1$ is fixed and $Z_j$ is determined by $Z_1,Y_2,\dots,Y_j$.
The order-$p$ Markov property and Lemma~\ref{lem:true-band-density} give
\begin{equation}
\P_x(Y_{j+1}\in A_j\mid Y_2,\dots,Y_j)
=\int_{A_j}p_\star(y\mid Z_j)\,dy
\ge c_G\Leb(A_j),\qquad j=1,\dots,k,
\label{eq:block-conditional-probability-bound}
\end{equation}
with no conditioning when $j=1$.
For $j=2,\dots,k$, the tower property and
\eqref{eq:block-conditional-probability-bound} imply
\begin{equation}
\begin{aligned}
\P_x\Bigl(\bigcap_{i=1}^j\{Y_{i+1}\in A_i\}\Bigr)
&=\E_x\Bigl[\prod_{i=1}^{j-1}1\{Y_{i+1}\in A_i\}
\,\P_x(Y_{j+1}\in A_j\mid Y_2,\dots,Y_j)\Bigr]\\
&\ge c_G\Leb(A_j)\,
\E_x\Bigl[\prod_{i=1}^{j-1}1\{Y_{i+1}\in A_i\}\Bigr]\\
&=c_G\Leb(A_j)\,
\P_x\Bigl(\bigcap_{i=1}^{j-1}\{Y_{i+1}\in A_i\}\Bigr).
\end{aligned}
\label{eq:block-rectangle-recursion}
\end{equation}
Starting from $\P_x(Y_2\in A_1)\ge c_G\Leb(A_1)$, which is
\eqref{eq:block-conditional-probability-bound} at $j=1$, iterating
\eqref{eq:block-rectangle-recursion} gives
\begin{equation}
\P_x\bigl((Y_2,\dots,Y_{k+1})\in A\bigr)
\ge c_G^k\prod_{j=1}^k\Leb(A_j)
=c_G^k\Leb_k(A).
\label{eq:block-rectangle-lower-bound}
\end{equation}
Finite additivity gives the inequality in \eqref{eq:block-rectangle-lower-bound}
for finite disjoint unions of these rectangles. Continuity under increasing
and decreasing limits and the monotone-class theorem extend it to all Borel
subsets of $(G_0\cup G_1)^k$.
\end{proof}
% The following deterministic result explains why the forced prefix can erase dependence on the initial weight. After the drain time, any two levels in the same slice are driven to the same value.

The following two lemmas provide another building block for the Doeblin minorization condition.

\begin{lemma}
\label{lem:drain}
For $\lambda\in\I$, define
\begin{equation}
\label{eq:deterministic-error-rule}
d(\lambda):=\begin{cases}
1, & \lambda\le0,\\[0.2em]
0, & \lambda>0,
\end{cases}
\quad
D(\lambda):=\lambda+\gamma\left(d(\lambda)-\frac1q\right).
\end{equation}
For every $\Lambda_r$ and every $\lambda,\lambda'\in\Lambda_r$,
\[
D^t(\lambda)=D^t(\lambda')
\quad\text{for every }t\ge T_*.
\]
\end{lemma}

\begin{proof}
First, $D$ maps $\I$ into itself. If $\lambda>0$, then $D(\lambda)=\lambda-\gamma/q\ge-\gamma/q$. If $\lambda\le0$, then $\lambda\in[-\gamma/q,0]$ and
\[
D(\lambda)\in\left[\gamma\left(1-\frac2q\right),\gamma\left(1-\frac1q\right)\right]\subset\I.
\]

Normalize $\lambda$ by setting
\[
w:=\frac{q\lambda}{\gamma}.
\]
The interval $\I$ becomes $[-1,U]$, where
\[
U:=\frac{q}{\gamma}\left\{\lambda_{\max}+\gamma\left(1-\frac1q\right)\right\}
=q\left(\frac1c+1\right)-1.
\]
Define the following map
\[
w\mapsto
\begin{cases}
w+q-1, & w\le0,\\[0.2em]
w-1, & w>0.
\end{cases}
\]
The half-open interval $(-1,q-1]$ is invariant under this map. Also, every trajectory starting in $[-1,U]$ enters $(-1,q-1]$ by time $\lceil U\rceil+1=\left\lceil q\left(\frac1c+1\right)\right\rceil=T_*$: while $w>0$ the map subtracts one, and after the first nonpositive value one more step sends the point into $(-1,q-1]$.

If $\lambda,\lambda'\in\Lambda_r$, then their normalized values differ by an integer multiple of $q$. The above map preserves the property that this difference is an integer multiple of $q$. Once the normalized $\lambda$ and $\lambda'$ lie in $(-1,q-1]$, they must be equal because they differ by an integer of $q$ and $(-1,q-1]$ has length $q$. This proves the claim.
\end{proof}

\begin{lemma}
\label{lem:drain-sequences}
For each $r\in \{1,\ldots, q\}$, there exist a common entrance level $\widehat\lambda_r\in\I$, a reference level $\overline\lambda_r\in\Lambda_r$, and a sequence
\[
c_{r,1},\dots,c_{r,p+B}\in\{0,1\}
\]
with the following property. For every initial value $\lambda\in\Lambda_r$, there is a prefix sequence
\[
d_{r,\lambda,1},\dots,d_{r,\lambda,n}\in\{0,1\}
\]
such that the full sequence
\begin{equation}
\label{eq:admissible-bit-sequence}
w_{r,\lambda}:=(d_{r,\lambda,1},\dots,d_{r,\lambda,n},c_{r,1},\dots,c_{r,p+B})
\end{equation}
is admissible from $\lambda$. If
\begin{equation}
\label{eq:admissible-weight-sequence}
\lambda^{r,\lambda}_0:=\lambda,
\quad
\lambda^{r,\lambda}_j
:=\lambda^{r,\lambda}_{j-1}+\gamma\left(w_{r,\lambda,j}-\frac1q\right),
\quad j=1,\dots,N,
\end{equation}
then
\[
w_{r,\lambda,j}\in\Esf(\lambda^{r,\lambda}_{j-1})
\quad\text{for every }j=1,\dots,N,
\]
where $\Esf(\lambda)$ is defined in \eqref{eq:admissible-error-set},
and
\[
\lambda^{r,\lambda}_n=\widehat\lambda_r,
\quad
\lambda^{r,\lambda}_N=\overline\lambda_r,
\]
where $N$ is defined in \eqref{eq:T*_L_n}.
\end{lemma}

\begin{proof}
Fix $r\in \{1,\ldots, q\}$. For the prefix, set
\[
\lambda^{r,\lambda}_0:=\lambda,
\quad
 d_{r,\lambda,j}:=d(\lambda^{r,\lambda}_{j-1}),
\quad
\lambda^{r,\lambda}_j:=D(\lambda^{r,\lambda}_{j-1}),
\quad j=1,\dots,n.
\]
Set $w_{r,\lambda,j}:=d_{r,\lambda,j}$ for $j=1,\dots,n$. Each prefix bit is admissible by definition \eqref{eq:deterministic-error-rule} of $d$, because $d(\lambda)\in\Esf(\lambda)$ for every $\lambda\in\I$. Since $n\ge T_*$, Lemma~\ref{lem:drain} implies that $\lambda^{r,\lambda}_n$ is the same for every starting $\lambda\in\Lambda_r$. Denote this common value by $\widehat\lambda_r$.

Starting from $\widehat\lambda_r$, continue the same rule for the final $p+B$ steps:
\[
\widehat\lambda_{r,0}:=\widehat\lambda_r,
\quad
c_{r,j}:=d(\widehat\lambda_{r,j-1}),
\quad
\widehat\lambda_{r,j}:=D(\widehat\lambda_{r,j-1}),
\quad j=1,\dots,p+B.
\]
Set $\overline\lambda_r:=\widehat\lambda_{r,p+B}$. For $j=n+1,\dots,N$, define
\[
w_{r,\lambda,j}:=c_{r,j-n},
\quad
\lambda^{r,\lambda}_j:=\widehat\lambda_{r,j-n}.
\]
The sequence $c_{r,1},\dots,c_{r,p+B}$ is common across all initial values in $\Lambda_r$. The full sequence is admissible at every step, reaches $\widehat\lambda_r$ after the prefix, and reaches $\overline\lambda_r$ after $N=Lq$ steps. Since $Lq$ is a multiple of $q$, Lemma~\ref{lem:phase} implies $\overline\lambda_r\in\Lambda_r$.
\end{proof}

% The forcing intervals are useful only if the true process can enter them uniformly. The next result obtains this lower bound directly from the compact-band positivity assumption.

% For an admissible pair $(\lambda,e)$ with $e\in\Esf(\lambda)$, define
% \[
% H(\lambda,e):=
% \begin{cases}
% G_1, & e=1,\\[0.3em]
% G_0, & e=0.
% \end{cases}
% \]

Combining all previous lemmas, we obtain the following result.

\begin{proposition}
\label{prop:minorization}
Assume Assumptions~\ref{ass:true-compact-density}, \ref{ass:locscale-F}, \ref{ass:finite-actions}, and \ref{ass:central-overlap}. Fix $r\in\{1,\dots,q\}$. Use the sequences $\lambda^{r,\lambda}_j$ and $w_{r,\lambda,j}$ defined in \eqref{eq:admissible-weight-sequence} and \eqref{eq:admissible-bit-sequence}, respectively, in Lemma~\ref{lem:drain-sequences}. Define
\[
S_{r,\lambda}:=
\prod_{j=1}^{n} G_{w_{r,\lambda,j}}
\subset (G_0\cup G_1)^n,
\]
with the convention that $\Leb_0(S_{r,\lambda})=1$ when $n=0$. Define
\[
\widehat\lambda_{r,0}:=\widehat\lambda_r,
\quad
\widehat\lambda_{r,j}:=\widehat\lambda_{r,j-1}+\gamma\left(c_{r,j}-\frac1q\right),
\quad j=1,\dots,p+B,
\]
and
\[
T_r:=\prod_{j=1}^{p+B}G_{c_{r,j}}
\subset (G_0\cup G_1)^{p+B}.
\]
Further, define the measure $\nu_r$:
\[
\nu_r(A):=\frac{1}{\Leb_{p+B}(T_r)}
\Leb_{p+B}\bigl(\{y\in T_r:(\Theta(y),\overline\lambda_r)\in A\}\bigr).
\]
Set
\[
\delta_r:=c_G^N\,\Leb_{p+B}(T_r)
\min_{\lambda\in\Lambda_r}\Leb_n(S_{r,\lambda}).
\]
Then $\delta_r>0$ and
\begin{equation}
Q_r^L(x,A)=P^{Lq}(x,A)
\ge \delta_r\,\nu_r(A)
\quad
\text{for every }x\in\Xsf_r\text{ and measurable }A\subseteq\Xsf_r.
\label{eq:slice-minorization}
\end{equation}
\end{proposition}

\begin{proof}
Fix $r\in \{1,\ldots, q\}$ and $x=(z,\gamma,\lambda)\in\Xsf_r$. Let
\[
D_{r,\lambda}:=S_{r,\lambda}\times T_r\subset (G_0\cup G_1)^N.
\]
On the event
\[
(Y_2,\dots,Y_{N+1})\in D_{r,\lambda},
\]
the observation at step $j$ belongs to $G_{w_{r,\lambda,j}}$ for $j=1,\dots,N$. By Lemma~\ref{lem:forcing} and the admissibility of the sequence,
\[
\err_j=w_{r,\lambda,j}
\quad\text{and}\quad
\lambda_{j+1}=\lambda^{r,\lambda}_j,
\quad j=1,\dots,N.
\]
Thus $\lambda_{N+1}=\lambda^{r,\lambda}_N=\overline\lambda_r$.

Since $N\ge p+B$, Lemma~\ref{lem:tail-state} gives
\[
(Z_{N+1},\Gamma_{N+1})
=
\Theta(Y_{N-p-B+2},\dots,Y_{N+1}).
\]
The last $p+B$ observations are exactly the coordinates in the common tail box $T_r$. Therefore, for any measurable $A\subseteq\Xsf_r$,
\[
P^N(x,A)
\ge
\P_x\Bigl((Y_2,\dots,Y_{N+1})\in E_{r,\lambda}(A)\Bigr),
\]
where
\[
E_{r,\lambda}(A):=
\Bigl\{(u,y)\in S_{r,\lambda}\times T_r:
(\Theta(y),\overline\lambda_r)\in A\Bigr\}.
\]
By Lemma~\ref{lem:block-lower},
\[
P^N(x,A)
\ge c_G^N\Leb_N(E_{r,\lambda}(A)).
\]
Fubini's theorem gives
\[
\Leb_N(E_{r,\lambda}(A))
=
\Leb_n(S_{r,\lambda})
\Leb_{p+B}\bigl(\{y\in T_r:(\Theta(y),\overline\lambda_r)\in A\}\bigr).
\]
Using the definition of $\nu_r$ and taking the minimum over the finite set $\Lambda_r$ yields
\[
P^N(x,A)
\ge
c_G^N\Leb_{p+B}(T_r)
\min_{\lambda'\in\Lambda_r}\Leb_n(S_{r,\lambda'})\,\nu_r(A)
=\delta_r\nu_r(A).
\]
Because $G_0$ and $G_1$ are nondegenerate intervals, every factor in each box has positive Lebesgue measure. Since $\Lambda_r$ is finite, $\delta_r>0$.
\end{proof}

\subsubsection{Proof of Theorem~\ref{thm:cycle}}
For Part 1, fix $r\in \{1,\ldots, q\}$. Let $\delta_r$ be the minorization constant from Proposition~\ref{prop:minorization}. Proposition~\ref{prop:minorization} gives
\[
Q_r^L(x,\cdot)\ge\delta_r\nu_r(\cdot)
\quad\text{for every }x\in\Xsf_r.
\]
Lemma~\ref{lem:doeblin-iterate}, applied to $Q=Q_r$, gives a unique invariant probability measure $\mu_r$ for $Q_r$ and the stated total-variation bound. Since
\[
\Law(X_{mq+r})=\eta_rQ_r^{m},
\]
we can define $\mu_r$ as its limit in total variation. 

When $r < q$, Lemma~\ref{lem:phase} implies that $\mu_rP$ is supported on $\Xsf_{r+1}$. Moreover,
\[
(\mu_rP)Q_{r+1}
=
\mu_rP^{q+1}
=
\mu_rQ_rP
=
\mu_rP,
\]
so $\mu_rP$ is invariant for $Q_{r+1}$. By uniqueness of the invariant law on slice $r+1$,
\[
\mu_rP=\mu_{r+1}.
\]
The same argument with $r=q$ gives $\mu_qP=\mu_1$. Averaging the cyclic identities gives
\[
\overline\mu P
=
\frac1q\sum_{r=1}^q\mu_rP
=
\frac1q\sum_{r=1}^q\mu_r
=\overline\mu.
\]
Thus $\overline\mu$ is invariant for $P$.

For Part 2, start the chain with $X_1\sim\overline\mu$. Part 1 implies that $(X_t)_{t\ge 1}$ is stationary. Since $\lambda_t$ is bounded almost surely, taking expectations in the bounded update
\[
\lambda_{t+1}-\lambda_t=\gamma\left(\err_t-\frac1q\right)
\]
under this stationary distribution gives
\[
0=\E[\lambda_{t+1}-\lambda_t]
=\gamma\left(\E[\err_t]-\frac1q\right),
\]
so $\E[\err_t]=1/q$ for every $t$.

\subsubsection{Doeblin minorization condition}

In this subsection, we revisit Doeblin minorization condition and its implication on the convergence to the stationary distribution with any initialization. We provide a self-contained proof for completeness.

\begin{lemma}
\label{lem:doeblin-iterate}
Let $Q$ be a Markov kernel on a measurable space $E$. Suppose that for some integer $L\ge1$, some $\delta\in(0,1]$, and some probability measure $\nu$ on $E$,
\begin{equation}
Q^L(x,A)\ge \delta\,\nu(A)
\quad\text{for every }x\in E\text{ and measurable }A\subseteq E.
\label{eq:doeblin-iterate}
\end{equation}
Then $Q$ has a unique invariant probability measure $\mu$, and for every probability measure $\eta$ on $E$,
\[
\|\eta Q^n-\mu\|_{\TV}
\le
(1-\delta)^{\lfloor n/L\rfloor},
\quad n\ge0.
\]
\end{lemma}

\begin{proof}
First apply the usual Doeblin decomposition to the kernel $\widetilde Q:=Q^L$. If $\delta=1$, then $\widetilde Q(x,\cdot)=\nu(\cdot)$ for every $x$ and the contraction below is immediate. If $0<\delta<1$, define
\[
R(x,A):=\frac{\widetilde Q(x,A)-\delta\nu(A)}{1-\delta}.
\]
Then $R$ is a Markov kernel and
\[
\widetilde Q(x,\cdot)=\delta\nu(\cdot)+(1-\delta)R(x,\cdot).
\]
Therefore, for any two probability measures $\eta_1,\eta_2$,
\[
\eta_1\widetilde Q-\eta_2\widetilde Q
=(1-\delta)(\eta_1R-\eta_2R),
\]
so
\begin{equation}
\|\eta_1\widetilde Q-\eta_2\widetilde Q\|_{\TV}
\le (1-\delta)\|\eta_1-\eta_2\|_{\TV}.
\label{eq:contraction-iterate}
\end{equation}
Thus $\eta\mapsto\eta\widetilde Q$ is a strict contraction on probability measures with the total-variation metric. Hence $\widetilde Q$ has a unique invariant probability measure $\mu$ and
\[
\|\eta \widetilde Q^k-\mu\|_{\TV}\le (1-\delta)^k,
\quad k\ge0.
\]

It remains to check that this same $\mu$ is invariant for $Q$. Since $\mu\widetilde Q=\mu$,
\[
(\mu Q)\widetilde Q=\mu Q^{L+1}=\mu Q^L Q=\mu Q.
\]
Thus $\mu Q$ is also invariant for $\widetilde Q$. By uniqueness for $\widetilde Q$, $\mu Q=\mu$. Therefore $\mu$ is invariant for $Q$. If $\pi Q=\pi$, then $\pi\widetilde Q=\pi$, so uniqueness for $\widetilde Q$ also gives $\pi=\mu$. Thus $\mu$ is the unique invariant law for $Q$.

Finally write $n=kL+j$ with $0\le j<L$. Since $\mu$ is invariant for $Q$ and hence for $\widetilde Q$,
\[
\eta Q^n=(\eta Q^j)\widetilde Q^k,
\]
and the contraction bound for $\widetilde Q$ gives
\[
\|\eta Q^n-\mu\|_{\TV}
=\|(\eta Q^j)\widetilde Q^k-\mu\|_{\TV}
\le (1-\delta)^k.
\]
Since $k=\lfloor n/L\rfloor$, the claim follows.
\end{proof}

\section{Proof of Theorem \ref{thm:oracle}}\label{sec:proof_oracle}

\subsection{Notation}
Fix $t$, $T$, $B$, and $\bar\alpha$ as in Theorem~\ref{thm:oracle}, and fix $\delta\in(0,1)$. We use $\Pi_t$, $\widehat J_{s|t}$, $\widehat D_{s|t}$, and $\widehat Q_{s|t}$ from Section~\ref{sec:bci-dynamic-programing}. Write
\[
\overline G_{t,B}(u):=\frac1B\sum_{j=1}^B G_{t-j}^\circ(u)
\]
for the average of the one-step conditional CDFs over the historical window.

\subsection{Lemmas}
We start by a lemma that bounds the Kolmogorov distance between two distributions based on their values on a grid.

\begin{lemma}
\label{lem:grid}
Let $H:[0,1]\to\R$ be nondecreasing and let $F:[0,1]\to\R$ be nondecreasing and $M$-Lipschitz, with $H(1)=F(1)$. Then for every integer $m\ge 1$,
\begin{equation}
\|H-F\|_\infty
\le
\max_{0\le \ell\le m}\bigl|H(\ell/m)-F(\ell/m)\bigr|+\frac{M}{m}.
\label{eq:grid}
\end{equation}
\end{lemma}

\begin{proof}
Note that $H(1)-F(1)=0$. Fix $u\in[0,1)$ and let $\ell=\lfloor mu\rfloor$.
\[
H(u)-F(u)
\le H\bigl((\ell+1)/m\bigr)-F(\ell/m).
\]
Then,
\[
\begin{aligned}
H(u)-F(u)
&\le
\bigl[H\bigl((\ell+1)/m\bigr)-F\bigl((\ell+1)/m\bigr)\bigr]
+
\bigl[F\bigl((\ell+1)/m\bigr)-F(\ell/m)\bigr]\\
&\le
\max_{0\le r\le m}|H(r/m)-F(r/m)|+\frac{M}{m}.
\end{aligned}
\]
Similarly,
\[
F(u)-H(u)
\le
F\bigl((\ell+1)/m\bigr)-H(\ell/m)
\le
\max_{0\le r\le m}|H(r/m)-F(r/m)|+\frac{M}{m}.
\]
Therefore \eqref{eq:grid} is proved.
\end{proof}

The next lemma controls the uniform deviation of the empirical PIT CDF from $\bar{G}_{t,B}$.

\begin{lemma}
\label{lem:empirical-process}
Assume \eqref{eq:key-lipschitz} and $B\ge 4$. Then for every $\delta\in(0,1)$,
\begin{equation}
\P\left(
\bigl\|\widehat G_t^{(B)}-\overline G_{t,B}\bigr\|_\infty
\le
\frac{2\bigl(\sqrt{\log(B/\delta)}+M\bigr)}{\sqrt B}
\right)
\ge 1-\delta.
\label{eq:empirical-process}
\end{equation}
\end{lemma}

\begin{proof}
Set $m:=\sqrt B-1$, so $m\ge 1$. Consider the grid
\[
\mathcal G_m:=\{\ell/m:0\le \ell\le \lfloor m\rfloor\}\cup\{1\}.
\]
For each $u\in\mathcal G_m$, set
\[
D_j(u):=1\{\beta_j\le u\}-G_j^{\circ}(u),
\qquad j=t-B,\dots,t-1.
\]
Since $G_j^{\circ}(u)=\E[1\{\beta_j\le u\}\mid \F_{j-1}]$, $\{D_j(u)\}_{j=t-B}^{t-1}$ is a martingale-difference sequence with respect to the filtration $\{\F_j\}_{j=t-B}^{t-1}$, with $\E[D_j(u)\mid\F_{j-1}]=0$. Also, $|D_j(u)|\le 1$ almost surely. Therefore Azuma--Hoeffding gives
\[
\P\left(
\left|\frac1B\sum_{j=t-B}^{t-1}D_j(u)\right|>x
\right)
\le 2\exp\left(-\frac{Bx^2}{2}\right),
\qquad x>0.
\]
Applying the union bound over $u\in\mathcal G_m$,
\[
\P\left(
\max_{u\in\mathcal G_m}
\left|\frac1B\sum_{j=t-B}^{t-1}D_j(u)\right|>x
\right)
\le 2(\sqrt B+1)\exp\left(-\frac{Bx^2}{2}\right).
\]
Choosing
\[
x=\sqrt{\frac{2\log\bigl(2(\sqrt B+1)/\delta\bigr)}{B}}
\]
shows that, with probability at least $1-\delta$,
\begin{equation}
\max_{u\in\mathcal G_m}
\bigl|\widehat G_t^{(B)}(u)-\overline G_{t,B}(u)\bigr|
\le
\sqrt{\frac{2\log\bigl(2(\sqrt B+1)/\delta\bigr)}{B}}.
\label{eq:grid-event}
\end{equation}
The same proof as Lemma~\ref{lem:grid}, using the mesh size of $\mathcal G_m$, gives
\[
\bigl\|\widehat G_t^{(B)}-\overline G_{t,B}\bigr\|_\infty
\le
\max_{u\in\mathcal G_m}
\bigl|\widehat G_t^{(B)}(u)-\overline G_{t,B}(u)\bigr|+\frac{M}{m}.
\]
Since $B\ge 4$ and $\delta\in(0,1)$,
\[
\sqrt{\frac{2\log\bigl(2(\sqrt B+1)/\delta\bigr)}{B}}
\le
\frac{2\sqrt{\log(B/\delta)}}{\sqrt B},
\qquad
\frac{M}{\sqrt B-1}\le \frac{2M}{\sqrt B}.
\]
Combining these displays yields \eqref{eq:empirical-process}. 
\end{proof}

The next lemma controls the uniform deviation of the empirical PIT CDF from $G_s^\circ$.
\begin{lemma}
\label{lem:policy-cdf-comparison}
Under \eqref{eq:kappa-assumption} and
\eqref{eq:key-lipschitz}, define
\begin{equation}
e_B(\delta):=\frac{2\bigl(\sqrt{\log(B/\delta)}+M\bigr)}{\sqrt B},
\qquad
d_{s|t}:=e_B(\delta)+\frac1B\sum_{j=1}^{B}\kappa_{t-j}+\kappa_s.
\label{eq:policy-cdf-error}
\end{equation}
There is an event $\mathcal A_t\in\F_{t-1}$ with
$\P(\mathcal A_t)\ge1-\delta$ on which
\begin{equation}
\sup_{u\in[0,1]}|G_s^\circ(u)-\widehat G_t^{(B)}(u)|
\le d_{s|t},\qquad s=t,\ldots,t+T-1.
\label{eq:policy-cdf-comparison}
\end{equation}
\end{lemma}

\begin{proof}
Since $B\ge4$ and \eqref{eq:key-lipschitz} holds,
Lemma~\ref{lem:empirical-process} gives
\begin{equation}
\mathcal A_t:=\left\{
\|\widehat G_t^{(B)}-\overline G_{t,B}\|_\infty\le e_B(\delta)
\right\}\in\F_{t-1}
\quad\text{satisfies}\quad\P(\mathcal A_t)\ge1-\delta.
\label{eq:policy-historical-event}
\end{equation}
On $\mathcal A_t$ in \eqref{eq:policy-historical-event}, the triangle inequality gives
\begin{equation}
\begin{aligned}
|G_s^\circ(u)-\widehat G_t^{(B)}(u)|
&\le|G_s^\circ(u)-u|+|u-\overline G_{t,B}(u)|
+|\overline G_{t,B}(u)-\widehat G_t^{(B)}(u)|\\
&\le\kappa_s+\frac1B\sum_{j=1}^B\kappa_{t-j}+e_B(\delta)
=d_{s|t}.
\end{aligned}
\label{eq:policy-full-history-comparison}
\end{equation}
Taking the supremum over $u$ in
\eqref{eq:policy-full-history-comparison} proves
\eqref{eq:policy-cdf-comparison}.
\end{proof}

The next lemma shows that an additional error changes the empirical optimal continuation value by at
most $|\lambda_t|/T$.
\begin{lemma}
\label{lem:policy-value-increments}
The values $\widehat J_{s|t}(r)$ are finite. For
$s=t,\ldots,t+T-1$ and $r=0,\ldots,s-t$,
\begin{equation}
|\widehat D_{s|t}(r)|\le\frac{|\lambda_t|}{T}.
\label{eq:policy-continuation-bound}
\end{equation}
\end{lemma}

\begin{proof}
Fix $s=t,\ldots,t+T$ and $r=0,\ldots,s-t$.
The terminal penalty in \eqref{eq:bci-terminal-cost} lies in
$[-|\lambda_t|,|\lambda_t|]$.
For $k=t,\ldots,t+T-1$ and $a\in[0,1]$,
the assumptions of Theorem~\ref{thm:oracle} give
\begin{equation}
L_{k|t}(a)+\tau1\{k=t\}K(a,\bar\alpha)\ge0.
\label{eq:policy-stage-cost-nonnegative}
\end{equation}
Fix $\pi\in\Pi_t$ and apply \eqref{eq:bci-state-dynamics} from
$\rho_{s|t}=r$, with $\alpha_{k|t}=\pi_{k|t}(\rho_{k|t})$ for
$k=s,\ldots,t+T-1$. Each state increment is zero or one, so
$0\le\rho_{t+T|t}\le r+t+T-s\le T$ for every PIT realization.
By \eqref{eq:policy-stage-cost-nonnegative},
\begin{equation}
\sum_{k=s}^{t+T-1}\bigl[L_{k|t}(\alpha_{k|t})
+\tau1\{k=t\}K(\alpha_{k|t},\bar\alpha)\bigr]\ge0.
\label{eq:policy-nonnegative-sum}
\end{equation}
Combining \eqref{eq:policy-nonnegative-sum} with
\eqref{eq:bci-terminal-cost} and taking expectation over the empirical PIT
draws, with the quantities chosen at time $t$ held fixed, gives
\begin{equation}
\begin{aligned}
&\E_{\beta\sim P_t^{(B)}}\Bigg[
\sum_{k=s}^{t+T-1}\bigl[L_{k|t}(\alpha_{k|t})
+\tau1\{k=t\}K(\alpha_{k|t},\bar\alpha)\bigr]
+\lambda_t\left(\frac{\rho_{t+T|t}}{T}-\bar\alpha\right)_+
\Bigg]\\
&\qquad\ge\E_{\beta\sim P_t^{(B)}}[-|\lambda_t|]
=-|\lambda_t|.
\end{aligned}
\label{eq:policy-expected-cost-lower-bound}
\end{equation}
Since $\widehat J_{s|t}(r)$ is the infimum over $\pi\in\Pi_t$ of the
expected costs in \eqref{eq:policy-expected-cost-lower-bound},
\begin{equation}
\widehat J_{s|t}(r)\ge\inf_{\pi\in\Pi_t}(-|\lambda_t|)=-|\lambda_t|.
\label{eq:policy-value-lower-bound}
\end{equation}
For the upper bound, choose $a_{k|t}=1$ at each remaining stage
$k=s,\ldots,t+T-1$. Since $\beta_{k|t}\in[0,1]$,
$1(1\ge\beta_{k|t})=1$, so \eqref{eq:bci-state-dynamics} gives the
terminal state $r+t+T-s\in\{0,\ldots,T\}$.
Then
\begin{equation}
\begin{aligned}
-|\lambda_t|&\le\widehat J_{s|t}(r)\\
&\le\sum_{k=s}^{t+T-1}\bigl[L_{k|t}(1)+\tau1\{k=t\}K(1,\bar\alpha)\bigr]
+\lambda_t\left(\frac{r+t+T-s}{T}-\bar\alpha\right)_+\\
&=\tau1\{s=t\}K(1,\bar\alpha)
+\lambda_t\left(\frac{r+t+T-s}{T}-\bar\alpha\right)_+\\
&\le|\lambda_t|+\tau K(1,\bar\alpha)<\infty.
\end{aligned}
\label{eq:policy-finite-values}
\end{equation}
Therefore, \eqref{eq:policy-finite-values} shows that
$\widehat J_{s|t}(r)$ is bounded from below and above.
The $1$-Lipschitz property of $x\mapsto(x-\bar\alpha)_+$
gives
\begin{equation}
|\widehat J_{t+T|t}(r+1)-\widehat J_{t+T|t}(r)|
\le\frac{|\lambda_t|}{T},\qquad r=0,\ldots,T-1.
\label{eq:policy-terminal-increment}
\end{equation}
Suppose that $|\widehat J_{s+1|t}(r+1)-\widehat J_{s+1|t}(r)|
\le|\lambda_t|/T$ for $r=0,\ldots,s-t$.
For $r=0,\ldots,s-t-1$ and any $a$ with $L_{s|t}(a)<\infty$,
\eqref{eq:policy-bellman-recursion} gives
\begin{equation}
\begin{aligned}
\widehat Q_{s|t}(r+1,a)-\widehat Q_{s|t}(r,a)
={}&\bigl(1-\widehat G_t^{(B)}(a)\bigr)
\bigl[\widehat J_{s+1|t}(r+1)-\widehat J_{s+1|t}(r)\bigr]\\
&+\widehat G_t^{(B)}(a)
\bigl[\widehat J_{s+1|t}(r+2)-\widehat J_{s+1|t}(r+1)\bigr].
\end{aligned}
\label{eq:policy-adjacent-objectives}
\end{equation}
The two coefficients in \eqref{eq:policy-adjacent-objectives} are
nonnegative and sum to one. The induction hypothesis therefore implies
\begin{equation}
\widehat Q_{s|t}(r,a)-\frac{|\lambda_t|}{T}
\le\widehat Q_{s|t}(r+1,a)
\le\widehat Q_{s|t}(r,a)+\frac{|\lambda_t|}{T}.
\label{eq:policy-adjacent-objective-bounds}
\end{equation}
Taking infima over $a$ in \eqref{eq:policy-adjacent-objective-bounds} gives
\begin{equation}
\widehat J_{s|t}(r)-\frac{|\lambda_t|}{T}
\le\widehat J_{s|t}(r+1)
\le\widehat J_{s|t}(r)+\frac{|\lambda_t|}{T}.
\label{eq:policy-adjacent-value-bounds}
\end{equation}
Starting from \eqref{eq:policy-terminal-increment}, backward induction
using \eqref{eq:policy-adjacent-value-bounds} proves
\eqref{eq:policy-continuation-bound}.
\end{proof}

\subsection{Proof of Theorem \ref{thm:oracle}}
\begin{proof}
Work on $\mathcal A_t$ defined in \eqref{eq:policy-historical-event}.
Fix $\pi\in\Pi_t$ and apply the dynamics in
\eqref{eq:bci-state-dynamics} to
$\beta=(\beta_t,\ldots,\beta_{t+T-1})$. Write the resulting variables as
$\rho_{s|t}^{\pi}$ and $\alpha_{s|t}^{\pi}$. Since the latter is known
before $\beta_s$ is observed, the definition of $G_s^\circ$ gives
\begin{equation}
\begin{aligned}
\P(\beta_s\le\alpha_{s|t}^{\pi}\mid\F_{s-1})
&=G_s^\circ(\alpha_{s|t}^{\pi}),\\
\E\bigl[\widehat J_{s+1|t}(\rho_{s+1|t}^{\pi})\mid\F_{s-1}\bigr]
&=\widehat J_{s+1|t}(\rho_{s|t}^{\pi})
+G_s^\circ(\alpha_{s|t}^{\pi})
\widehat D_{s|t}(\rho_{s|t}^{\pi}).
\end{aligned}
\label{eq:policy-true-transition-value}
\end{equation}
Combining \eqref{eq:policy-true-transition-value} with
\eqref{eq:policy-bellman-recursion} yields the identity
\begin{equation}
\begin{aligned}
&L_{s|t}(\alpha_{s|t}^{\pi})
+\tau1\{s=t\}K(\alpha_{s|t}^{\pi},\bar\alpha)
+\E\bigl[\widehat J_{s+1|t}(\rho_{s+1|t}^{\pi})\mid\F_{s-1}\bigr]
-\widehat J_{s|t}(\rho_{s|t}^{\pi})\\
&\quad=\widehat Q_{s|t}(\rho_{s|t}^{\pi},\alpha_{s|t}^{\pi})
-\widehat J_{s|t}(\rho_{s|t}^{\pi})
+
\bigl[G_s^\circ(\alpha_{s|t}^{\pi})
-\widehat G_t^{(B)}(\alpha_{s|t}^{\pi})\bigr]
\widehat D_{s|t}(\rho_{s|t}^{\pi}).
\end{aligned}
\label{eq:policy-bellman-residual-identity}
\end{equation}
By Lemmas~\ref{lem:policy-cdf-comparison} and~\ref{lem:policy-value-increments},
\begin{equation}
\left|
\bigl[G_s^\circ(\alpha_{s|t}^{\pi})
-\widehat G_t^{(B)}(\alpha_{s|t}^{\pi})\bigr]
\widehat D_{s|t}(\rho_{s|t}^{\pi})
\right|
\le\frac{|\lambda_t|}{T}d_{s|t}.
\label{eq:policy-residual-error-bound}
\end{equation}
By \eqref{eq:policy-exact-selection}, we have
\begin{equation}
\widehat Q_{s|t}(\rho_{s|t}^{\pi},\alpha_{s|t}^{\pi})
-\widehat J_{s|t}(\rho_{s|t}^{\pi})\ge0.
\label{eq:policy-nonnegative-residual}
\end{equation}
For $\pi=\widehat\pi_t$, \eqref{eq:policy-exact-selection} gives equality
in \eqref{eq:policy-nonnegative-residual}.
By \eqref{eq:policy-finite-values}, for $s=t,\ldots,t+T-1$ and $r=0,\ldots,s-t$,
\[
\widehat J_{s+1|t}(r)\ge-|\lambda_t|,
\quad \widehat J_{s+1|t}(r+1)\ge-|\lambda_t|.
\]
Since $L_{s|t}\ge0$, $\tau\ge0$, $K(\cdot,\bar\alpha)\ge0$, and
$0\le\widehat G_t^{(B)}(\widehat\pi_{s|t}(r))\le1$,
\eqref{eq:policy-bellman-recursion} and
\eqref{eq:policy-exact-selection} give
\begin{equation}
\begin{aligned}
0&\le L_{s|t}\bigl(\widehat\pi_{s|t}(r)\bigr)
+\tau1\{s=t\}K\bigl(\widehat\pi_{s|t}(r),\bar\alpha\bigr)
\\
&=\widehat J_{s|t}(r)
-\bigl(1-\widehat G_t^{(B)}(\widehat\pi_{s|t}(r))\bigr)\widehat J_{s+1|t}(r)
-\widehat G_t^{(B)}(\widehat\pi_{s|t}(r))\widehat J_{s+1|t}(r+1)\\
&\le\widehat J_{s|t}(r)
+\bigl(1-\widehat G_t^{(B)}(\widehat\pi_{s|t}(r))\bigr)|\lambda_t|
+\widehat G_t^{(B)}(\widehat\pi_{s|t}(r))|\lambda_t|\\
&=\widehat J_{s|t}(r)+|\lambda_t|\\
&\le2|\lambda_t|+\tau K(1,\bar\alpha).
\end{aligned}
\label{eq:policy-selected-stage-cost-bound}
\end{equation}
By \eqref{eq:policy-selected-stage-cost-bound} and
\eqref{eq:bci-terminal-cost}, the bound is finite.
\begin{equation}
-|\lambda_t|\le J_t^{P_t^\star}(\widehat\pi_t)
\le T\bigl(2|\lambda_t|+\tau K(1,\bar\alpha)\bigr)+|\lambda_t|<\infty.
\label{eq:policy-selected-true-cost-finite}
\end{equation}
By \eqref{eq:policy-finite-values} and \eqref{eq:policy-cdf-error}, on
$\{J_t^{P_t^\star}(\pi)=\infty\}$,
\begin{equation}
J_t^{P_t^\star}(\pi)=+\infty>
\widehat J_{t|t}(0)-\frac{|\lambda_t|}{T}
\sum_{s=t}^{t+T-1}d_{s|t}>-\infty.
\label{eq:policy-infinite-objective-bound}
\end{equation}
Now work on $\{J_t^{P_t^\star}(\pi)<\infty\}\in\F_{t-1}$.
Since $\rho_{t|t}^{\pi}=0$ and
$\alpha_{t|t}^{\pi}=\pi_{t|t}(0)$, \eqref{eq:policy-objective} and
\eqref{eq:bci-terminal-cost} give
\begin{equation}
\begin{aligned}
0&\le\E\Bigg[\sum_{s=t}^{t+T-1}
\bigl\{L_{s|t}(\alpha_{s|t}^{\pi})
+\tau1\{s=t\}K(\alpha_{s|t}^{\pi},\bar\alpha)\bigr\}
\,\Bigm|\,\F_{t-1}\Bigg]\\
&=J_t^{P_t^\star}(\pi)
-\E\bigl[\widehat J_{t+T|t}(\rho_{t+T|t}^{\pi})\mid\F_{t-1}\bigr]\\
&\le J_t^{P_t^\star}(\pi)+|\lambda_t|<\infty.
\end{aligned}
\label{eq:policy-conditional-objective-decomposition}
\end{equation}
By \eqref{eq:policy-finite-values},
\begin{equation}
\E\bigl[|\widehat J_{s|t}(\rho_{s|t}^{\pi})|\mid\F_{t-1}\bigr]
\le|\lambda_t|+\tau K(1,\bar\alpha)<\infty,
\qquad s=t,\ldots,t+T.
\label{eq:policy-value-conditional-integrability}
\end{equation}
Since $\F_{t-1}\subseteq\F_{s-1}$ and $\widehat J_{t|t}(0)$ is
$\F_{t-1}$-measurable, \eqref{eq:policy-value-conditional-integrability}
and the tower property give
\begin{equation}
\begin{aligned}
&\sum_{s=t}^{t+T-1}\E\Big[
\E\bigl[\widehat J_{s+1|t}(\rho_{s+1|t}^{\pi})\mid\F_{s-1}\bigr]
-\widehat J_{s|t}(\rho_{s|t}^{\pi})\,\Bigm|\,\F_{t-1}\Big]\\
&\quad=\sum_{s=t}^{t+T-1}\Big\{
\E\bigl[\widehat J_{s+1|t}(\rho_{s+1|t}^{\pi})\mid\F_{t-1}\bigr]
-\E\bigl[\widehat J_{s|t}(\rho_{s|t}^{\pi})\mid\F_{t-1}\bigr]
\Big\}\\
&\quad=\sum_{s=t+1}^{t+T}
\E\bigl[\widehat J_{s|t}(\rho_{s|t}^{\pi})\mid\F_{t-1}\bigr]
-\sum_{s=t}^{t+T-1}
\E\bigl[\widehat J_{s|t}(\rho_{s|t}^{\pi})\mid\F_{t-1}\bigr]\\
&\quad=\E\bigl[\widehat J_{t+T|t}(\rho_{t+T|t}^{\pi})\mid\F_{t-1}\bigr]
-\widehat J_{t|t}(0).
\end{aligned}
\label{eq:policy-conditional-value-telescoping}
\end{equation}
Taking $\F_{t-1}$-conditional expectations in
\eqref{eq:policy-bellman-residual-identity}, summing over $s$, and using
\eqref{eq:policy-conditional-objective-decomposition} and
\eqref{eq:policy-conditional-value-telescoping} gives
\begin{equation}
J_t^{P_t^\star}(\pi)-\widehat J_{t|t}(0)
=\E\Bigg[\sum_{s=t}^{t+T-1}\Big\{
\widehat Q_{s|t}(\rho_{s|t}^{\pi},\alpha_{s|t}^{\pi})
-\widehat J_{s|t}(\rho_{s|t}^{\pi})
+\bigl[G_s^\circ(\alpha_{s|t}^{\pi})
-\widehat G_t^{(B)}(\alpha_{s|t}^{\pi})\bigr]
\widehat D_{s|t}(\rho_{s|t}^{\pi})
\Big\}\,\Bigm|\,\F_{t-1}\Bigg].
\label{eq:policy-residual-telescoping}
\end{equation}
Since $\mathcal A_t\in\F_{t-1}$, the
bounds \eqref{eq:policy-residual-error-bound} and
\eqref{eq:policy-nonnegative-residual} hold throughout these conditional
expectations. By \eqref{eq:policy-selected-true-cost-finite},
\eqref{eq:policy-residual-telescoping} applies to $\pi=\widehat\pi_t$.
Applying \eqref{eq:policy-residual-error-bound} and
\eqref{eq:policy-nonnegative-residual} to
\eqref{eq:policy-residual-telescoping}, and using
\eqref{eq:policy-infinite-objective-bound} when
$J_t^{P_t^\star}(\pi)=\infty$, gives
\begin{equation}
\begin{aligned}
J_t^{P_t^\star}(\widehat\pi_t)
&\le\widehat J_{t|t}(0)
+\frac{|\lambda_t|}{T}\sum_{s=t}^{t+T-1}d_{s|t},\\
J_t^{P_t^\star}(\pi)
&\ge\widehat J_{t|t}(0)
-\frac{|\lambda_t|}{T}\sum_{s=t}^{t+T-1}d_{s|t},
\qquad\pi\in\Pi_t.
\end{aligned}
\label{eq:policy-two-value-bounds}
\end{equation}
Subtracting the two inequalities in \eqref{eq:policy-two-value-bounds},
\begin{equation}
\begin{aligned}
0\le J_t^{P_t^\star}(\widehat\pi_t)
-\inf_{\pi\in\Pi_t}J_t^{P_t^\star}(\pi)
&\le\frac{2|\lambda_t|}{T}\sum_{s=t}^{t+T-1}d_{s|t}\\
&=2|\lambda_t|\left[
e_B(\delta)+\frac1B\sum_{j=1}^B\kappa_{t-j}
+\frac1T\sum_{s=t}^{t+T-1}\kappa_s
\right].
\end{aligned}
\label{eq:policy-refined-oracle-bound}
\end{equation}
By monotonicity of $(\kappa_s)$,
\begin{equation}
\frac1B\sum_{j=1}^B\kappa_{t-j}
+\frac1T\sum_{s=t}^{t+T-1}\kappa_s\le2\kappa_{t-B}.
\label{eq:policy-calibration-simplification}
\end{equation}
Substitute \eqref{eq:policy-cdf-error} and
\eqref{eq:policy-calibration-simplification} into
\eqref{eq:policy-refined-oracle-bound} to obtain
\eqref{eq:main-theorem-bound}. The probability statement follows from
\eqref{eq:policy-historical-event}.
\end{proof}

\section{A theoretical result on the length dominance of BCI}\label{subsec:dominance}

In this section, we present a setting in which the expected interval lengths of ACI is provably longer than those of BCI. 

\subsection{Setting}
We use the time indexing in Section~\ref{sec:setup}, so that $C_t$ predicts $Y_t$. Let $V_t$ be the covariate vector observed before $Y_t$, and define
\begin{equation}
\begin{aligned}
W_t&:=\bigl(V_t,(V_{t-1},Y_{t-1}),\dots,(V_{t-p},Y_{t-p})\bigr),\\
\mathcal H_{t-1}&:=\sigma\bigl((V_s,Y_s):s<t\bigr),
\qquad
\mathcal G_{t-1}:=\mathcal H_{t-1}\vee\sigma(V_t).
\end{aligned}
\label{eq:dominance-information}
\end{equation}
Use the interval $[\ell_a,h_a]$ from Assumption~\ref{ass:finite-actions}. 

For ACI, we consider a truncated version as in Assumption~\ref{ass:finite-actions} so that the update can be made stationary. With step size $\gamma_A>0$, the update in \eqref{eqn:aci} becomes
\begin{equation}
\begin{aligned}
\alpha_t^A&:=
\begin{cases}
0,&a_t<\ell_a,\\
a_t,&\ell_a\le a_t\le h_a,\\
1,&a_t>h_a,
\end{cases}\\
\err_t^A&:=1\{\beta_t\le\alpha_t^A\},
\qquad
a_{t+1}:=a_t+\gamma_A(\bar\alpha-\err_t^A),
\end{aligned}
\label{eq:dominance-aci}
\end{equation}
where $\beta_t$ is the PIT in \eqref{eq:beta_t}.

For BCI, consider $\tau>0$ and
\begin{equation}
K(b,a):=b\log\frac ba+(1-b)\log\frac{1-b}{1-a},
\qquad b\in[0,1],\quad a\in(0,1),
\label{eq:dominance-kl}
\end{equation}
with $0\log0:=0$. Similar to \eqref{eqn:tilde_alpha}, define
\begin{equation}
\begin{aligned}
\alpha_t^B&\in\argmin_{a\in[\ell_a,h_a]}
\left\{|C_t(1-a)|+D_{t|t}(0)a+\tau K(a,\bar\alpha)\right\},
&&0\le\lambda_t\le\lambda_{\max},\\
\alpha_t^B&=1,&&\lambda_t<0,\\
\alpha_t^B&=0,&&\lambda_t>\lambda_{\max}.
\end{aligned}
\label{eq:dominance-bci}
\end{equation}
In particular, \eqref{eq:dominance-bci} uses the minimizer at $\lambda_t=0$ and $\lambda_t=\lambda_{\max}$. Retain $\gamma=c\lambda_{\max}$ with $c\in(0,1)$ and the update in \eqref{eqn:lambda-update}:
\begin{equation}
\err_t^B:=1\{\beta_t\le\alpha_t^B\},
\qquad
\lambda_{t+1}:=\lambda_t+c\lambda_{\max}(\err_t^B-\bar\alpha).
\label{eq:dominance-bci-update}
\end{equation}

\subsection{Assumptions}

\begin{assumption}
\label{ass:dominance-calibration}
$Y_t$ given $\mathcal G_{t-1}$ has a continuous CDF $F^\star(W_t,\cdot)$ and the $(C_t,\alpha_t^B)$ are $\mathcal G_{t-1}$-measurable, and
\begin{equation}
\Law(\beta_t\mid\mathcal G_{t-1})=\mathrm{Unif}[0,1].
\label{eq:dominance-calibration}
\end{equation}
Further, $(\alpha_t,\lambda_t,D_{t|t}(0))$ are $\mathcal H_{t-1}$-measurable.
\end{assumption}

\begin{remark}
\label{rem:dominance-calibration}
If the conditional CDF is known, then Assumption~\ref{ass:pred-sets} and \eqref{eq:dominance-calibration} hold with
\begin{equation}
\begin{aligned}
C_t(1-a)&=\Bigl[(F^\star)^{-1}(W_t,a/2),
(F^\star)^{-1}(W_t,1-a/2)\Bigr],\qquad 0<a<1,\\
\beta_t&=2\min\left\{F^\star(W_t,Y_t),1-F^\star(W_t,Y_t)\right\}.
\end{aligned}
\label{eq:dominance-central-intervals}
\end{equation}
If the multi-step ahead forecast intervals and $\lambda_t$ are $\mathcal H_{t-1}$-measurable, \eqref{eqn:bci-bellman-update} also makes $D_{t|t}(0)$ $\mathcal H_{t-1}$-measurable.
\end{remark}

\begin{assumption}
\label{ass:dominance-length}
There are constants $\bar s<\infty$ and $\mu_S>0$, a $\mathcal G_{t-1}$-measurable variable $S_t\in[0,\bar s]$, and a nonincreasing function $w:[0,1]\to[0,\infty)$, differentiable at $\bar\alpha$, such that
\begin{equation}
|C_t(1-a)|=S_tw(a),\qquad a\in[0,1],
\qquad
\E[S_t\mid\mathcal H_{t-1}]=\mu_S.
\label{eq:dominance-length-scale}
\end{equation}
For constants $0<v_-\le v_+<\infty$,
\begin{equation}
v_-(a-\bar\alpha)^2
\le w(a)-w(\bar\alpha)-w'(\bar\alpha)(a-\bar\alpha)
\le v_+(a-\bar\alpha)^2,
\qquad a\in[0,1].
\label{eq:dominance-curvature}
\end{equation}
\end{assumption}

\begin{remark}
The requirement $\E[S_t\mid\mathcal H_{t-1}]=\mu_S$ in \eqref{eq:dominance-length-scale} holds, for example, if $S_t=\sigma_0(V_t)$ for a measurable function $\sigma_0$ taking values in $[0,\bar s]$, the covariates are stationary, $V_t$ is independent of $\mathcal H_{t-1}$, and $\E\sigma_0(V_t)>0$.
\end{remark}

\begin{assumption}
\label{ass:dominance-tuning}
There are finite constants $D_-\le D_+$ such that
\begin{equation}
D_-\le D_{t|t}(0)\le D_+.
\label{eq:dominance-continuation-bounds}
\end{equation}
Define
\begin{equation}
r_I:=\max\{h_a,1-\ell_a\},
\qquad
C_\star:=\max\left\{\bar s[-w'(\bar\alpha)]-D_-,D_+\right\}.
\label{eq:dominance-constants}
\end{equation}
The step sizes and regularization parameter satisfy
\begin{equation}
\begin{gathered}
\gamma_A(1-\bar\alpha)\le\ell_a,
\qquad \gamma_A\bar\alpha\le1-h_a,
\qquad \ell_a\le\bar\alpha\le h_a,\\
v_-\gamma_A>v_+c,
\qquad
\tau>\frac{v_+r_IC_\star}
{\bar\alpha(1-\bar\alpha)(v_-\gamma_A-v_+c)}.
\end{gathered}
\label{eq:dominance-tuning}
\end{equation}
\end{assumption}

\subsection{Main result}
With these assumptions, we can show that BCI has a strictly shorter expected interval length than ACI.

\begin{theorem}
\label{thm:dominance}
Under Assumptions~\ref{ass:pred-sets}, \ref{ass:dominance-calibration}, \ref{ass:dominance-length}, and \ref{ass:dominance-tuning}, suppose that the ACI path $(V_t,Y_t,a_t,\alpha_t^A)_{t\ge1}$ and the BCI path $(V_t,Y_t,\lambda_t,\alpha_t^B)_{t\ge1}$ are stationary. Then, for every $t$,
\begin{equation}
\E|C_t(1-\alpha_t^B)|<\E|C_t(1-\alpha_t^A)|.
\label{eq:dominance-conclusion}
\end{equation}
\end{theorem}

\begin{remark}
\label{rem:dominance-stationarity}
For ACI, \citet[Appendix~A.5]{gibbs2021adaptive} give sufficient conditions for the existence of an invariant distribution in a finite-state hidden Markov model. For BCI, Theorem~\ref{thm:stationarity}, formalized in Theorem~\ref{thm:cycle}, establishes an invariant distribution under its stated assumptions. Initializing the respective joint processes from these invariant distributions makes their paths stationary. Therefore, the stationarity of ACI and BCI paths is possible.
\end{remark}

\begin{proof}
We first verify that $a_t$ and $\lambda_t$ have finite second moments. For ACI, define
\begin{equation}
L_A:=\ell_a-\gamma_A(1-\bar\alpha),
\qquad U_A:=h_a+\gamma_A\bar\alpha,
\qquad \mathcal I_A:=[L_A,U_A]\subseteq[0,1],
\label{eq:dominance-aci-state-interval}
\end{equation}
where the inclusion follows from \eqref{eq:dominance-tuning}. For $a_t\in\mathcal I_A$, \eqref{eq:dominance-aci} gives, almost surely,
\begin{equation}
\begin{aligned}
a_{t+1}&\in
\begin{cases}
[L_A,\ell_a+\gamma_A\bar\alpha],&a_t<\ell_a,\\
[a_t-\gamma_A(1-\bar\alpha),a_t+\gamma_A\bar\alpha],&\ell_a\le a_t\le h_a,\\
[h_a-\gamma_A(1-\bar\alpha),U_A],&a_t>h_a,
\end{cases}\\
&\subseteq\mathcal I_A.
\end{aligned}
\label{eq:dominance-aci-invariance}
\end{equation}
If $a_1<L_A$, \eqref{eq:dominance-aci} gives
\begin{equation}
\begin{aligned}
n_-&:=\left\lceil\frac{L_A-a_1}{\gamma_A\bar\alpha}\right\rceil<\infty,\\
a_{1+k}&=a_1+k\gamma_A\bar\alpha<L_A,\qquad 0\le k<n_-,\\
a_{1+n_-}&=a_1+n_-\gamma_A\bar\alpha
\in[L_A,L_A+\gamma_A\bar\alpha)\subseteq\mathcal I_A.
\end{aligned}
\label{eq:dominance-aci-entry-below}
\end{equation}
If $a_1>U_A$, \eqref{eq:dominance-aci} gives, almost surely,
\begin{equation}
\begin{aligned}
n_+&:=\left\lceil\frac{a_1-U_A}{\gamma_A(1-\bar\alpha)}\right\rceil<\infty,\\
a_{1+k}&=a_1-k\gamma_A(1-\bar\alpha)>U_A,\qquad 0\le k<n_+,\\
a_{1+n_+}&=a_1-n_+\gamma_A(1-\bar\alpha)
\in(U_A-\gamma_A(1-\bar\alpha),U_A]\subseteq\mathcal I_A.
\end{aligned}
\label{eq:dominance-aci-entry-above}
\end{equation}
By \eqref{eq:dominance-aci-invariance}--\eqref{eq:dominance-aci-entry-above}, $1\{a_t\notin\mathcal I_A\}\to0$ almost surely. Stationarity and bounded convergence give
\begin{equation}
\P(a_1\notin\mathcal I_A)
=\lim_{t\to\infty}\P(a_t\notin\mathcal I_A)=0.
\label{eq:dominance-aci-state-support}
\end{equation}
For BCI, the interval $\I=[-\gamma\bar\alpha,\lambda_{\max}+\gamma(1-\bar\alpha)]$ is invariant under \eqref{eq:dominance-bci-update}, as shown in the proof of Theorem~\ref{thm:global-guarantee}: \eqref{eq:dominance-bci} gives $\alpha_t^B=1$ for $\lambda_t<0$ and $\alpha_t^B=0$ for $\lambda_t>\lambda_{\max}$. Write $L_B:=-\gamma\bar\alpha$ and $U_B:=\lambda_{\max}+\gamma(1-\bar\alpha)$. By \eqref{eq:dominance-bci}--\eqref{eq:dominance-calibration}, almost surely, if $\lambda_1<L_B$,
\begin{equation}
\begin{aligned}
n_{B,-}&:=\left\lceil\frac{L_B-\lambda_1}{\gamma(1-\bar\alpha)}\right\rceil<\infty,\\
\lambda_{1+k}&=\lambda_1+k\gamma(1-\bar\alpha)<L_B,\qquad 0\le k<n_{B,-},\\
\lambda_{1+n_{B,-}}&=\lambda_1+n_{B,-}\gamma(1-\bar\alpha)
\in[L_B,L_B+\gamma(1-\bar\alpha))\subseteq\I;
\end{aligned}
\label{eq:dominance-bci-entry-below}
\end{equation}
if $\lambda_1>U_B$,
\begin{equation}
\begin{aligned}
n_{B,+}&:=\left\lceil\frac{\lambda_1-U_B}{\gamma\bar\alpha}\right\rceil<\infty,\\
\lambda_{1+k}&=\lambda_1-k\gamma\bar\alpha>U_B,\qquad 0\le k<n_{B,+},\\
\lambda_{1+n_{B,+}}&=\lambda_1-n_{B,+}\gamma\bar\alpha
\in(U_B-\gamma\bar\alpha,U_B]\subseteq\I.
\end{aligned}
\label{eq:dominance-bci-entry-above}
\end{equation}
By invariance and \eqref{eq:dominance-bci-entry-below}--\eqref{eq:dominance-bci-entry-above}, $1\{\lambda_t\notin\I\}\to0$ almost surely. Stationarity gives $\P(\lambda_t\in\I)=1$ by the limit argument in \eqref{eq:dominance-aci-state-support}. Both $a_t$ and $\lambda_t$ are consequently bounded.

By Assumption~\ref{ass:dominance-calibration}, for $j\in\{A,B\}$,
\begin{equation}
\E[\err_t^j\mid\mathcal G_{t-1}]=\alpha_t^j,
\qquad
\E\err_t^j=\E \alpha_t^j.
\label{eq:dominance-conditional-errors}
\end{equation}
Taking expectations in the two stationary updates gives
\begin{equation}
\begin{aligned}
0&=\E(a_{t+1}-a_t)
=\gamma_A(\bar\alpha-\E \alpha_t^A),\\
0&=\E(\lambda_{t+1}-\lambda_t)
=\gamma(\E \alpha_t^B-\bar\alpha).
\end{aligned}
\label{eq:dominance-first-moments}
\end{equation}
Hence $\E \alpha_t^A=\E \alpha_t^B=\bar\alpha$. Since $(\err_t^j)^2=\err_t^j$, \eqref{eq:dominance-conditional-errors} and \eqref{eq:dominance-first-moments} imply
\begin{equation}
\E(\err_t^j-\bar\alpha)^2
=(1-2\bar\alpha)\E\err_t^j+\bar\alpha^2
=\bar\alpha(1-\bar\alpha),\qquad j\in\{A,B\}.
\label{eq:dominance-error-second-moments}
\end{equation}
As a result,
\begin{equation}
\begin{gathered}
0=\E(a_{t+1}^2-a_t^2)
=2\gamma_A\E[a_t(\bar\alpha-\alpha_t^A)]
+\gamma_A^2\bar\alpha(1-\bar\alpha),\\
\Longrightarrow\quad\E[a_t(\alpha_t^A-\bar\alpha)]
=\frac{\gamma_A}{2}\bar\alpha(1-\bar\alpha).
\end{gathered}
\label{eq:dominance-aci-cross-moment}
\end{equation}
Because $0\le a_t\le1$ and $\bar\alpha\in[\ell_a,h_a]$, \eqref{eq:dominance-aci} implies
\begin{equation}
(\alpha_t^A-a_t)(\alpha_t^A-\bar\alpha)=
\begin{cases}
a_t\bar\alpha\ge0,&a_t<\ell_a,\\
0,&\ell_a\le a_t\le h_a,\\
(1-a_t)(1-\bar\alpha)\ge0,&a_t>h_a.
\end{cases}
\label{eq:dominance-aci-clipping-sign}
\end{equation}
Using \eqref{eq:dominance-first-moments}, \eqref{eq:dominance-aci-cross-moment}, and \eqref{eq:dominance-aci-clipping-sign},
\begin{equation}
\E(\alpha_t^A-\bar\alpha)^2
=\E[a_t(\alpha_t^A-\bar\alpha)]
+\E[(\alpha_t^A-a_t)(\alpha_t^A-\bar\alpha)]
\ge\frac{\gamma_A}{2}\bar\alpha(1-\bar\alpha).
\label{eq:dominance-aci-variance}
\end{equation}

For BCI, we have
\begin{equation}
\begin{gathered}
0=\E(\lambda_{t+1}^2-\lambda_t^2)
=2\gamma\E[\lambda_t(\alpha_t^B-\bar\alpha)]
+\gamma^2\bar\alpha(1-\bar\alpha),\\
\Longrightarrow\quad\E[\lambda_t(\alpha_t^B-\bar\alpha)]
=-\frac{c\lambda_{\max}}2\bar\alpha(1-\bar\alpha).
\end{gathered}
\label{eq:dominance-bci-cross-moment}
\end{equation}
Define $z_t^B:=1-\lambda_t/\lambda_{\max}$. By \eqref{eq:dominance-first-moments} and \eqref{eq:dominance-bci-cross-moment},
\begin{equation}
\E[z_t^B(\alpha_t^B-\bar\alpha)]
=\E(\alpha_t^B-\bar\alpha)
-\frac{\E[\lambda_t(\alpha_t^B-\bar\alpha)]}{\lambda_{\max}}
=\frac c2\bar\alpha(1-\bar\alpha).
\label{eq:dominance-bci-normalized-moment}
\end{equation}

We next bound the deviation $\alpha_t^B-\bar\alpha$ when $0\le\lambda_t\le\lambda_{\max}$. The feasible action $\bar\alpha$ in \eqref{eq:dominance-bci} gives
\begin{equation}
S_t\{w(\alpha_t^B)-w(\bar\alpha)\}
+D_{t|t}(0)(\alpha_t^B-\bar\alpha)+\tau K(\alpha_t^B,\bar\alpha)\le0.
\label{eq:dominance-bci-optimality}
\end{equation}
For the KL divergence in \eqref{eq:dominance-kl}, it is well-known that
\begin{equation}
\begin{gathered}
K(\bar\alpha,\bar\alpha)=0,
\qquad
\left.\frac{\partial K(b,\bar\alpha)}{\partial b}\right|_{b=\bar\alpha}=0,
\\
\frac{\partial^2 K(b,\bar\alpha)}{\partial b^2}
=\frac1{b(1-b)}\ge4,\qquad b\in(0,1),\\
K(b,\bar\alpha)\ge2(b-\bar\alpha)^2,\qquad b\in[0,1],
\end{gathered}
\label{eq:dominance-kl-quadratic}
\end{equation}
Since $w'(\bar\alpha)\le0$, \eqref{eq:dominance-continuation-bounds} implies
\begin{equation}
-D_+\le S_t[-w'(\bar\alpha)]-D_{t|t}(0)
\le\bar s[-w'(\bar\alpha)]-D_-.
\label{eq:dominance-linear-coefficient}
\end{equation}
Combining \eqref{eq:dominance-curvature}, \eqref{eq:dominance-bci-optimality}, \eqref{eq:dominance-kl-quadratic}, and \eqref{eq:dominance-linear-coefficient} gives
\begin{equation}
\begin{aligned}
2\tau(\alpha_t^B-\bar\alpha)^2
&\le\tau K(\alpha_t^B,\bar\alpha)\\
&\le\{S_t[-w'(\bar\alpha)]-D_{t|t}(0)\}(\alpha_t^B-\bar\alpha)
-S_tv_-(\alpha_t^B-\bar\alpha)^2\\
&\le C_\star|\alpha_t^B-\bar\alpha|.
\end{aligned}
\label{eq:dominance-bci-deviation-inequality}
\end{equation}
If $\alpha_t^B-\bar\alpha\ne0$, division by $2\tau|\alpha_t^B-\bar\alpha|$ yields
\begin{equation}
|\alpha_t^B-\bar\alpha|\le\frac{C_\star}{2\tau},
\qquad 0\le\lambda_t\le\lambda_{\max};
\label{eq:dominance-bci-deviation}
\end{equation}
the inequality also holds when $\alpha_t^B-\bar\alpha=0$.

Decompose the BCI variance using \eqref{eq:dominance-first-moments}:
\begin{equation}
\E(\alpha_t^B-\bar\alpha)^2
=\E[z_t^B(\alpha_t^B-\bar\alpha)]+\E[(\alpha_t^B-z_t^B)(\alpha_t^B-\bar\alpha)].
\label{eq:dominance-bci-variance-decomposition}
\end{equation}
For $\lambda_t<0$, $\alpha_t^B=1$ and $z_t^B>1$; for $\lambda_t>\lambda_{\max}$, $\alpha_t^B=0$ and $z_t^B<0$. Therefore
\begin{equation}
(\alpha_t^B-z_t^B)(\alpha_t^B-\bar\alpha)\le0,
\qquad \lambda_t\notin[0,\lambda_{\max}].
\label{eq:dominance-bci-boundary-sign}
\end{equation}
For $0\le\lambda_t\le\lambda_{\max}$, $z_t^B\in[0,1]$ and $\alpha_t^B\in[\ell_a,h_a]$, so
\begin{equation}
|\alpha_t^B-z_t^B|\le\max\{h_a,1-\ell_a\}=r_I.
\label{eq:dominance-bci-interior-distance}
\end{equation}
Substituting \eqref{eq:dominance-bci-normalized-moment}, \eqref{eq:dominance-bci-deviation}, \eqref{eq:dominance-bci-boundary-sign}, and \eqref{eq:dominance-bci-interior-distance} into \eqref{eq:dominance-bci-variance-decomposition} yields
\begin{equation}
\E(\alpha_t^B-\bar\alpha)^2
\le\frac c2\bar\alpha(1-\bar\alpha)+\frac{r_IC_\star}{2\tau}.
\label{eq:dominance-bci-variance}
\end{equation}

It remains to account for the dependence between $S_t$ and $\alpha_t^B$. By \eqref{eq:dominance-curvature}, for $a\in[0,1]$,
\begin{equation}
\begin{aligned}
0\le w(a)&\le w(\bar\alpha)+w'(\bar\alpha)(a-\bar\alpha)+v_+(a-\bar\alpha)^2\\
&\le w(\bar\alpha)+|w'(\bar\alpha)|+v_+=:C_w<\infty.
\end{aligned}
\label{eq:dominance-uniform-length-bound}
\end{equation}
Since $0\le S_t\le\bar s$, \eqref{eq:dominance-length-scale} and \eqref{eq:dominance-uniform-length-bound} give, for $j\in\{A,B\}$,
\begin{equation}
\begin{aligned}
\E|C_t(1-\alpha_t^j)|&=\E[S_tw(\alpha_t^j)]\le\bar s C_w<\infty,\\
\E\bigl|\operatorname{Cov}(S_t,w(\alpha_t^j)\mid\mathcal H_{t-1})\bigr|
&\le\E[S_tw(\alpha_t^j)]
+\E\bigl\{\E[S_t\mid\mathcal H_{t-1}]\E[w(\alpha_t^j)\mid\mathcal H_{t-1}]\bigr\}\\
&\le2\bar s C_w<\infty.
\end{aligned}
\label{eq:dominance-length-covariance-integrability}
\end{equation}
For ACI, $\alpha_t^A$ is $\mathcal H_{t-1}$-measurable, so \eqref{eq:dominance-length-scale} gives
\begin{equation}
\E[S_tw(\alpha_t^A)]
=\E\{w(\alpha_t^A)\E[S_t\mid\mathcal H_{t-1}]\}
=\mu_S\E w(\alpha_t^A).
\label{eq:dominance-aci-length}
\end{equation}
For BCI, condition on $\mathcal H_{t-1}$, so that $\lambda_t$ and $D_{t|t}(0)$ are fixed. When $0\le\lambda_t\le\lambda_{\max}$, define
\begin{equation}
L_t(a,s):=sw(a)+D_{t|t}(0)a+\tau K(a,\bar\alpha).
\label{eq:dominance-scaled-objective}
\end{equation}
For $s_1<s_2$, let $b_i$ be any minimizer of $L_t(\cdot,s_i)$ on $[\ell_a,h_a]$, for $i=1,2$. Optimality of $b_i$ implies
\begin{equation}
\begin{gathered}
L_t(b_1,s_1)\le L_t(b_2,s_1),\qquad L_t(b_2,s_2)\le L_t(b_1,s_2),\\
(s_2-s_1)\{w(b_2)-w(b_1)\}\le0.
\end{gathered}
\label{eq:dominance-length-monotonicity}
\end{equation}
Thus $w(b_2)\le w(b_1)$, including when minimizers are nonunique. If $\lambda_t\notin[0,\lambda_{\max}]$, $\alpha_t^B$ is fixed given $\mathcal H_{t-1}$. Let $(S_t',b_t')$ be a conditionally independent copy of $(S_t,\alpha_t^B)$ given $\mathcal H_{t-1}$. By \eqref{eq:dominance-length-monotonicity}
\begin{equation}
\begin{aligned}
2\operatorname{Cov}(S_t,w(\alpha_t^B)\mid\mathcal H_{t-1})
&=\E\left[(S_t-S_t')\{w(\alpha_t^B)-w(b_t')\}
\mid\mathcal H_{t-1}\right]\\
&\le0.
\end{aligned}
\label{eq:dominance-negative-covariance}
\end{equation}
By \eqref{eq:dominance-length-scale} and \eqref{eq:dominance-negative-covariance},
\begin{equation}
\begin{aligned}
&\E[S_tw(\alpha_t^B)]\\
&=\E\operatorname{Cov}(S_t,w(\alpha_t^B)\mid\mathcal H_{t-1})
+\E\left\{\E[S_t\mid\mathcal H_{t-1}]
\E[w(\alpha_t^B)\mid\mathcal H_{t-1}]\right\}\\
&\le\mu_S\E w(\alpha_t^B).
\end{aligned}
\label{eq:dominance-bci-length}
\end{equation}

Finally, \eqref{eq:dominance-curvature} and $\E \alpha_t^A=\E \alpha_t^B=\bar\alpha$ give
\begin{equation}
\begin{aligned}
\E w(\alpha_t^A)&\ge w(\bar\alpha)+v_-\E(\alpha_t^A-\bar\alpha)^2,\\
\E w(\alpha_t^B)&\le w(\bar\alpha)+v_+\E(\alpha_t^B-\bar\alpha)^2.
\end{aligned}
\label{eq:dominance-expected-curvature}
\end{equation}
Combining \eqref{eq:dominance-aci-length}, \eqref{eq:dominance-bci-length}, and \eqref{eq:dominance-expected-curvature} with the variance bounds \eqref{eq:dominance-aci-variance} and \eqref{eq:dominance-bci-variance},
\begin{equation}
\begin{aligned}
\E|C_t(1-\alpha_t^A)|-\E|C_t(1-\alpha_t^B)|
&\ge\mu_S\{\E w(\alpha_t^A)-\E w(\alpha_t^B)\}\\
&\ge\mu_S\left\{v_-\E(\alpha_t^A-\bar\alpha)^2
-v_+\E(\alpha_t^B-\bar\alpha)^2\right\}\\
&\ge\frac{\mu_S}{2}\left\{
\bar\alpha(1-\bar\alpha)(v_-\gamma_A-v_+c)
-\frac{v_+r_IC_\star}{\tau}\right\}>0,
\end{aligned}
\label{eq:dominance-length-gap}
\end{equation}
where strict positivity follows from \eqref{eq:dominance-tuning}. This proves \eqref{eq:dominance-conclusion}.
\end{proof}

%%%%%%%%%%%%%%%%%%%%%%%%%%%%%%%%%%%%%%%%%%%%%%%%%%%%%%%%%%%%%%%%%%%%%%%%%%%%%%%
%%%%%%%%%%%%%%%%%%%%%%%%%%%%%%%%%%%%%%%%%%%%%%%%%%%%%%%%%%%%%%%%%%%%%%%%%%%%%%%

% This document was modified from the file originally made available by
% Pat Langley and Andrea Danyluk for ICML-2K. This version was created
% by Iain Murray in 2018, and modified by Alexandre Bouchard in
% 2019 and 2021 and by Csaba Szepesvari, Gang Niu and Sivan Sabato in 2022.
% Modified again in 2023 and 2024 by Sivan Sabato and Jonathan Scarlett.
% Previous contributors include Dan Roy, Lise Getoor and Tobias
% Scheffer, which was slightly modified from the 2010 version by
% Thorsten Joachims & Johannes Fuernkranz, slightly modified from the
% 2009 version by Kiri Wagstaff and Sam Roweis's 2008 version, which is
% slightly modified from Prasad Tadepalli's 2007 version which is a
% lightly changed version of the previous year's version by Andrew
% Moore, which was in turn edited from those of Kristian Kersting and
% Codrina Lauth. Alex Smola contributed to the algorithmic style files.

%%%%%%%%%%%%%%%%%%%%%%%%%%%%%%%%%%%%%%%%%%%%%%%%%%%%%%%%%%%%

\end{document}